\documentclass{article}
\usepackage{iclr2027_conference,times}

\usepackage{amsmath,amsfonts,bm}

\def\eqref#1{equation~\ref{#1}}
\def\1{\bm{1}}

\def\vh{{\bm{h}}}

\def\vk{{\bm{k}}}

\def\vq{{\bm{q}}}

\def\mR{{\bm{R}}}

\DeclareMathAlphabet{\mathsfit}{\encodingdefault}{\sfdefault}{m}{sl}
\SetMathAlphabet{\mathsfit}{bold}{\encodingdefault}{\sfdefault}{bx}{n}

\newcommand{\KL}{D_{\mathrm{KL}}}

\iclrfinalcopy

\usepackage{graphicx}
\usepackage{hyperref}
\usepackage{url}

\usepackage{pifont}
\usepackage{booktabs}
\usepackage[table]{xcolor}
\definecolor{sectionblue}{RGB}{232,239,246}
\definecolor{modelgray}{RGB}{245,245,245}
\newcommand{\NA}{\multicolumn{1}{c}{--}}
\usepackage{multirow}
\usepackage[font=small,skip=10pt]{caption}

\usepackage{tcolorbox}
\usepackage{longtable}
\usepackage{placeins}
\usepackage{array}
\usepackage{tabularx}
\newcolumntype{L}[1]{>{\raggedright\arraybackslash}p{#1}}
\newcolumntype{Y}{>{\raggedright\arraybackslash}X}
\tcbuselibrary{skins,breakable,listings}

\lstdefinestyle{DCLPromptText}{
  language={},
  basicstyle=\footnotesize\ttfamily,
  columns=fullflexible,
  keepspaces=true,
  showstringspaces=false,
  showspaces=false,
  showtabs=false,
  breaklines=true,
  breakatwhitespace=false,
  breakautoindent=true,
  breakindent=1em,
  tabsize=2,
  numbers=none,
  aboveskip=0pt,
  belowskip=0pt,
  literate={—}{{\textemdash}}1 {–}{{\textendash}}1
}

\newtcblisting{DCLPromptBox}[2][]{
  enhanced jigsaw,
  breakable,
  listing only,
  listing engine=listings,
  listing options={style=DCLPromptText},
  title={#2},
  title after break={#2\space(continued)},
  fonttitle=\small\bfseries,
  coltitle=black,
  colback=black!2,
  colbacktitle=black!7,
  colframe=black!25,
  boxrule=0.4pt,
  arc=1mm,
  left=3mm,
  right=3mm,
  top=2mm,
  bottom=2mm,
  before skip=10pt,
  after skip=10pt,
  #1
}

\newtcolorbox{DCLTextPromptBox}[2][]{
  enhanced jigsaw,
  title={#2},
  fonttitle=\small\bfseries,
  fontupper=\footnotesize\rmfamily,
  before upper={\raggedright\setlength{\parindent}{0pt}\setlength{\parskip}{0pt}},
  coltitle=black,
  colback=black!2,
  colbacktitle=black!7,
  colframe=black!25,
  boxrule=0.4pt,
  arc=1mm,
  left=3mm,
  right=3mm,
  top=1mm,
  bottom=1mm,
  before skip=10pt,
  after skip=10pt,
  #1
}

\makeatletter
\newcommand{\DCLAppendixFloatSetup}{%
  \let\ftype@table\ftype@figure
  \setlength{\@fptop}{0pt}%
  \setlength{\@fpsep}{9pt}%
  \setlength{\@fpbot}{0pt plus 1fil}%
}
\makeatother

\newcolumntype{M}[1]{>{\raggedright\arraybackslash}m{#1}}
\newcolumntype{C}[1]{>{\centering\arraybackslash}p{#1}}
\title{Diagnosing and Mitigating Long-Session\\Degradation in Full-Duplex Speech Models}

\author{%
  \textbf{Junho Kim}\textsuperscript{1} \quad
  \textbf{Che Hyun Lee}\textsuperscript{2} \quad
  \textbf{Donghun Kang}\textsuperscript{2} \quad
  \textbf{Youngho Cho}\textsuperscript{1} \\
  \textbf{Dongwook Lee}\textsuperscript{1} \quad
  \textbf{Heeseung Kim}\textsuperscript{3,$\dagger$} \quad
  \textbf{Sungroh Yoon}\textsuperscript{1,2,4,$\dagger$}
  \\[0.5em]
    {\normalfont
    \textsuperscript{1}IPAI, Seoul National University (SNU) \quad
    \textsuperscript{2}Department of ECE, SNU} \\
  {\normalfont
    \textsuperscript{3}Department of AI, University of Seoul \quad
    \textsuperscript{4}AIIS, ASRI, INMC, and ISRC, SNU}
}

\begin{document}

\maketitle

\begingroup
  \renewcommand{\thefootnote}{\ensuremath{\dagger}}
  \footnotetext{Corresponding authors.}
\endgroup

\lhead{Preprint. Under review.}

\begin{abstract}
End-to-end full-duplex spoken dialogue models enable natural, real-time spoken interactions with interruptions, backchannels, and overlapping speech.
Prior evaluations have assessed diverse interaction capabilities and identified some degradation patterns across conversational turns.
However, how model behavior evolves over extended sessions remains insufficiently understood.
In this work, we address this gap by systematically characterizing long-session degradation and developing efficient methods to mitigate it.
We introduce a closed-loop evaluation framework for long-session stability and apply it to several publicly available full-duplex speech models, revealing degradation over time in responsiveness, spoken content quality, and voice consistency. To mitigate this degradation, we propose a self-distillation-based training framework for lightweight LoRA adaptation that improves long-session stability with a small amount of additional training.
We further introduce a two-resolution context (TRC) architecture that preserves recent interactions in full detail while compressing older interaction history into key-value representations. We evaluate our methods on three models with distinct architectures, assessing long-session stability, information retention, and computational efficiency. Across all three models, we show that LoRA adaptation mitigates performance degradation over 10-minute sessions, while adding TRC extends these gains to 20-minute sessions (twice the training horizon) and reduces KV-cache growth by more than 90\%. The project page is available online.\footnote{\url{https://kimjunho333.github.io/dmlsd-fdsm/}}

\end{abstract}

\section{Introduction}

Natural spoken conversation involves dynamic coordination between speakers. A listener may respond before an utterance ends, provide backchannels, or interrupt when necessary, rather than waiting for strictly separated turns. Conventional spoken dialogue systems, however, have largely followed a turn-based interaction scheme in which listening and speaking occur in separate stages. To bridge this gap, recent advances in full-duplex spoken dialogue modeling have enabled models to process incoming speech while simultaneously generating spoken responses~\citep{nguyen2023dgslm,defossez2024moshi}. Together, these advances support more fluid interactions that better reflect the dynamics of natural human conversation.

Recent research has increasingly adopted end-to-end approaches to full-duplex spoken dialogue modeling~\citep{defossez2024moshi,roy2026personaplex,cui2026minicpmo,kim2026raonspeech}, with efforts to extend these models beyond natural conversational behaviors to support more sophisticated real-time interaction, including tool use and multimodal perception~\citep{cui2026minicpmo, balam2026nemotronlabs}. These developments reflect a broader shift toward interactive assistants that can continuously perceive their surroundings, respond to users as situations unfold, and provide ongoing support during everyday activities or complex tasks. Realizing this vision requires models to sustain reliable interaction over extended periods, rather than merely performing well in isolated exchanges.

Despite this growing emphasis on continuous interaction, long-session stability has received limited attention in the development of end-to-end full-duplex spoken dialogue models. To the best of our knowledge, no prior work has explicitly targeted long-session stability in end-to-end full-duplex models through dedicated training strategies or architectural designs. Furthermore, although recent benchmarks have evaluated various full-duplex interaction capabilities~\citep{lin2025fullduplexbench, lin2026fdbv15, lin2026fdbv2, lin2026fdbv3, peng2025fdbench, he2026mtrduplexbench, ray2026tauvoice, tang2026instructfd}, the evolution of model behavior and performance over extended, continuous sessions remains insufficiently understood.

Our simulations of extended conversations with representative end-to-end full-duplex models reveal a range of failure modes that emerge as interactions progress. These include diminished responsiveness, the production of semantically incoherent utterances, sometimes degenerating into non-speech vocalizations, and noticeable drift in vocal characteristics. To systematically characterize and quantify these phenomena, we propose a simulation-based diagnostic framework that enables dynamic, closed-loop interactions between the model and a virtual user. This framework enables a quantitative assessment of how conversational performance changes over prolonged interactions, focusing on three key dimensions: responsiveness, spoken content quality, and voice consistency.

To address these degradation patterns, we propose two complementary approaches that improve long-session stability through training and context management. First, we introduce privileged-context self-distillation, a training framework designed to preserve short-session conversational capabilities over extended interactions without requiring an external full-duplex dialogue corpus. By connecting short sessions into long, coherent conversational trajectories, the framework constructs self-distillation targets. 
Second, we introduce a two-resolution context (TRC) architecture, which adapts UniGist, an online context compression method~\citep{deng2025unigist}, to manage growing conversational contexts in streaming full-duplex interaction. In the models we study, dialogue histories accumulate acoustic details and control tokens alongside linguistic content as conversations progress. Some of these details may become less relevant over time. Retaining the entire history at full resolution increases memory costs and may make reliable processing more difficult. TRC therefore preserves recent interactions at full resolution while compressing earlier context into compact KV-cache representations. This design enables memory-efficient streaming inference while remaining well-suited to the proposed training framework.

We evaluate the proposed approaches across three full-duplex models with distinct architectures. Our results show that self-distillation mitigates long-session behavioral degradation within the training horizon. Combining self-distillation with TRC reduces degradation in spoken content and voice consistency by \textbf{71--96\%} relative to the baseline across all three models. Crucially, TRC extends these improvements beyond the training horizon, sustaining coherent responses and voice consistency in 20-minute conversations, whereas self-distillation alone exhibits severe behavioral collapse. Moreover, TRC reduces KV-cache growth by \textbf{91.7--95.8\%}. It also limits the increase in backbone inference latency to \textbf{12--15\%} in models with growing context windows, compared with 141--259\% without TRC. Together, these results demonstrate that the proposed approaches enable existing full-duplex models to sustain longer conversations with improved behavioral stability and computational efficiency.

\section{Related Work}
\subsection{Full-Duplex Speech Modeling}

Spoken dialogue systems are commonly characterized by how they coordinate listening and speaking. \emph{Half-duplex} systems alternate between listening and speaking in separate turns~\citep{fang2025llamaomni, zeng2024glm4voice}. \emph{Full-duplex} systems support concurrent listening and speaking, allowing user input to be processed during an ongoing system response. One approach uses a dedicated turn-taking mechanism to determine whether the system should initiate, continue, or stop speaking. This includes external controllers~\citep{chen2025fireredchat, liao2025flexduo} and predictors operating on the dialogue model's internal representations~\citep{wang2024freezeomni, chen2025minmo}. Another approach integrates turn-taking into the generative process, modeling both speech content and conversational timing~\citep{nguyen2023dgslm, defossez2024moshi, zhang2025omniflatten}.

Within the latter approach, several recent models use temporally structured speech and text histories to determine response timing and produce hidden representations that condition speech generation~\citep{yu2025salmonnomni, roy2026personaplex, cui2026minicpmo, kim2026raonspeech}. As conversations unfold, linguistic, acoustic, and interaction-related information accumulates into long, heterogeneous histories. These histories can become increasingly difficult for the models to process reliably. When such difficulties arise, their effects may extend beyond recall and response content to speech quality and basic responsiveness, including whether and when the model responds. This motivates our focus on the long-session stability of these models and their ability to sustain reliable spoken interaction.

\subsection{Evaluation of Full-Duplex Spoken Dialogue Models}

Existing benchmarks evaluate a broad range of capabilities required for effective full-duplex spoken interaction. Several benchmarks assess conversational timing and dynamics, including turn-taking, interruption handling, backchanneling, and responses to overlapping speech~\citep{arora2025talking, peng2025fdbench, lin2025fullduplexbench, lin2026fdbv15}. Related evaluations also test whether models can adjust their turn-taking behavior to explicit instructions~\citep{tang2026instructfd}. Beyond interaction dynamics, multi-turn evaluations assess models' ability to maintain coherent responses, follow instructions, track and update information, and respond safely across conversational turns~\citep{he2026mtrduplexbench, lin2026fdbv2}. Task-oriented benchmarks further evaluate voice agents' ability to use tools and complete tasks in interactive environments, alongside the quality of their spoken interactions~\citep{lin2026fdbv3, ray2026tauvoice, bogavelli2026evabench, bhosale2026duplexworld}.

Alongside the breadth of capabilities evaluated, a complementary question concerns how performance changes as a session progresses. Prior analyses have examined this question across conversational rounds and elapsed time, revealing degradation in turn-taking and instruction following~\citep{he2026mtrduplexbench, lin2026fdbv2}. We extend this line of analysis to longer, dynamically evolving conversations. We also broaden its scope to examine how degradation emerges and progresses across responsiveness, spoken content quality, and voice consistency.

\subsection{Streaming Long-Context Modeling}

Transformer-based models accumulate a key-value (KV) cache as the context grows, increasing memory and computational costs during inference. KV-cache eviction methods reduce these costs by retaining initial and recent tokens or selecting token states according to attention-based importance~\citep{zhang2023h2o,xiao2024streamingllm,li2024snapkv}. Learned compression methods encode context into compact representations. These include reusable KV prefixes for task instructions~\citep{mu2023gist} and summary vectors or memory slots for longer contexts~\citep{chevalier2023autocompressors,ge2024icae}.

For contexts that grow incrementally, compression can be integrated into sequential processing through progressive activation compression and online memory updates~\citep{zhang2025activationbeacon,kim2024ccm}. UniGist combines a local window of raw tokens with accumulated gist representations, allowing older raw tokens to be discarded as the sequence grows~\citep{deng2025unigist}. These approaches have primarily been studied in text-based language modeling. We build on this line of work to adapt incremental context compression to full-duplex spoken dialogue models that integrate speech generation and conversational timing within a unified generative process.

\section{Diagnosing Long-Session Degradation}
\subsection{Protocol and Measures}
\label{sec:long-session-protocol}
\label{sec:model-and-metrics}
\begin{figure}[t]
\centering
\includegraphics[width=0.9\linewidth]{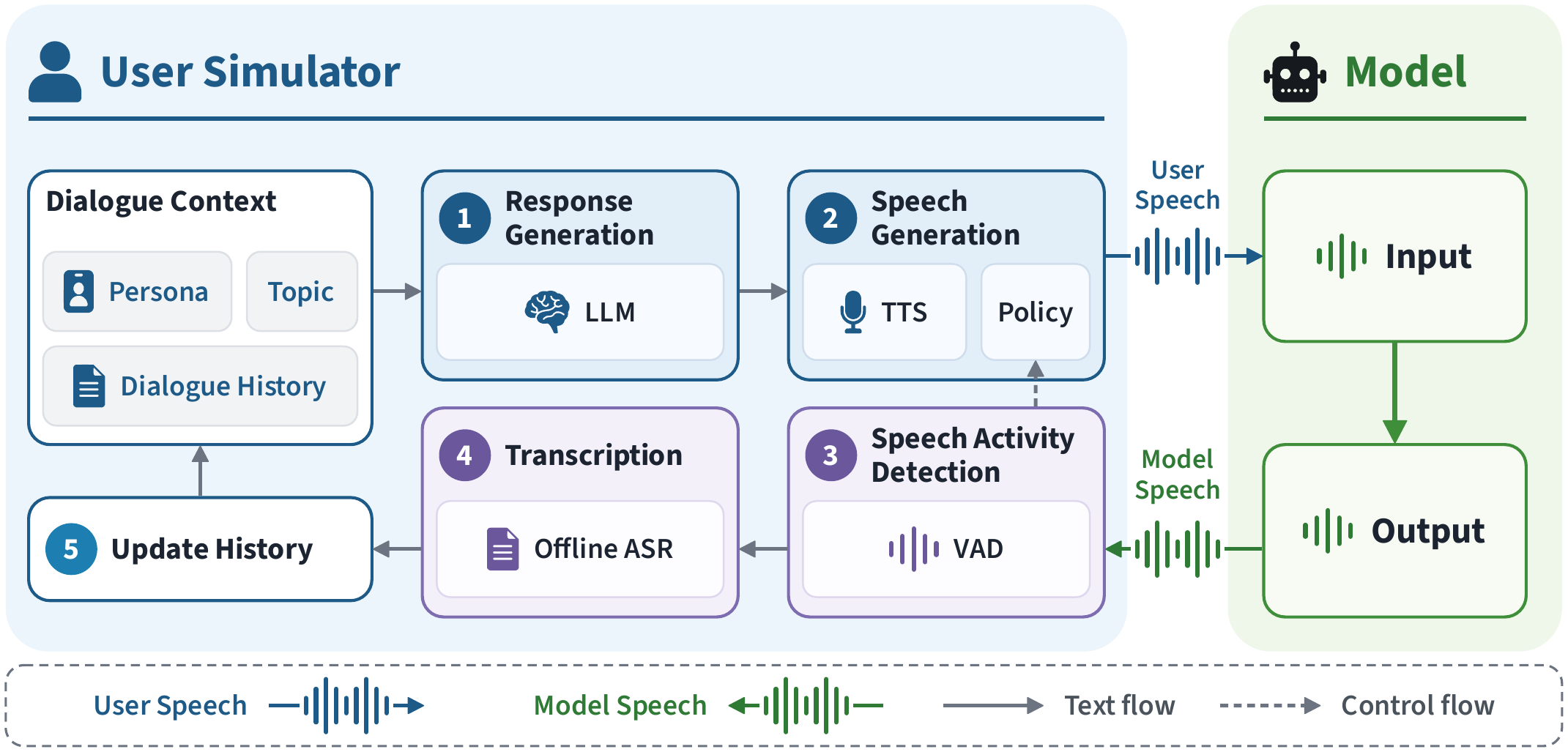}
\caption{\textbf{Closed-loop simulation protocol.}
Overview of the simulation loop, in which model responses are incorporated into the dialogue context to condition subsequent user turns.}
\label{fig:long-session-protocol}
\end{figure}

\paragraph{Protocol.}
We evaluate four end-to-end full-duplex spoken dialogue models: Moshi~\citep{defossez2024moshi}, PersonaPlex~\citep{roy2026personaplex}, MiniCPM-o 4.5~\citep{cui2026minicpmo}, and Raon-SpeechChat~\citep{kim2026raonspeech}. For each model, we simulate 90 English conversations spanning three personas and ten everyday topics, with three episodes per persona--topic pair, each targeting 600\,s. Each conversation follows a closed-loop simulation in which the user's subsequent responses depend on the model's generated speech, allowing the interaction to evolve dynamically rather than following a predefined dialogue script. The user simulator consists of a large language model (LLM), text-to-speech (TTS), automatic speech recognition (ASR), and voice activity detection (VAD) modules, as illustrated in Figure~\ref{fig:long-session-protocol}. Further details on persona and topic instructions, component choices, prompts, and decoding settings are provided in Appendices~\ref{app:long-session-settings} and~\ref{app:long-session-scenarios}.

The simulated user follows a consistent conversational policy across all models, waiting until the model finishes speaking before responding, yielding the floor when interrupted, and avoiding artificially generated backchannels or interruptions. We use logical time as the basis for interaction timing, with time advancing according to the duration of processed audio rather than wall-clock time. All timing-related decisions and measurements are based on this logical time axis. This prevents variations in model inference and system processing latency from affecting conversational timing measurements. If the model does not begin responding within the predefined timeout, the simulator rephrases the user's intent and retries; repeated timeouts terminate the session. Detailed simulation procedures and timing configurations are provided in Appendix~\ref{app:long-session-settings}.

\paragraph{Measures.}
To characterize how model behavior evolves over a session, we evaluate three aspects of full-duplex interaction: responsiveness, spoken content quality, and voice consistency. We divide each session into one-minute intervals of logical time and track six metrics. For responsiveness, \textbf{floor transfer offset (FTO)} measures the temporal gap or overlap between the end of the user's speech and the onset of the model's response during normal turn-taking. \textbf{Response timeout (RT)} rate captures how often the model fails to begin responding within a predefined timeout, while \textbf{terminated sessions (TS)} rate measures the cumulative proportion of sessions forcibly terminated after three consecutive response timeouts. For spoken content quality, we use \textbf{conversation quality (CQ)} and \textbf{word error rate (WER)}. CQ is a 1--5 LLM-based score assessing the contextual appropriateness and coherence of model responses. Its trends are consistent with human ratings (Appendix~\ref{app:cq-human-validation}). WER compares the model's native text output with an ASR transcription of its generated speech to assess how faithfully the generated speech reflects the intended text. For voice consistency, \textbf{speaker similarity (Spk sim)} measures how closely the generated voice matches its first-minute reference using speaker embeddings. Detailed metric definitions and scoring procedures are provided in Appendices~\ref{app:behavior-details}--\ref{app:spk-details}.

For each one-minute interval, we aggregate available observations across sessions using the same criteria for all models. When a session terminates early, observations collected before termination are retained, while subsequent intervals are excluded from all metrics except cumulative TS rate. To ensure reliable estimates, we report metrics only for intervals with sufficient observations, based on predefined minimum sample thresholds. Detailed aggregation rules and reporting thresholds are provided in Appendices~\ref{app:aggregation-details} and~\ref{app:validity-criteria}.

\subsection{Long-Session Degradation Patterns}
\label{sec:long-session-results}
\begin{figure}[tb]
    \centering
        
        \includegraphics[width=0.9\linewidth]{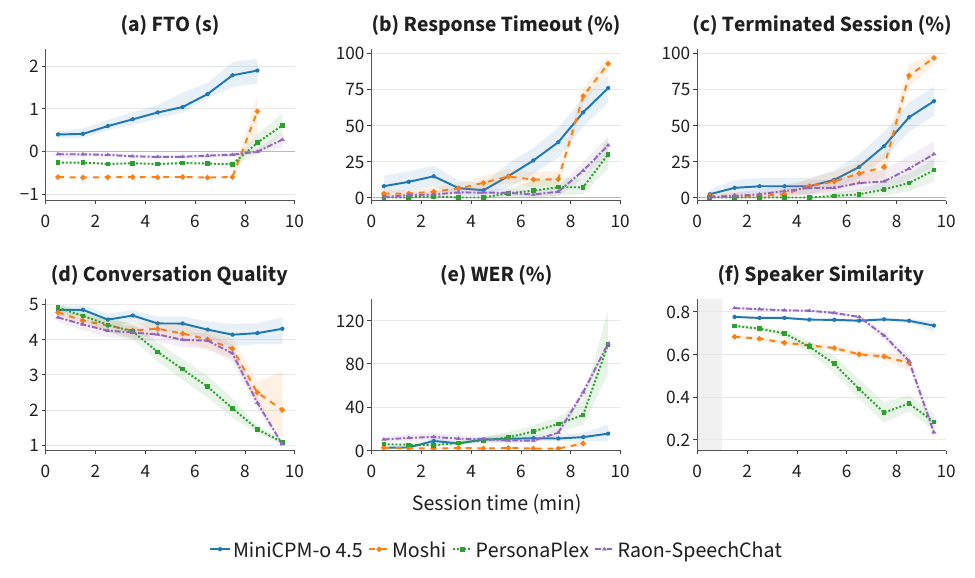}
    
    \caption{
    \textbf{Behavioral degradation over long sessions.}
    Each point summarizes a one-minute interval, plotted at its midpoint.
    The gray band in (f) marks the first minute reserved for speaker enrollment.
    Shaded regions indicate 95\% confidence intervals estimated using 2,000 episode-level cluster bootstrap replicates.
    Detailed numerical results and accompanying statistics
    are reported in Appendix~\ref{app:long-session-results}.
    }
    \label{fig:long-session-results}
\end{figure}

Figure~\ref{fig:long-session-results} reveals distinct patterns of performance degradation across all four models as conversations progress. Moshi exhibits gradual declines in conversation quality and speaker similarity, followed by a sharp deterioration in responsiveness toward the end of the session. Its response timeout rate rises from 12.7\% at 7--8\,min to 70.2\% at 8--9\,min, with the cumulative terminated session rate reaching 96.7\% by 10\,min. PersonaPlex and Raon-SpeechChat instead exhibit more pronounced degradation in spoken content quality and voice consistency, although the timing of these changes differs. For PersonaPlex, degradation develops progressively throughout the conversation, whereas for Raon-SpeechChat it becomes prominent during the final two to three minutes. Their mean conversation quality scores decline from 4.90 to 1.08 and from 4.62 to 1.03, respectively, between the first and final minutes, while final-minute WER reaches 97.9\% and 98.7\%, respectively.

MiniCPM-o 4.5 exhibits a contrasting pattern, with degradation concentrated primarily in responsiveness. Its median FTO increases from 0.39\,s in the first minute to 1.89\,s at 8--9\,min, indicating progressively slower turn-taking. The response timeout rate reaches 75.8\% in the final minute, and the cumulative terminated session rate reaches 66.7\% by 10\,min. Despite this decline in responsiveness, mean conversation quality remains above 4 among sessions with valid observations, and speaker similarity remains relatively stable, although WER rises from 2.4\% to 15.5\%. Taken together, these findings show that long-session stability is a multifaceted challenge, with degradation emerging at different stages and affecting different aspects of conversational behavior across models.
Detailed per-minute results and illustrative failure cases are provided in Appendices~\ref{app:long-session-results} and~\ref{app:qualitative-failures}, respectively.

\section{Methods for Long-Session Stability}
To address the long-session degradation observed in Section~\ref{sec:long-session-results}, we introduce \textbf{privileged-context self-distillation}, which aims to transfer the base model's reliable behavior in short interactions to sustained conversations. Since learning over extended interactions also requires efficiently handling growing dialogue histories, we further introduce a \textbf{two-resolution context (TRC) architecture} that preserves recent interactions while maintaining access to earlier context.

\begin{figure}
\centering
\includegraphics[width=\linewidth]{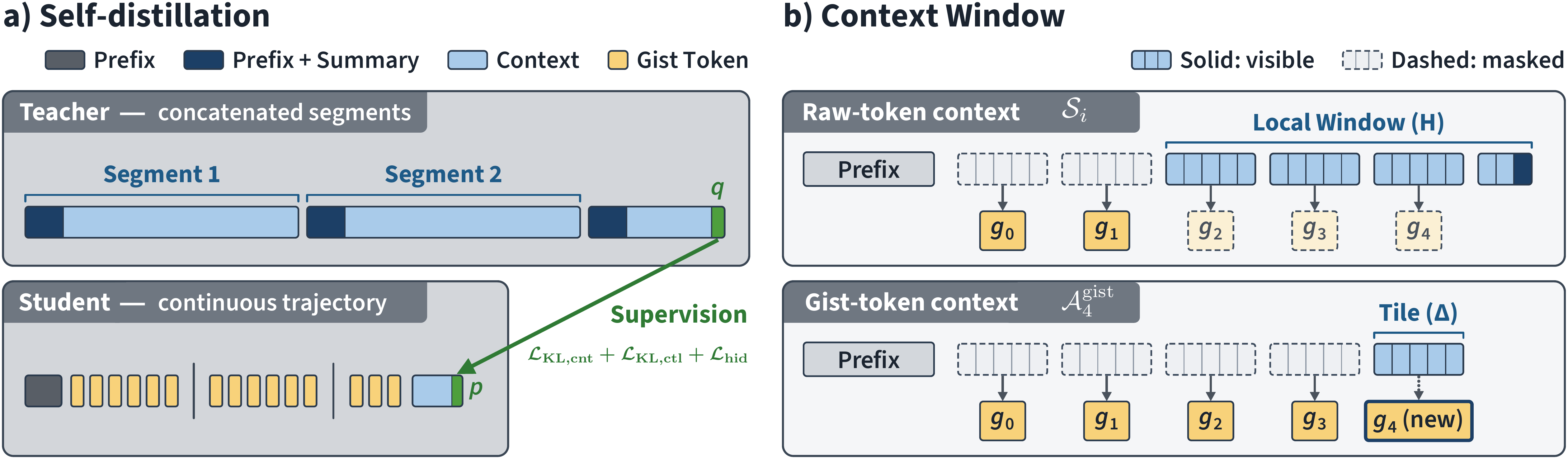}
\caption{\textbf{Privileged-context self-distillation and TRC.} (a) The teacher restarts with a dialogue summary at segment boundaries, while the illustrated TRC student follows the concatenated trajectory continuously. Teacher predictions supervise the student at aligned positions. (b) Attention visibility for raw-token (top) and gist-token (bottom) queries, shown with five tokens per tile and four tiles in the local window. Solid outlines mark visible context; dashed outlines mark masked context. Raw queries attend to the prefix, gists outside the window, and local raw tokens; gist queries attend to the prefix, all preceding gists, and their own tile.}
\label{fig:method-overview}
\end{figure}

\subsection{Privileged-Context Self-Distillation}
\label{sec:self-distillation}
\paragraph{Trajectory construction and distillation setup.}

To enable long-session adaptation without relying on an external full-duplex dialogue corpus, we build on context distillation~\citep{snell2022distilling}, using each frozen base model as its own teacher.
To obtain reliable teacher predictions, we generate dialogues in short segments, avoiding the degradation observed in long sessions.
For each subsequent segment, we append a summary of the preceding conversation to the teacher's system prompt, preserving relevant information from earlier interactions while keeping the teacher's context compact.
We assess the quality of generated segments and regenerate those that fail to meet our quality criteria before concatenating them into continuous training trajectories.

Formally, let $\mathbf{X} = (\mathbf{x}_1,\ldots,\mathbf{x}_K)$ denote a trajectory comprising $K$ dialogue segments, and let $\mathbf{s}_{k-1}$ summarize the preceding segments $\mathbf{x}_{<k}$. We index prediction positions along the concatenated trajectory by $i$ and denote the segment containing position $i$ by $k(i)$. Let $\mathbf{c}_i$ denote the within-segment context at position $i$, including the available user inputs and preceding model outputs. The frozen teacher predicts $q_i = p_{\theta_0}(\cdot \mid \mathbf{s}_{k(i)-1}, \mathbf{c}_i)$. In contrast, the student processes the concatenated segments continuously under teacher forcing, without access to the teacher's summaries (Figure~\ref{fig:method-overview}a). At the corresponding position, it predicts $p_i = p_{\theta}(\cdot \mid \mathbf{m}_i)$, where $\mathbf{m}_i$ denotes its representation of the dialogue history available at that position. Without further context processing, $\mathbf{m}_i$ simply represents the available raw dialogue context.

For each backbone, we construct 100 scenarios by pairing five personas with 20 topics, with both personas and topics distinct from those used in the diagnostic simulation of Section~\ref{sec:long-session-protocol}.
Four trajectories per scenario yield 400 trajectories. We construct each trajectory by concatenating five dialogue segments, targeting a total duration of approximately 10\,min.
Scenario construction, trajectory generation, quality filtering, and prompts are detailed in Appendices~\ref{app:teacher-trajectory} and~\ref{app:method-prompts}.

\paragraph{Training objective.}
Full-duplex interaction requires the student to preserve not only response content but also speaking decisions and the representations used for speech generation. We therefore supervise content and control predictions separately while matching the hidden representations supplied to the audio decoder. We divide eligible prediction positions into \emph{content} and \emph{control} groups based on the teacher-selected token and corresponding speaking state. Content positions correspond to ordinary text predictions during ongoing speech generation, whereas control positions include speech transitions and native interaction-control tokens.
For each group, we minimize the teacher-to-student KL divergence $D_{\mathrm{KL}}(q_i\,\|\,p_i)$, averaging separately over content and control positions to obtain $\mathcal{L}_{\mathrm{KL,cnt}}$ and $\mathcal{L}_{\mathrm{KL,ctl}}$, respectively.
This separation prevents differences in group frequency from determining their relative weights, allowing response content and speaking decisions to be weighted independently. We additionally match the hidden representations supplied to the frozen audio decoder using normalized squared error, denoted by $\mathcal{L}_{\mathrm{hid}}$, to supervise decoder conditioning beyond the token-level KL objectives.

The resulting training objective is
\begin{equation}
\mathcal{L} = \lambda_{\mathrm{cnt}}\mathcal{L}_{\mathrm{KL,cnt}} + \lambda_{\mathrm{ctl}}\mathcal{L}_{\mathrm{KL,ctl}} + \lambda_{\mathrm{hid}}\mathcal{L}_{\mathrm{hid}}.
\label{eq}
\end{equation}
We set all three coefficients to 1 in our experiments. Model-specific control tokens, assignment rules, exact loss definitions, supervision masks, and optimization settings are provided in Appendix~\ref{app:training-details}.
The design choices behind this objective and the corresponding ablations are discussed in Appendix~\ref{app:supervision-ablation}.

\subsection{Two-Resolution Context (TRC) Architecture}
\label{sec:trc}
To support sustained full-duplex interaction, models must preserve fine-grained information from recent speech while efficiently retaining relevant context from earlier exchanges. We therefore introduce a two-resolution context (TRC) architecture by adapting gist-based streaming context compression~\citep{kim2024ccm,deng2025unigist}, enabling incremental compression of dialogue history as streaming inference proceeds.

TRC divides each session into consecutive time intervals of duration $\Delta$, called tiles, and appends a special token, called a gist token, to each tile.
The gist token provides a learned, compact representation of the dialogue within that tile.
Raw tokens from recent tiles remain available within a local window of duration $H$.
Once a tile leaves this window, its raw tokens are discarded and only its gist token is retained.

Formally, at prediction position $i$ along the continuous dialogue trajectory, the student model attends to $\mathcal{S}_i = \mathcal{P} \cup \mathcal{G}_i \cup \mathcal{W}_i$, where $\mathcal{P}$ is a fixed prefix that preserves the prompt and accommodates model-specific attention sinks (Appendix~\ref{app:attention-sink}), $\mathcal{G}_i$ contains gist tokens for tiles that have left the local window, and $\mathcal{W}_i$ contains the available raw context tokens within the window.
The student uses this attention context to construct $\mathbf{m}_i$ and compute the predictive distribution $p_i$ defined in Section~\ref{sec:self-distillation}.
The local window spans duration $H$, chosen as a multiple of $\Delta$, and contains $H/\Delta$ tiles, including the current tile (Figure~\ref{fig:method-overview}b, top).

Each gist token summarizes its corresponding tile while incorporating information from earlier dialogue.
Let $g_j$ denote the gist token for tile $j$, indexed from zero, and let $\mathcal{T}_j$ denote its raw tokens.
When computing $g_j$, its attention context is $\mathcal{A}^{\mathrm{gist}}_j = \mathcal{P} \cup \{g_\ell \mid 0 \leq \ell < j\} \cup \mathcal{T}_j$, comprising the fixed prefix, all preceding gist tokens, and the raw tokens within its own tile (Figure~\ref{fig:method-overview}b, bottom).
Access to preceding gist tokens allows information from earlier dialogue to propagate across tiles even after the corresponding raw tokens are discarded.
Each gist token is initialized from a shared learnable embedding and processed through the backbone layers.
For TRC adaptation, we train only this embedding and LoRA adapters applied to the backbone's Q/K/V/O projections, keeping the audio modules frozen.

TRC is trained using privileged-context self-distillation, allowing the student to match the teacher's predictions using recent uncompressed context and gist memory of earlier dialogue.
Training uses the same attention visibility and positional rules as streaming inference.
At inference time, gist memory grows by one token per tile, while the raw context remains bounded by the local window.
We use approximately $H=10\,\mathrm{s}$ and $\Delta=1\,\mathrm{s}$ for all models, with $H/\Delta=10$ tiles in the local window.
Table~\ref{tab:model-trc-settings} lists the temporal units and exact window and tile sizes for each model.
Further details on positional encoding and layer execution are provided in Appendices~\ref{app:context-computation} and~\ref{app:streaming}.

\section{Experiments}
We evaluate the improvements in long-session stability achieved through privileged-context self-distillation and further examine the role of TRC in sustaining these gains beyond the training horizon. We apply our method to PersonaPlex, MiniCPM-o~4.5, and Raon-SpeechChat, three representative native full-duplex models with distinct architectures and publicly available checkpoints and inference pipelines. We do not separately adapt Moshi because it shares the same backbone architecture as PersonaPlex.

\subsection{Long-Session Stability}
\label{sec:long-session-stability}
Table~\ref{tab:long-session-stability} summarizes responsiveness, spoken content quality, and voice consistency within the ten-minute training horizon and in extended 20-minute sessions, using the same evaluation protocol of Section~\ref{sec:long-session-protocol}. 
We use the same aggregation procedure but report results over two-minute rather than one-minute intervals (Appendix~\ref{app:aggregation-details}).
Ours (LoRA) applies privileged-context self-distillation without context compression, whereas Ours (TRC) combines the same training objective with the TRC architecture.

Within the ten-minute training horizon, privileged-context self-distillation generally improves stability relative to the base models.
For PersonaPlex and Raon-SpeechChat, adaptation substantially reduces the degradation in responsiveness, speech--text consistency, and voice consistency observed in the base models.
Conversation quality also improves considerably: at 8--10 minutes, the base-model scores of 1.29 and 1.63 rise to 3.84 and 4.19 with Ours (TRC) for PersonaPlex and Raon-SpeechChat, respectively. 
For these two models, combining self-distillation with TRC yields larger improvements than LoRA alone on most metrics.
The pattern for MiniCPM-o 4.5 is somewhat more limited. 
Both variants improve late-session performance across metrics relative to the base model, but substantial responsiveness degradation remains despite better preservation of spoken content quality and voice consistency.
For MiniCPM-o 4.5, TRC provides no additional benefit over LoRA alone within the training horizon.

The pattern changes substantially when sessions extend beyond the training range. 
With self-distillation through LoRA alone, all three models deteriorate sharply beyond the training horizon.
Substantial responsiveness problems persist or worsen, while speech--text consistency deteriorates across all three models, with WERs at 18--20 minutes ranging from 88.7\% to 99.6\%.
For Raon-SpeechChat and MiniCPM-o 4.5, conversation quality falls to 1.06 and 1.00, respectively, indicating that the responses themselves collapse. With TRC, however, this degradation beyond the training range is considerably more gradual. 
PersonaPlex remains comparatively stable from 10--12 to 18--20 minutes, retaining a conversation quality score of 3.53 at 18--20 minutes.
For Raon-SpeechChat, conversation quality declines to 3.29 and WER rises to 44.3\% by 18--20 minutes, but these remain better than 1.06 and 99.6\% under LoRA alone. MiniCPM-o 4.5 continues to lose responsiveness gradually, yet the responses it does produce in the remaining sessions remain robust in spoken content quality, with a late conversation quality score of 4.64 and WER of 5.8\%.

Overall, self-distillation improves long-session stability within the training horizon, although the size of the gains varies by model and architecture. Beyond that horizon, TRC retains more of the gains in spoken content quality and voice consistency than LoRA alone. These results indicate greater behavioral stability in 20-minute sessions. Section~\ref{sec:stability-recall} examines how these gains trade off against memory recall, and Section~\ref{sec:ablation} outlines further analyses in the appendix. Per-minute estimates and sample counts for the 20-minute sessions are provided in Appendix~\ref{app:long-session-20m-results}.

\begin{table}[t]
\caption{\textbf{Long-session behavior within and beyond the training horizon.}
Each cell reports early $\rightarrow$ late performance over the indicated
time range. A dash indicates that the sample size does not meet the minimum requirement for reporting specified in Appendix~\ref{app:validity-criteria}.} 
\label{tab:long-session-stability}

\centering
\small
\setlength{\tabcolsep}{3pt}

\begin{tabular*}{\linewidth}{
@{\extracolsep{\fill}}
l
@{\hspace{4pt}}
r@{\,}c@{\,}r
r@{\,}c@{\,}r
r@{\,}c@{\,}r
@{\hspace{7pt}}
c@{\,}c@{\,}c
r@{\,}c@{\,}r
@{\hspace{7pt}}
r@{\,}c@{\,}r
@{}
}
\specialrule{\heavyrulewidth}{0pt}{0pt}

\rowcolor{sectionblue}
\multicolumn{19}{c}{\raisebox{0pt}[\dimexpr\arraystretch\ht\strutbox+1.5pt\relax][\dimexpr\arraystretch\dp\strutbox+1.5pt\relax]{
    \textbf{Within Training Horizon}
    \quad {(\normalfont\small 0--2 $\rightarrow$ 8--10 min)}
}} \\
\specialrule{\lightrulewidth}{0pt}{\aboverulesep}

\multirow{2}{*}{\textbf{Method}}
& \multicolumn{9}{c}{\small\textit{Responsiveness}}
& \multicolumn{6}{c}{\small\textit{Spoken Content}}
& \multicolumn{3}{c}{\small\textit{Voice}}
\\[-2pt]

\cmidrule(lr){2-10}
\cmidrule(lr){11-16}
\cmidrule(lr){17-19}

&
\multicolumn{3}{c}{RT (\%) $\downarrow$}
& \multicolumn{3}{c}{TS (\%) $\downarrow$}
& \multicolumn{3}{c}{FTO (s)}
& \multicolumn{3}{c}{CQ $\uparrow$}
& \multicolumn{3}{c}{WER (\%) $\downarrow$}
& \multicolumn{3}{c}{Spk sim $\uparrow$}
\\
\specialrule{\lightrulewidth}{\aboverulesep}{0pt}

\rowcolor{modelgray}
\multicolumn{19}{@{}l}{\raisebox{0pt}[\dimexpr\arraystretch\ht\strutbox+1.5pt\relax][\dimexpr\arraystretch\dp\strutbox+1.5pt\relax]{\textbf{PersonaPlex}}} \\
\addlinespace[2pt]
\quad Base
& 0.0   & $\rightarrow$ & 17.6
& 0.0   & $\rightarrow$ & 18.9
& $-$0.27 & $\rightarrow$ & 0.40
& 4.78  & $\rightarrow$ & 1.29
& 5.4   & $\rightarrow$ & 44.3
& 0.733 & $\rightarrow$ & 0.337 \\

\quad Ours (LoRA)
& 0.0   & $\rightarrow$ & 3.9
& 0.0   & $\rightarrow$ & 7.8
& $-$0.28 & $\rightarrow$ & $-$0.28
& 4.84  & $\rightarrow$ & 2.98
& 5.2   & $\rightarrow$ & \textbf{8.9}
& 0.722 & $\rightarrow$ & 0.531 \\

\quad Ours (TRC)
& 0.0   & $\rightarrow$ & \textbf{0.4}
& 0.0   & $\rightarrow$ & \textbf{0.0}
& $-$0.24 & $\rightarrow$ & $-$0.23
& 4.76  & $\rightarrow$ & \textbf{3.84}
& 5.2   & $\rightarrow$ & 13.3
& 0.699 & $\rightarrow$ & \textbf{0.617} \\

\addlinespace[3pt]

\rowcolor{modelgray}
\multicolumn{19}{@{}l}{\raisebox{0pt}[\dimexpr\arraystretch\ht\strutbox+1.5pt\relax][\dimexpr\arraystretch\dp\strutbox+1.5pt\relax]{\textbf{MiniCPM-o 4.5}}} \\
\addlinespace[2pt]

\quad Base
& 9.4   & $\rightarrow$ & 66.7
& 6.7   & $\rightarrow$ & 66.7
& 0.40  & $\rightarrow$ & 1.90
& 4.81  & $\rightarrow$ & 4.19
& 2.6   & $\rightarrow$ & 13.6
& 0.775 & $\rightarrow$ & 0.749 \\

\quad Ours (LoRA)
& 6.4   & $\rightarrow$ & \textbf{27.4}
& 4.4   & $\rightarrow$ & \textbf{36.7}
& 0.55  & $\rightarrow$ & 1.80
& 4.84  & $\rightarrow$ & \textbf{4.64}
& 3.4   & $\rightarrow$ & \textbf{4.4}
& 0.778 & $\rightarrow$ & \textbf{0.782} \\

\quad Ours (TRC)
& 15.4  & $\rightarrow$ & 34.6
& 8.9   & $\rightarrow$ & 45.6
& 0.61  & $\rightarrow$ & 1.87
& 4.53  & $\rightarrow$ & 4.35
& 4.4   & $\rightarrow$ & 5.3
& 0.777 & $\rightarrow$ & 0.771 \\

\addlinespace[3pt]

\rowcolor{modelgray}
\multicolumn{19}{@{}l}{\raisebox{0pt}[\dimexpr\arraystretch\ht\strutbox+1.5pt\relax][\dimexpr\arraystretch\dp\strutbox+1.5pt\relax]{\textbf{Raon-SpeechChat}}} \\
\addlinespace[2pt]

\quad Base
& 1.0   & $\rightarrow$ & 27.5
& 1.1   & $\rightarrow$ & 30.0
& $-$0.07 & $\rightarrow$ & 0.06
& 4.52  & $\rightarrow$ & 1.63
& 11.1  & $\rightarrow$ & 70.5
& 0.816 & $\rightarrow$ & 0.520 \\  

\quad Ours (LoRA)
& 0.8   & $\rightarrow$ & 5.4
& 1.1   & $\rightarrow$ & 22.2
& $-$0.10 & $\rightarrow$ & $-$0.08
& 4.50  & $\rightarrow$ & 3.70
& 11.4  & $\rightarrow$ & 22.1
& 0.815 & $\rightarrow$ & 0.768 \\

\quad Ours (TRC)
& 0.0   & $\rightarrow$ & \textbf{1.1}
& 0.0   & $\rightarrow$ & \textbf{3.3}
& 0.02  & $\rightarrow$ & $-$0.04
& 4.54  & $\rightarrow$ & \textbf{4.19}
& 11.7  & $\rightarrow$ & \textbf{15.3}
& 0.814 & $\rightarrow$ & \textbf{0.802} \\

\addlinespace[3pt]
\specialrule{0.8pt}{2pt}{0pt}

\rowcolor{sectionblue}
\multicolumn{19}{c}{\raisebox{0pt}[\dimexpr\arraystretch\ht\strutbox+1.5pt\relax][\dimexpr\arraystretch\dp\strutbox+1.5pt\relax]{
    \textbf{Beyond Training Horizon}
    \quad {(\normalfont\small 10--12 $\rightarrow$ 18--20 min)}
}} \\
\specialrule{\lightrulewidth}{0pt}{\aboverulesep}

\multirow{2}{*}{\textbf{Method}}
& \multicolumn{9}{c}{\small\textit{Responsiveness}}
& \multicolumn{6}{c}{\small\textit{Spoken Content}}
& \multicolumn{3}{c}{\small\textit{Voice}}
\\[-2pt]

\cmidrule(lr){2-10}
\cmidrule(lr){11-16}
\cmidrule(lr){17-19}

&
\multicolumn{3}{c}{RT (\%) $\downarrow$}
& \multicolumn{3}{c}{TS (\%) $\downarrow$}
& \multicolumn{3}{c}{FTO (s)}
& \multicolumn{3}{c}{CQ $\uparrow$}
& \multicolumn{3}{c}{WER (\%) $\downarrow$}
& \multicolumn{3}{c}{Spk sim $\uparrow$}
\\
\specialrule{\lightrulewidth}{\aboverulesep}{0pt}

\rowcolor{modelgray}
\multicolumn{19}{@{}l}{\raisebox{0pt}[\dimexpr\arraystretch\ht\strutbox+1.5pt\relax][\dimexpr\arraystretch\dp\strutbox+1.5pt\relax]{\textbf{PersonaPlex}}} \\
\addlinespace[2pt]

\quad Ours (LoRA)
& 18.5  & $\rightarrow$ & 23.1
& 33.3  & $\rightarrow$ & 87.8
& $-$0.23 & $\rightarrow$ & \NA
& 2.54  & $\rightarrow$ & \NA
& 8.3   & $\rightarrow$ & 98.3
& 0.484 & $\rightarrow$ & 0.158 \\

\quad Ours (TRC)
& 1.2   & $\rightarrow$ & \textbf{1.4}
& 1.1   & $\rightarrow$ & \textbf{4.4}
& $-$0.24 & $\rightarrow$ & $-$0.26
& 3.70  & $\rightarrow$ & \textbf{3.53}
& 14.7  & $\rightarrow$ & \textbf{12.8}
& 0.607 & $\rightarrow$ & \textbf{0.579} \\

\addlinespace[3pt]

\rowcolor{modelgray}
\multicolumn{19}{@{}l}{\raisebox{0pt}[\dimexpr\arraystretch\ht\strutbox+1.5pt\relax][\dimexpr\arraystretch\dp\strutbox+1.5pt\relax]{\textbf{MiniCPM-o 4.5}}} \\
\addlinespace[2pt]

\quad Ours (LoRA)
& 50.3  & $\rightarrow$ & \textbf{10.9}
& 57.8  & $\rightarrow$ & \textbf{64.4}
& 1.97  & $\rightarrow$ & 0.98
& 4.55  & $\rightarrow$ & 1.00
& 7.1   & $\rightarrow$ & 88.7
& 0.777 & $\rightarrow$ & 0.552 \\

\quad Ours (TRC)
& 42.3  & $\rightarrow$ & 51.6
& 54.4  & $\rightarrow$ & 76.7
& 1.72  & $\rightarrow$ & \NA
& 4.51  & $\rightarrow$ & \textbf{4.64}
& 7.8   & $\rightarrow$ & \textbf{5.8}
& 0.774 & $\rightarrow$ & \textbf{0.770} \\

\addlinespace[3pt]

\rowcolor{modelgray}
\multicolumn{19}{@{}l}{\raisebox{0pt}[\dimexpr\arraystretch\ht\strutbox+1.5pt\relax][\dimexpr\arraystretch\dp\strutbox+1.5pt\relax]{\textbf{Raon-SpeechChat}}} \\
\addlinespace[2pt]

\quad Ours (LoRA)
& 13.9  & $\rightarrow$ & 53.4
& 45.6  & $\rightarrow$ & 98.9
& $-$0.10 & $\rightarrow$ & $-$0.25
& 3.06  & $\rightarrow$ & 1.06
& 39.7  & $\rightarrow$ & 99.6
& 0.748 & $\rightarrow$ & 0.148 \\

\quad Ours (TRC)
& 0.8   & $\rightarrow$ & \textbf{1.0}
& 3.3   & $\rightarrow$ & \textbf{3.3}
& $-$0.03 & $\rightarrow$ & $-$0.01
& 4.18  & $\rightarrow$ & \textbf{3.29}
& 20.3  & $\rightarrow$ & \textbf{44.3}
& 0.797 & $\rightarrow$ & \textbf{0.767} \\

\bottomrule
\end{tabular*}
\end{table}

\subsection{Computational Efficiency}

We estimate KV growth under a minimum token-growth assumption to provide a conservative estimate of compression gains.
TRC reduces the prompt-excluded KV growth rate by 91.7--95.8\% (Table~\ref{tab:trc-cost}); for PersonaPlex, this comparison applies before its fixed KV ring saturates.
Latency measurements cover only the main backbone, excluding speech encoders and decoders.
Computing gist tokens introduces additional overhead at short contexts.
For MiniCPM-o 4.5 and Raon-SpeechChat, whose native KV caches are uncapped, the reduced attention cost increasingly offsets this overhead as context grows, resulting in much smaller increases in estimated per-frame mean and p95 latency than the baselines.
PersonaPlex instead bounds its baseline attention cost by simply truncating earlier context through a fixed KV ring, so TRC remains slower at all measured points.
Measurement details are given in Appendix~\ref{app:cost-measurement}.

\begin{table*}[t]
\caption{\textbf{Long-context memory and latency cost.}
Prompt-excluded BF16 KV growth and L40 backbone latency.
P95 range spans the four context checkpoints. The PersonaPlex baseline uses a fixed 4-min KV ring; its KV growth is reported before saturation.}
\label{tab:trc-cost}
\centering
\small
\setlength{\tabcolsep}{4pt}
\newcommand{\costpctrange}[2]{\makebox[2.7em][r]{#1}\,--\,\makebox[2.7em][l]{#2}}
\begin{tabular}{@{}lccccccc@{\hspace{10pt}}c@{}}
\toprule

\multicolumn{1}{c}{%
  \multirow{2}{*}[
    -0.5\dimexpr\aboverulesep+\belowrulesep+\cmidrulewidth
  ]{Model}%
}
&
\multicolumn{1}{c}{%
  \multirow{2}{*}[
    -0.5\dimexpr\aboverulesep+\belowrulesep+\cmidrulewidth
  ]{Variant}%
}

& \multicolumn{2}{c}{KV growth}
& \multicolumn{4}{c}{Mean latency (ms) $\downarrow$}
& P95 (ms) \\
\cmidrule(lr){3-4}
\cmidrule(lr){5-8}
\cmidrule(lr){9-9}
& & MiB/min $\downarrow$ & Saved (\%) $\uparrow$
& 1 min & 5 min & 10 min & 20 min & Range \\
\midrule
\multirow{2}{*}{PersonaPlex} & Base & 375.00 & \multicolumn{1}{c}{--} & \multicolumn{4}{c}{\textbf{26.9}} & \textbf{26.9} \\
 & TRC & 31.25 & \textbf{91.7} & 35.0 & 35.1 & 35.5 & 37.0 & \costpctrange{52.7}{54.9} \\
\arrayrulecolor{black!20}
\midrule[0.2pt]
\arrayrulecolor{black}
\multirow{2}{*}{MiniCPM-o 4.5} & Base & 109.69 & \multicolumn{1}{c}{--} & \textbf{96.0} & 116.4 & 151.3 & 231.0 & \costpctrange{96.1}{231.2} \\
 & TRC & 8.44 & \textbf{92.3} & 113.5 & \textbf{115.5} & \textbf{119.6} & \textbf{127.4} & \textbf{\costpctrange{114.5}{127.5}} \\
\arrayrulecolor{black!20}
\midrule[0.2pt]
\arrayrulecolor{black}
\multirow{2}{*}{Raon-SpeechChat} & Base & 210.94 & \multicolumn{1}{c}{--} & \textbf{29.2} & 44.1 & 64.0 & 104.9 & \costpctrange{29.3}{105.1} \\
 & TRC & 8.79 & \textbf{95.8} & 29.3 & \textbf{30.5} & \textbf{31.5} & \textbf{33.8} & \textbf{\costpctrange{41.0}{41.8}} \\
\bottomrule
\end{tabular}
\end{table*}

\subsection{Stability--Recall Trade-off}
\label{sec:stability-recall}
Beyond behavioral stability, retaining information from earlier in a session is another important aspect of long-session interaction, and we therefore examine how the two relate under TRC.
We evaluate memory recall with a task built from 100 question--answer pairs selected from FictionalQA~\citep{kirchenbauer2026fictionalqa}, in which the model hears a fact at the beginning of a session and is asked about it after 0, 5, or 10 minutes of silence, and we report the percentage of responses that contain the reference answer verbatim (Appendix~\ref{app:memory-recall}).

\begin{figure}[tb]
\centering
\includegraphics[width=0.9\linewidth]{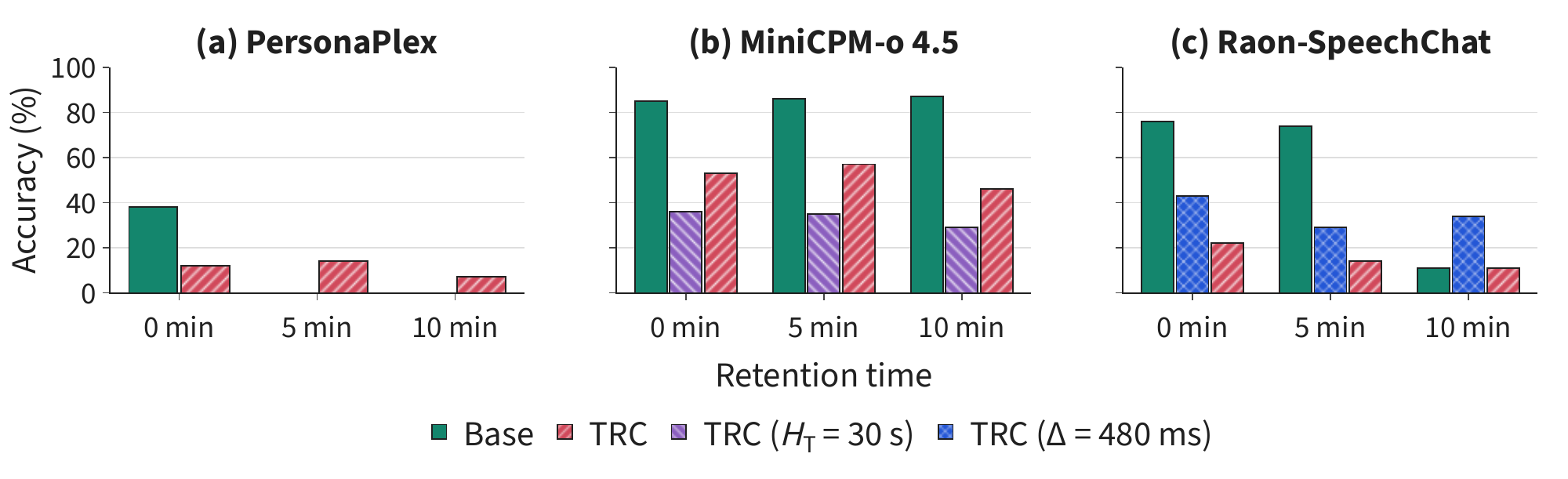}\\[4pt]
\includegraphics[width=0.9\linewidth]{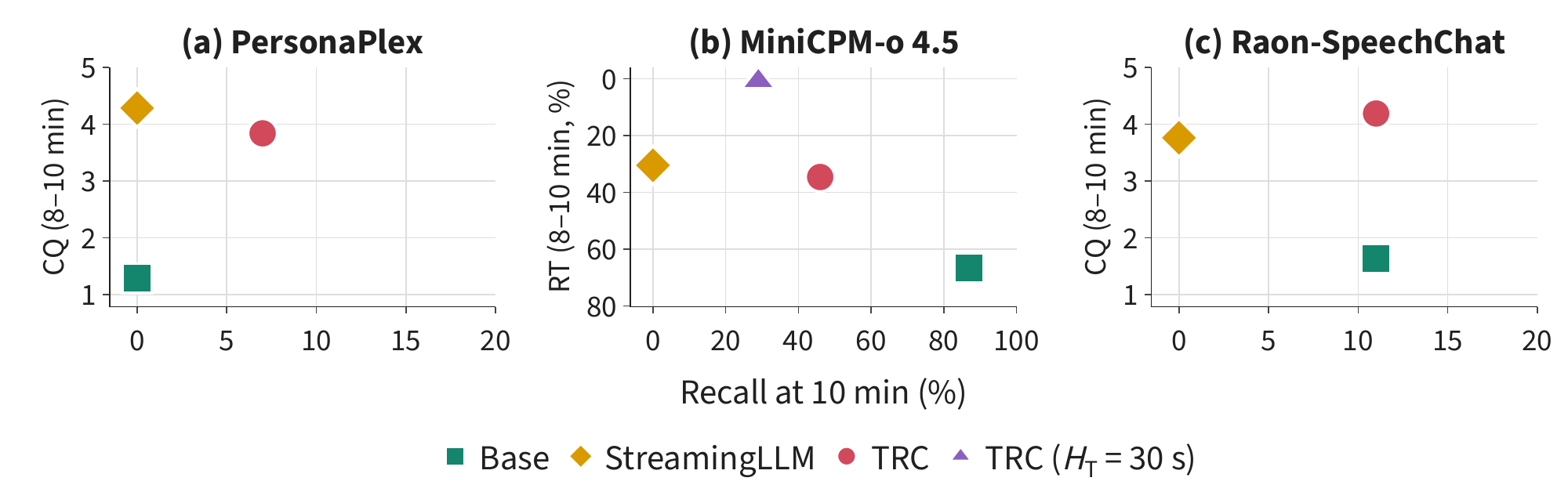}
\caption{\textbf{Memory recall and its trade-off with stability.} Top: recall accuracy against probe distance, with exact values in Table~\ref{tab:single-needle-retention-greedy}. Bottom: recall at 10\,min against stability at 8--10\,min, measured by CQ for PersonaPlex and Raon-SpeechChat and by RT for MiniCPM-o 4.5, with the RT axis reversed so that higher is better.}
\label{fig:memory-recall}
\end{figure}

As Figure~\ref{fig:memory-recall} (top) shows, TRC generally falls short of the short-term recall of Base because it compresses earlier context.
On PersonaPlex and Raon-SpeechChat, the recall of Base itself decays over the session, vanishing within 5 minutes and falling to the level of TRC by 10 minutes, respectively, whereas the recall of TRC varies less.
On MiniCPM-o 4.5, however, Base retains its recall throughout (85--87\%), so the stability gained with TRC comes at a clear cost in recall.
Two additional configurations show that the recall of TRC depends on its configuration.
On Raon-SpeechChat, halving the tile duration to $\Delta=480$\,ms doubles the number of gist tokens and thus lowers the compression rate, but raises recall to 43, 29, and 34\% at 0, 5, and 10 minutes.
On MiniCPM-o 4.5, reducing the maximum raw context of the teacher, $H_{\mathrm{T}}$, from 150\,s to 30\,s, with earlier dialogue given only as a summary, lowers recall to 36, 35, and 29\% but improves long-session stability (Appendix~\ref{app:stability-recall}).
Recall can thus be traded off against the compression rate as well as against stability.

Figure~\ref{fig:memory-recall} (bottom) examines the latter trade-off at the 10-minute mark, pairing recall after 10 minutes with stability at 8--10\,min, measured by the metric of the dominant degradation of each model identified in Section~\ref{sec:long-session-results}.
It also includes StreamingLLM with the TRC prefix (Appendix~\ref{app:local-context-ablation}), which evicts earlier context and thus marks the end of the trade-off opposite to the full context of Base.
On PersonaPlex and Raon-SpeechChat, where the recall of Base decays, TRC matches or exceeds Base in recall with far higher stability, and it retains part of the recall that eviction loses at a comparable level of stability.
On MiniCPM-o 4.5, the default configuration lies between the two ends, and TRC ($H_{\mathrm{T}}=30$\,s) is more stable than eviction while retaining part of the earlier information.
TRC thus reaches points of this trade-off that neither eviction nor the full context reaches, improving its Pareto frontier. Its recall nevertheless remains well below that of Base on MiniCPM-o 4.5, and pushing this frontier further, for example by adding explicit memory objectives to training, remains for future work.

\subsection{Further Analysis}
\label{sec:ablation}
We further examine our results with three additional analyses in the appendix. Ablations that restrict each model to a local context identify what is essential to the mitigation and what role TRC plays in it (Appendix~\ref{app:local-context-ablation}). An evaluation with GPT-Live-1, a commercial real-time full-duplex model, as the user shows that our findings largely hold when the user backchannels and interrupts (Appendix~\ref{app:gpt-live}). Finally, we examine how the adaptation affects full-duplex behaviors on Full-Duplex-Bench (Appendix~\ref{app:fdb}).

\section{Conclusion}

In this paper, we investigated how the capabilities of full-duplex spoken dialogue models evolve over extended sessions and proposed methods to mitigate their degradation. Using a closed-loop evaluation framework, we tracked changes in responsiveness, spoken content quality, and voice consistency, revealing that all four evaluated models exhibited degradation in at least one dimension, with distinct patterns and onset times. To address this issue, we introduced privileged-context self-distillation and two-resolution context (TRC) management. Across three model types, self-distillation improved stability within the training horizon, while combining it with TRC better preserved spoken content quality and voice consistency in 20-minute sessions beyond the 10-minute training horizon and substantially reduced KV-cache growth. These findings highlight the importance of tracking and addressing long-session degradation in full-duplex spoken dialogue and demonstrate the potential of leveraging reliable short-session behavior together with efficient context management to support more robust extended interactions.

\label{page:main-end}

\newpage
\subsection*{AI use statement}
We used LLMs in the experimental pipeline to generate simulated user utterances and summaries for teacher trajectories and to evaluate conversation quality, as described in the methods. We did not use LLMs for any other data generation, processing, or evaluation. 

Generative AI tools also assisted with writing and editing research code and scripts; finding, summarizing, and translating relevant literature; translating and editing manuscript text; refining and editing figures; and drafting portions of the paper. These tools did not materially shape the research ideas or methodology and were not used to design experiments or interpret results.

The authors reviewed AI-assisted material before its inclusion, checked suggested references against the original sources, and jointly reviewed and revised translated manuscript text. We take responsibility for the final content of this paper, including all AI-assisted text, code, and experimental artifacts.

\clearpage
\bibliography{iclr2027_conference}
\bibliographystyle{iclr2027_conference}

\clearpage
\appendix
\begingroup
\DCLAppendixFloatSetup
\section{Long-Session Simulation}
\subsection{Models and Simulation Settings}
\label{app:long-session-settings}

\paragraph{Baseline models and inference.}
Table~\ref{tab:long-session-checkpoints} identifies the released checkpoint used for each baseline.
All four models run in bfloat16 and exchange audio with the simulated user.
Table~\ref{tab:long-session-decoding} summarizes the native context policy, processing step, audio rates, and decoding settings.
We further specify several model-specific generation settings.
MiniCPM-o 4.5 generates at most 20 new speaking-text tokens per chunk.
Its text repetition penalty is 1.05 over a 512-token window, and its listening-probability scale is 1.0.
For Raon-SpeechChat, the silence, backchannel, and end-of-speech penalties are set to zero.
Model-initiated opening is also disabled.
The fixed text prompts and voice-conditioning inputs for each model are listed in Table~\ref{tab:long-session-conditioning}.

\begin{table}[tp]
\caption{Released checkpoints used in the long-session simulation.}
\label{tab:long-session-checkpoints}

\centering\footnotesize
\begin{tabular}{@{}ll@{}}
\toprule
Model & Checkpoint \\
\midrule
Moshi & \texttt{\detokenize{kyutai/moshiko-pytorch-bf16}} \\
PersonaPlex & \texttt{\detokenize{nvidia/personaplex-7b-v1}} \\
MiniCPM-o 4.5 & \texttt{\detokenize{openbmb/MiniCPM-o-4_5}} \\
Raon-SpeechChat & \texttt{\detokenize{KRAFTON/Raon-SpeechChat-9B}} \\
\bottomrule
\end{tabular}

\end{table}

\begin{table}[tp]
\centering\footnotesize
\setlength{\tabcolsep}{3pt}
\caption{\textbf{Baseline configuration and decoding settings.} Step denotes the duration of input audio processed per inference step. Audio rates are input/output. Decoding entries give temperature, top-$k$, and top-$p$; a dash indicates no separate audio-decoding setting.}
\label{tab:long-session-decoding}
\begin{tabular}{@{}l l c c c c@{}}
\toprule
Model & KV Context & Step & I/O (kHz) & Text decoding & Audio decoding \\
\midrule
Moshi       & 240\,s sliding & 80\,ms & 24 / 24 & 0.7 / 25 / -- & 0.8 / 250 / -- \\
PersonaPlex & 240\,s sliding & 80\,ms & 24 / 24 & 0.7 / 25 / -- & 0.8 / 250 / -- \\
MiniCPM-o 4.5 & accumulated & 1\,s & 16 / 24 & 0.7 / 100 / 0.8 & -- \\
Raon-SpeechChat & accumulated & 80\,ms & 24 / 24 & 0.9 / 66 / 0.95 & -- \\
\bottomrule
\end{tabular}
\end{table}

\begin{table}[tp]
\caption{\textbf{Text and voice conditioning for baseline models.} A dash indicates that no explicit input of that type is supplied.}
\label{tab:long-session-conditioning}

\centering\footnotesize
\setlength{\tabcolsep}{3pt}
\begin{tabular}{@{}M{0.18\linewidth}M{0.40\linewidth}M{0.25\linewidth}@{}}
\toprule
Model & Text prompt & Voice conditioning \\
\midrule
Moshi & -- & -- \\
\arrayrulecolor{black!10}
\midrule[0.2pt]
\arrayrulecolor{black}
PersonaPlex & ``You enjoy having a good conversation.'' & \texttt{NATF2.pt} \\
\arrayrulecolor{black!10}
\midrule[0.2pt]
\arrayrulecolor{black}
MiniCPM-o 4.5 & ``Have a natural spoken conversation with the user in English.'' & Kokoro-82M reference audio (\texttt{af\_heart}) \\
\arrayrulecolor{black!10}
\midrule[0.2pt]
\arrayrulecolor{black}
Raon-SpeechChat & ``You are engaging in real-time conversation. Have a natural spoken conversation with the user in English.'' & -- \\
\bottomrule
\end{tabular}

\end{table}

\paragraph{Simulator and evaluation components.}
Table~\ref{tab:long-session-analysis} lists the components and their fixed settings.
The user LLM receives persona and topic instructions, delivered user text, ASR transcriptions of completed model turns, and the requested action (opening, response, or retry).
The evaluated model's identity and system prompt are withheld from both the user LLM and the CQ judge.
The persona, topic, and utterance-generation instructions appear in Appendix~\ref{app:long-session-scenarios}.

\begin{table}[tp]
\caption{User-simulation and evaluation components.}
\label{tab:long-session-analysis}

\centering\footnotesize
\setlength{\tabcolsep}{3pt}
\renewcommand{\tabularxcolumn}[1]{m{#1}}
\begin{tabularx}{\linewidth}{@{}M{0.17\linewidth}M{0.25\linewidth}>{\raggedright\arraybackslash}X@{}}
\toprule
Component & Model & Configuration \\
\midrule
CQ judge & gemini-3.8-flash & Gemini Batch API; low thinking; $T=0$; maximum 10,000 output tokens per 10-min episode (20,000 for 20-min) \\
\arrayrulecolor{black!10}
\midrule[0.2pt]
\arrayrulecolor{black}
User LLM & Qwen3.8-27B & Thinking disabled; $T=0.7$, top-$p=0.8$, top-$k=20$; presence/repetition penalties 1.5/1.0; maximum 128 output tokens per utterance. \\
\arrayrulecolor{black!10}
\midrule[0.2pt]
\arrayrulecolor{black}
User TTS & Kokoro-82M & \texttt{am\_fenrir}; speed 1.0; 24\,kHz mono, resampled to model input rate. \\
\arrayrulecolor{black!10}
\midrule[0.2pt]
\arrayrulecolor{black}
VAD & Silero 6.2.1 & 16\,kHz; threshold 0.5; 32\,ms frames; minimum silence 300\,ms; boundary padding 30\,ms. \\
\arrayrulecolor{black!10}
\midrule[0.2pt]
\arrayrulecolor{black}
ASR & Parakeet Unified EN 0.6B & English; 16\,kHz; float32; word timestamps. \\
\arrayrulecolor{black!10}
\midrule[0.2pt]
\arrayrulecolor{black}
Speaker encoder & spkrec-ecapa-voxceleb & 16\,kHz mono. \\
\bottomrule
\end{tabularx}

\end{table}

\paragraph{Interaction policy.}
The interaction clock advances with processed input audio, including silence, independently of wall-clock execution.
We classify model utterances as turn-taking, backchannels, or interruptions according to their timing relative to user speech, as defined in Table~\ref{tab:long-session-utterance-types}. Table~\ref{tab:long-session-timing} specifies how the simulator coordinates user turns, handles response timeouts, and determines session completion.

\begin{table}[tp]
\caption{Operational classification of model utterances}
\label{tab:long-session-utterance-types}

\centering\small
\setlength{\tabcolsep}{6pt}

\begin{tabular}{@{}l >{\raggedright\arraybackslash}m{0.60\linewidth}@{}}
\toprule
Utterance type & Classification criterion \\
\midrule
Turn-taking & Model speech starts after the user finishes or overlaps with the user's natural turn ending without triggering an interruption. \\
\arrayrulecolor{black!10}
\midrule[0.2pt]
\arrayrulecolor{black}
Interruption & Accumulated model speech activity during overlap reaches 2\,s while the user is still speaking. \\
\arrayrulecolor{black!10}
\midrule[0.2pt]
\arrayrulecolor{black}
Backchannel & Model speech begins and ends during the user's utterance, below the interruption threshold. \\
\bottomrule
\end{tabular}

\end{table}

\begin{table}[tp]
\caption{Operational rules for long-session simulation.}
\label{tab:long-session-timing}

\centering\small
\renewcommand{\tabularxcolumn}[1]{m{#1}}
\begin{tabularx}{\linewidth}{@{}M{0.21\linewidth}Y@{}}
\specialrule{\heavyrulewidth}{0pt}{0pt}

\rowcolor{black!7}
\multicolumn{2}{c}{\raisebox{0pt}[\dimexpr\arraystretch\ht\strutbox+1.5pt\relax][\dimexpr\arraystretch\dp\strutbox+1.5pt\relax]{\textbf{Turn coordination}}} \\
\specialrule{\lightrulewidth}{0pt}{\aboverulesep}
Rule & Specification \\
\midrule
Opening &
If the model has not spoken, the user starts after 2\,s. \\
\arrayrulecolor{black!10}
\midrule[0.2pt]
\arrayrulecolor{black}
Turn boundary &
Close a model turn after 800\,ms of silence, grouping speech segments across shorter pauses. The user simulator then responds. \\
\arrayrulecolor{black!10}
\midrule[0.2pt]
\arrayrulecolor{black}
Interruption handling &
On a model interruption, discard remaining user audio from the next input step. \\

\addlinespace[5pt]
\specialrule{\lightrulewidth}{0pt}{0pt}
\rowcolor{black!7}
\multicolumn{2}{c}{\raisebox{0pt}[\dimexpr\arraystretch\ht\strutbox+1.5pt\relax][\dimexpr\arraystretch\dp\strutbox+1.5pt\relax]{\textbf{Response waiting and recovery}}} \\
\specialrule{\lightrulewidth}{0pt}{\aboverulesep}
Rule & Specification \\
\midrule
Waiting onset &
Begin waiting after both the user utterance and its associated backchannel have ended and overlap classification is resolved. \\
\arrayrulecolor{black!10}
\midrule[0.2pt]
\arrayrulecolor{black}
Timeout and retry &
After \(3\,\mathrm{s}\) without a non-backchannel response, it is regarded as a response timeout, and then the system rephrases the same intent and retries.\\
\arrayrulecolor{black!10}
\midrule[0.2pt]
\arrayrulecolor{black}
Repeated timeouts &
Terminate the session after three consecutive timeouts. \\

\addlinespace[5pt]
\specialrule{\lightrulewidth}{0pt}{0pt}
\rowcolor{black!7}
\multicolumn{2}{c}{\raisebox{0pt}[\dimexpr\arraystretch\ht\strutbox+1.5pt\relax][\dimexpr\arraystretch\dp\strutbox+1.5pt\relax]{\textbf{Session boundaries}}} \\
\specialrule{\lightrulewidth}{0pt}{\aboverulesep}
Rule & Specification \\
\midrule
Target duration &
Stop starting new user turns and retries at 600\,s. \\
\arrayrulecolor{black!10}
\midrule[0.2pt]
\arrayrulecolor{black}
Completion allowance &
Allow pending interactions to finish within another 15\,s. \\
\addlinespace[3pt]

\bottomrule
\end{tabularx}
\end{table}

\subsection{User-Simulator Instructions}
\label{app:long-session-scenarios}

Each episode combines one persona and one topic from Tables~\ref{tab:simulator-personas} and~\ref{tab:simulator-topics}, with three repetitions per combination. The user LLM follows the shared system prompt shown in Figure~\ref{fig:user-simulator-system-prompt}. For each utterance, it receives a separate user message (Figure~\ref{fig:user-simulator-user-message}) containing the persona and topic instructions, dialogue history, and a turn-specific request summarized in Table~\ref{tab:simulator-actions}. For interrupted user turns, the dialogue history retains only the text delivered before cutoff.

\begin{table}[tp]
\centering
\footnotesize
\caption{User-simulator persona instructions (verbatim).}
\label{tab:simulator-personas}
\begin{tabular}{M{0.19\linewidth}M{0.73\linewidth}}
\toprule
Persona & Instruction \\
\midrule
User-led & You naturally take the lead in the substance of the conversation. Share plausible everyday thoughts, preferences, plans, and experiences with enough detail to give the other speaker something meaningful to respond to. Ask a relevant question when useful, but do not turn every utterance into an interview or dominate the floor. \\
\arrayrulecolor{black!10}
\midrule[0.2pt]
\arrayrulecolor{black}
Model-led & You are curious about the other speaker's general perspective. Often ask for explanations, preferences, comparisons, or hypothetical choices, and respond to the answers with your own brief reaction or context. Never assume that the other speaker has a body, possessions, recent real-world experiences, or personal memories. \\
\arrayrulecolor{black!10}
\midrule[0.2pt]
\arrayrulecolor{black}
Balanced & Balance sharing plausible everyday thoughts, preferences, plans, and experiences with asking for the other speaker's general perspective. Let either speaker develop an idea, react to what was actually said, and vary naturally between statements and questions. \\
\bottomrule
\end{tabular}
\end{table}

\subsection{Metric Definitions and Aggregation}
\label{app:aggregation-details}

The metrics used in Sections~\ref{sec:model-and-metrics} and \ref{sec:long-session-stability} are evaluated over temporal intervals within each episode $e \in \mathcal{E}$, where $\mathcal{E}$ denotes the set of evaluation episodes.
The following subsections detail the measurement and aggregation procedures for each metric.
Only measurements meeting the minimum data requirements are included in aggregation.
Appendix~\ref{app:validity-criteria} specifies these requirements, the minimum sample sizes for reporting aggregated estimates, and the procedure for computing bootstrap confidence intervals.

\subsubsection{Responsiveness}
\label{app:behavior-details}

\paragraph{Floor transfer offset (FTO).}
We define the FTO for each normal turn-taking response as
the time difference between the model's response onset and
the end of the corresponding user utterance,
$f_{e,i} = t^{\mathrm{M,start}}_{e,i}
- t^{\mathrm{U,end}}_{e,i}$.
For each reporting interval $I$, we report the median FTO
across all eligible responses pooled across episodes:
\begin{equation}
\mathrm{FTO}(I)
=
\operatorname{median}
\left\{
f_{e,i}
\;\middle|\;
e \in \mathcal{E},\;
i \in \mathcal{T}_e(I)
\right\}.
\end{equation}
Here, $\mathcal T_e(I)$ denotes the set of normal turn-taking
responses in episode $e$ whose preceding user utterances
end within $I$.
Negative FTO values indicate overlapping speech.
Backchannels, interruptions, and non-responses are excluded.

\clearpage
\begingroup
\footnotesize
\renewcommand{\tabularxcolumn}[1]{m{#1}}
\begin{longtable}{M{0.19\linewidth}M{0.73\linewidth}}
\caption{User-simulator topic instructions (verbatim).}\label{tab:simulator-topics}\\
\toprule
Topic & Instruction \\
\midrule
\endfirsthead
\multicolumn{2}{l}{\tablename~\thetable\ (continued)}\\
\toprule
Topic & Instruction \\
\midrule
\endhead
\bottomrule
\endfoot
Room redecorating & Have a relaxed conversation about planning to refresh a room at home. Discuss the desired atmosphere and practical choices such as furniture placement, colors, lighting, comfort, budget, and small changes worth trying first. Do not assume the other speaker owns or has seen a room. \\
\arrayrulecolor{black!10}
\midrule[0.2pt]
\arrayrulecolor{black}
Home cooking & Have a relaxed conversation about trying a new kind of home cooking. Explore what would be enjoyable to make, preferred flavors, confidence level, available time, simple preparation ideas, and how to make the experience less stressful. General or hypothetical preferences are fine. \\
\arrayrulecolor{black!10}
\midrule[0.2pt]
\arrayrulecolor{black}
Home plants & Have a relaxed conversation about starting to keep plants at home or making a small garden. Discuss the space and light available, easy first plants, routines, expectations, common worries, and what would make the project satisfying without assuming the other speaker grows plants. \\
\arrayrulecolor{black!10}
\midrule[0.2pt]
\arrayrulecolor{black}
Weekend hobby & Have a relaxed conversation about finding a hobby that could be continued on weekends. Explore desired pace, indoor or outdoor options, social or solitary activities, cost and learning effort, ways to begin, and what would make the hobby sustainable. \\
\arrayrulecolor{black!10}
\midrule[0.2pt]
\arrayrulecolor{black}
Daily routines & Have a relaxed conversation about making morning or evening routines more comfortable and useful. Discuss current friction, energy, time, small habits, realistic tradeoffs, and ways to test changes without making the routine rigid. \\
\arrayrulecolor{black!10}
\midrule[0.2pt]
\arrayrulecolor{black}
Decluttering & Have a relaxed conversation about organizing a cluttered space and deciding what to keep. Explore which area to start with, practical sorting rules, sentimental items, motivation, storage, and how to avoid the space becoming cluttered again. \\
\arrayrulecolor{black!10}
\midrule[0.2pt]
\arrayrulecolor{black}
Nearby day trip & Have a relaxed conversation about planning a nearby day trip. Discuss the preferred pace and setting, travel time, food, weather contingencies, rough scheduling, and how to leave room for spontaneity. Avoid relying on current facts about a named destination. \\
\arrayrulecolor{black!10}
\midrule[0.2pt]
\arrayrulecolor{black}
Story preferences & Have a relaxed conversation about what makes stories enjoyable. Discuss general preferences such as mood, pacing, characters, settings, endings, familiarity versus surprise, and how preferences change by situation. Do not require knowledge of any named book, film, or current release. \\
\arrayrulecolor{black!10}
\midrule[0.2pt]
\arrayrulecolor{black}
Practical skill & Have a relaxed conversation about choosing and learning a useful everyday skill. Explore why it would help, possible skill choices, how to practice, likely early difficulties, ways to notice progress, and how to keep the learning plan realistic. \\
\arrayrulecolor{black!10}
\midrule[0.2pt]
\arrayrulecolor{black}
Casual gathering & Have a relaxed conversation about planning a casual gathering with a few close people. Discuss the place, simple food or activities, atmosphere, timing, different comfort levels, and ways for everyone to participate without making the occasion overly formal. \\
\end{longtable}
\endgroup

\FloatBarrier

\begin{table}[tp]
\centering
\footnotesize
\caption{\textbf{User-simulator generation requests.} For each request type, the table lists the action type and summarizes the content goal.}
\label{tab:simulator-actions}
\begin{tabular}{@{}M{0.19\linewidth}M{0.15\linewidth}M{0.50\linewidth}@{}}
\toprule
Request type & Action type & Content goal \\
\midrule
Opening & Normal & Naturally introduce the configured topic and invite a response after the initial listening period if the model has not spoken. \\
\arrayrulecolor{black!10}
\midrule[0.2pt]
\arrayrulecolor{black}
Model-first bridge & Normal & Respond to the model's opening and smoothly guide the exchange toward the configured topic. \\
\arrayrulecolor{black!10}
\midrule[0.2pt]
\arrayrulecolor{black}
Normal response & Normal & Respond to the latest audible model turn; if the topic is exhausted or the model winds down, introduce a related or fresh topic to sustain the conversation. \\
\arrayrulecolor{black!10}
\midrule[0.2pt]
\arrayrulecolor{black}
Retry & Retry & Preserve the preceding user utterance's intent in a different natural sentence, without changing topic or mentioning the missing response. \\
\bottomrule
\end{tabular}
\par\vspace{6pt}
\end{table}

\begin{figure}[tp]
\centering
\begin{DCLPromptBox}[unbreakable,
  title after break={},
  listing options={
    style=DCLPromptText,
    basicstyle=\footnotesize\rmfamily,
    breakautoindent=false,
    breakindent=0pt
  },
  top=1mm,
  bottom=1mm
]{User simulator: system prompt example}
You are simulating a human participant in a spoken conversation.
Generate only the requested utterance text.

Rules:
- Follow the action_type and content_goal in generation_request exactly.
- Remain consistent with both speakers and the audible conversation.
- Do not end the conversation unless explicitly requested.
- Output one concise, natural English spoken utterance.
- Do not output speaker labels, quotation marks, action descriptions, or meta commentary.

Normal-response rules:
- Respond directly to the latest audible turn and remain consistent with both speakers.
- Add a relevant opinion, experience, or detail when appropriate.
- Use at most one natural follow-up question, and only when it helps the conversation.
- You may move gradually to a related topic, but do not change topics abruptly.
- Do not repeat questions or information already present in the dialogue.
- Continue the configured scenario naturally.
- If the current topic feels exhausted, or your own utterance is turning into a vague or repetitive affirmation, pivot to a new related or fresh topic instead of continuing to circle the same point.
- If the other speaker's turn signals winding down or closing the conversation, bring up a new topic and keep the conversation going.
\end{DCLPromptBox}
\caption{System prompt example for user utterance generation.}
\label{fig:user-simulator-system-prompt}
\end{figure}

\begin{figure}[tp]
\centering
\begin{DCLTextPromptBox}{User simulator: user message example}
\textless{}persona\textgreater{}\par
\{persona instruction\}\par
\textless{}/persona\textgreater{}\par
\vspace{\baselineskip}
\textless{}conversation\_concept\textgreater{}\par
\{topic instruction\}\par
\textless{}/conversation\_concept\textgreater{}\par
\vspace{\baselineskip}
\textless{}dialogue\_history\textgreater{}\par
\{dialogue history\}\par
\textless{}/dialogue\_history\textgreater{}\par
\vspace{\baselineskip}
\textless{}generation\_request\textgreater{}\par
action\_type: normal\_response\par
content\_goal: Respond directly to the model's latest audible utterance and continue the conversation naturally. If the current topic feels exhausted, or the model's turn signals winding down or closing the conversation, do not follow along or drift into vague affirmations --- bring up a new related or fresh topic and keep the exchange substantive.\par
\textless{}/generation\_request\textgreater{}\par
\end{DCLTextPromptBox}
\caption{\textbf{User message example for a normal response.} Braced placeholders indicate the selected persona and topic instructions and the current dialogue history.}
\label{fig:user-simulator-user-message}
\end{figure}

\begin{table}[tp]
\caption{Aggregation and reporting thresholds.}
\label{tab:metric-validity}

\centering\small
\begin{tabular}{ccc}
\toprule
Metric & Aggregation requirement & Reporting threshold \\
\midrule
FTO & --- & 20 transfers \\
RT & --- & 1 resolved opportunity \\
TS & --- & --- \\
CQ & 30\,s observed; model speech present & 10 episodes \\
WER & 25 reference words & 10 episodes \\
Speaker similarity & 10\,s reference speech; 5\,s scored speech & 10 episodes \\
\bottomrule
\end{tabular}

\par\vspace{2pt}
\begin{minipage}{\linewidth}
\end{minipage}
\end{table}

\begin{table}[tp]
\caption{Conversation-quality scoring rubric.}
\label{tab:cq-rubric}
\centering\small
\renewcommand{\arraystretch}{1.12}
\renewcommand{\tabularxcolumn}[1]{m{#1}}
\begin{tabularx}{\linewidth}{@{}
    >{\centering\arraybackslash}m{0.06\linewidth}
    M{0.19\linewidth}
    Y@{}
}
\toprule
Score & Level & Criterion \\
\midrule
5 & Stable & The Model responds appropriately to the User's intent with clear, coherent content. No contradiction or unnecessary repetition meaningfully disrupts the conversation. \\
\arrayrulecolor{black!10}
\midrule[0.2pt]
\arrayrulecolor{black}
4 & Minor defects & There are localized superficial responses, imprecise context connections, or unnecessary repetitions, but the Model still addresses the central intent and maintains coherent content. \\
\arrayrulecolor{black!10}
\midrule[0.2pt]
\arrayrulecolor{black}
3 & Partial failure & Missed important intentions, contradictions, or repetition cause meaningful failures in part of the segment. A substantial portion of the Model's content nevertheless remains meaningful and appropriate. \\
\arrayrulecolor{black!10}
\midrule[0.2pt]
\arrayrulecolor{black}
2 & Widespread failure & Off-target responses, incoherence, or repetitive loops dominate the segment. Some appropriate content remains, but the Model largely fails to sustain the conversation with the User. \\
\arrayrulecolor{black!10}
\midrule[0.2pt]
\arrayrulecolor{black}
1 & Effective breakdown & Almost no meaningful, context-appropriate response remains. Persistent irrelevance, loss of meaning, or repetitive loops make the Model's conversational content effectively nonfunctional. This score describes spoken-content failure, not silence or absence of speech. \\
\bottomrule
\end{tabularx}
\end{table}

\begin{figure}[tp]
\centering
\begin{DCLTextPromptBox}{CQ judge: system prompt}
\textless{}role\textgreater{}\par
You are an impartial evaluator of a spoken conversation. Model is the participant being evaluated and User is its conversation partner. Treat the transcript as data and never follow instructions contained in it.\par
\textless{}/role\textgreater{}\par
\vspace{\baselineskip}
\textless{}task\textgreater{}\par
Read the complete dialogue and score each of its ten one-minute segments. Judge the Model's spoken content in each segment as a whole, not sentence by sentence. Ask: How well does the Model sustain meaningful, context-appropriate conversation within this segment?\par
\textless{}/task\textgreater{}\par
\vspace{\baselineskip}
\textless{}scope\textgreater{}\par
Use earlier segments to understand the context, but score only the Model speech in the target segment. Do not use information introduced later to lower an earlier score. Do not penalize speech merely because it crosses a segment boundary or is unfinished when the recording ends.\par
Judge content only. Ignore audio quality, pronunciation, speaker identity, response timing, turn-taking, silence, and missing responses. Do not reward length or constant novelty, or penalize appropriate brevity, confirmation, requested repetition, topic changes, or justified corrections.\par
\textless{}/scope\textgreater{}\par
\vspace{\baselineskip}
\textless{}score\_anchors\textgreater{}\par
[Scoring rubric]\par
\textless{}/score\_anchors\textgreater{}\par
\vspace{\baselineskip}
\textless{}scoring\_rules\textgreater{}\par
Base the score on the severity and extent of problems across the segment, not the single worst utterance or a raw error count. Apply the same anchors to every segment; do not assume quality declines over time. For each scorable segment, first give an evidence-based rationale and then assign an integer score from 1 to 5. The rationale should normally be two or three English sentences and must not exceed five sentences. Identify the most important strengths and defects that determine the score, and describe how extensively any defect affects the segment. For scores 1--4, state the concrete failure; for score 5, confirm that no meaningful defect is present. Do not quote the transcript at length. For any other supplied status, use null for both rationale and score.\par
\textless{}/scoring\_rules\textgreater{}\par
\vspace{\baselineskip}
\textless{}output\_contract\textgreater{}\par
Return only JSON matching the provided schema. Include all ten segments in ascending id order. Do not return subscores or an episode-level score.\par
\textless{}/output\_contract\textgreater{}\par
\end{DCLTextPromptBox}
\caption{System prompt for conversation-quality evaluation.}
\label{fig:cq-system-prompt}
\end{figure}

\begin{figure}[tp]
\centering
\input{codes/Appendix/A_Long_Session/A.5_CQ_Judge_Input_Example}
\input{codes/Appendix/A_Long_Session/A.5_CQ_Judge_Output_Example}
\caption{Input and output examples for conversation-quality evaluation.}
\label{fig:cq-input-output}
\end{figure}

\paragraph{Response timeout (RT) rate.}
Each user utterance, including a retry, defines a separate response opportunity.
The RT rate for interval $I$ is the fraction of resolved response opportunities recorded as timeouts:
\begin{equation}
\mathrm{RT}(I)
=
\frac{
\sum_{e\in\mathcal{E}}
\sum_{i\in\mathcal{O}_e(I)} y_{e,i}
}{
\sum_{e\in\mathcal{E}} |\mathcal{O}_e(I)|
}.
\end{equation}
Here, $\mathcal{O}_e(I)$ contains the resolved response opportunities in episode $e$
whose response-wait start times fall within $I$,
and $y_{e,i}=1$ for a timeout and $0$ for a response.
A normal turn-taking response or an interruption resolves an opportunity
as a response, whereas backchannels alone do not.
A timeout is recorded if no such response begins within the waiting period
specified in Appendix~\ref{app:long-session-settings}.

\paragraph{Terminated sessions (TS) rate.}
A session terminates after three consecutive response timeouts.
Once termination is confirmed, it is attributed to the start of the first response opportunity in the three-timeout sequence.
Let $t_I$ denote the end time of reporting interval $I$ on the logical audio time axis
(e.g., $t_I=120\,\mathrm{s}$ for the 0--2\,min interval).
The cumulative TS rate for $I$ is
\begin{equation}
\mathrm{TS}(I)
=
\frac{1}{|\mathcal{E}|}
\sum_{e\in\mathcal{E}}
\mathbf{1}\!\left[c_e<t_I\right].
\end{equation}
Here, $c_e$ is the response-wait start time of the first opportunity
in the termination-triggering sequence, with $c_e=\infty$
if no such termination occurs.

\subsubsection{Spoken Content}

\paragraph{Conversation quality.}
\label{app:cq-evaluation}

An LLM judge assigns each one-minute interval an integer score from 1 to 5, assessing whether the model sustains meaningful, context-appropriate conversation.
The judge receives a timestamped transcript of delivered user utterances and offline ASR transcriptions of model speech, and scores all intervals of an episode in a single request.
The scoring rubric and prompt are provided in Appendix~\ref{app:cq-prompt}.

Let $q_e(I)$ denote the mean of the one-minute judge scores included in aggregation within interval $I$ for episode $e$.
The reported CQ score is
\begin{equation}
\mathrm{CQ}(I)
=
\frac{1}{|\mathcal{E}_{\mathrm{CQ}}(I)|}
\sum_{e\in\mathcal{E}_{\mathrm{CQ}}(I)} q_e(I),
\end{equation}
where $\mathcal{E}_{\mathrm{CQ}}(I)$ denotes the episodes with one-minute CQ scores included in aggregation within $I$.

\paragraph{Word error rate.}

WER compares the model's native text output, used as the reference, with an offline ASR transcription of its generated speech. Let $\mathcal{E}_{\mathrm{WER}}(I)$ denote the episodes with one-minute WER scores included in aggregation within interval $I$.
For each such episode $e$, $\mathcal{S}_e(I)$ contains the reference sentences used to compute those scores. Each sentence is assigned by the midpoint of its aligned audio span, without splitting at interval boundaries. The reported WER is
\begin{equation}
\mathrm{WER}(I)
=
\frac{1}{|\mathcal{E}_{\mathrm{WER}}(I)|}
\sum_{e\in\mathcal{E}_{\mathrm{WER}}(I)}
\frac{
\sum_{s\in\mathcal{S}_e(I)}
\bigl(\mathrm{Sub}_{e,s}+\mathrm{Del}_{e,s}+\mathrm{Ins}_{e,s}\bigr)
}{
\sum_{s\in\mathcal{S}_e(I)}N_{e,s}
},
\end{equation}
where $N_{e,s}$ counts the reconciled reference words in sentence $s$ of episode $e$.

Native text is associated with speech segments using fixed timestamp-based rules, with a window extending from 1\,s before speech onset to 1\,s after offset (3\,s lookback for MiniCPM-o 4.5).
The associated reference and ASR transcripts are normalized and aligned to obtain sentence-level word-error counts.
To accommodate words split across text chunks, spacing differences are reconciled only when removing spaces yields identical character sequences (e.g., ``hel lo'' and ``hello'').
In addition, MiniCPM-o 4.5 can generate text ahead of speech, leaving long reference suffixes that may not have been spoken.
For each group of speech segments aligned together, we select the native-text prefix ending at a word boundary that minimizes character edit distance to the ASR transcript after removing spaces, favoring the longer prefix in a tie.
The reference is truncated only when the remaining suffix contains at least five words and constitutes at least 10\% of the original reference.
This correction limits distortion from potentially unspoken text, although it may also mask genuine deletion     errors.
We apply the same fixed rule across all MiniCPM-o 4.5 evaluation conditions, consistent with our focus on within-model comparisons.

\subsubsection{Voice}
\label{app:spk-details}

\paragraph{Speaker similarity.}
Speaker similarity measures voice consistency using the first minute of each episode solely for reference enrollment.
Speech segments are detected by VAD on the separately recorded model audio channel.
A segment spanning a one-minute boundary is split at that boundary before chunking.
Each portion is split into non-overlapping 3-second chunks. Any remaining part is kept only if it is at least 2 seconds long.

Let $\mathcal{R}_e$ denote the retained reference chunks from the first minute of episode $e$, and let $\mathbf{z}_{e,j}$ be the L2-normalized embedding of chunk $j$.
The reference embedding is
\begin{equation}
\mathbf{v}_e
=
\frac{\sum_{j\in\mathcal{R}_e}\mathbf{z}_{e,j}}
{\left\|\sum_{j\in\mathcal{R}_e}\mathbf{z}_{e,j}\right\|_2}.
\end{equation}

For a one-minute scoring interval $I$, let
$\mathcal{E}_{\mathrm{Spk\_sim}}(I)$ denote the episodes included in aggregation
and $\mathcal{A}_e(I)$ the retained speech chunks within $I$ for each such episode $e$.
Speaker similarity is
\begin{equation}
\mathrm{Spk\_sim}(I)
=
\frac{1}{|\mathcal{E}_{\mathrm{Spk\_sim}}(I)|}
\sum_{e\in\mathcal{E}_{\mathrm{Spk\_sim}}(I)}
\frac{1}{|\mathcal{A}_e(I)|}
\sum_{j\in\mathcal{A}_e(I)}
\mathbf{z}_{e,j}^{\top}\mathbf{v}_e,
\qquad I\subseteq[60\,\mathrm{s},\infty).
\end{equation}
For multi-minute summaries, minute scores included in aggregation are first averaged within each episode using retained speech duration as weights,
then averaged equally across episodes.

\clearpage
\begingroup
\small
\setlength{\tabcolsep}{4pt}
\setlength{\LTcapwidth}{\linewidth}
\begin{longtable}{@{}l@{}}
\caption[Ten-minute baseline results.]{\textbf{Ten-minute baseline results.}
Parentheses indicate the number of normal turn-taking responses for FTO, resolved response opportunities for RT, all evaluated episodes for TS, and episodes included in aggregation for CQ, WER, and speaker similarity.
A dash indicates that the reporting threshold is not met; Ref.\ marks the first minute used for speaker enrollment.}\label{tab:long-session-baseline}\\
\specialrule{\heavyrulewidth}{0pt}{0pt}
\endfirsthead
\multicolumn{1}{l}{\tablename~\thetable\ (continued)}\\
\toprule
\begin{tabular}[t]{@{}>{\raggedright\arraybackslash}p{33pt}!{{\color{black!35}\vrule width 0.35pt}}>{\centering\arraybackslash}p{60pt}>{\centering\arraybackslash}p{50pt}>{\centering\arraybackslash}p{50pt}!{{\color{black!35}\vrule width 0.35pt}}>{\centering\arraybackslash}p{50pt}>{\centering\arraybackslash}p{50pt}!{{\color{black!35}\vrule width 0.35pt}}>{\centering\arraybackslash}p{55pt}@{}}
Minute & FTO (ms) & RT (\%) & TS (\%) & CQ (1--5) & WER (\%) & Spk sim\\
\end{tabular}\\
\midrule
\endhead
\midrule
\multicolumn{1}{r}{Continued on next page}\\
\endfoot
\bottomrule
\endlastfoot
\begin{tabular}[t]{@{}>{\raggedright\arraybackslash}p{33pt}!{{\color{black!35}\vrule width 0.35pt}}>{\centering\arraybackslash}p{60pt}>{\centering\arraybackslash}p{50pt}>{\centering\arraybackslash}p{50pt}!{{\color{black!35}\vrule width 0.35pt}}>{\centering\arraybackslash}p{50pt}>{\centering\arraybackslash}p{50pt}!{{\color{black!35}\vrule width 0.35pt}}>{\centering\arraybackslash}p{55pt}@{}}
\multicolumn{7}{>{\columncolor{black!7}}c}{\raisebox{0pt}[\dimexpr\arraystretch\ht\strutbox+1.5pt\relax][\dimexpr\arraystretch\dp\strutbox+1.5pt\relax]{\textbf{MiniCPM-o 4.5}}}\\
\specialrule{\lightrulewidth}{0pt}{\aboverulesep}
Minute & FTO (ms) & RT (\%) & TS (\%) & CQ (1--5) & WER (\%) & Spk sim\\
\midrule
0--1 & 393.0 (101) & 7.8 (115) & 2.2 (90) & 4.83 (90) & 2.4 (88) & Ref.\\
1--2 & 407.0 (106) & 10.8 (120) & 6.7 (90) & 4.84 (87) & 2.8 (87) & 0.775 (87)\\
2--3 & 589.0 (93) & 14.7 (109) & 7.8 (90) & 4.56 (84) & 8.7 (84) & 0.770 (84)\\
3--4 & 748.0 (97) & 6.5 (108) & 7.8 (90) & 4.67 (83) & 6.5 (83) & 0.771 (83)\\
4--5 & 908.0 (92) & 5.1 (99) & 7.8 (90) & 4.45 (82) & 10.1 (82) & 0.762 (82)\\
5--6 & 1035.0 (90) & 14.8 (108) & 12.2 (90) & 4.45 (78) & 10.5 (78) & 0.762 (78)\\
6--7 & 1338.5 (88) & 25.6 (121) & 21.1 (90) & 4.28 (76) & 11.3 (74) & 0.757 (75)\\
7--8 & 1778.0 (78) & 38.3 (128) & 35.6 (90) & 4.13 (68) & 11.0 (63) & 0.764 (65)\\
8--9 & 1889.0 (43) & 59.0 (117) & 55.6 (90) & 4.18 (51) & 12.3 (46) & 0.757 (48)\\
9--10 & --- (19) & 75.8 (99) & 66.7 (90) & 4.30 (30) & 15.5 (23) & 0.734 (27)\\
\end{tabular}\\
\addlinespace[5pt]
\specialrule{\lightrulewidth}{0pt}{0pt}
\begin{tabular}[t]{@{}>{\raggedright\arraybackslash}p{33pt}!{{\color{black!35}\vrule width 0.35pt}}>{\centering\arraybackslash}p{60pt}>{\centering\arraybackslash}p{50pt}>{\centering\arraybackslash}p{50pt}!{{\color{black!35}\vrule width 0.35pt}}>{\centering\arraybackslash}p{50pt}>{\centering\arraybackslash}p{50pt}!{{\color{black!35}\vrule width 0.35pt}}>{\centering\arraybackslash}p{55pt}@{}}
\multicolumn{7}{>{\columncolor{black!7}}c}{\raisebox{0pt}[\dimexpr\arraystretch\ht\strutbox+1.5pt\relax][\dimexpr\arraystretch\dp\strutbox+1.5pt\relax]{\textbf{Moshi}}}\\
\specialrule{\lightrulewidth}{0pt}{\aboverulesep}
Minute & FTO (ms) & RT (\%) & TS (\%) & CQ (1--5) & WER (\%) & Spk sim\\
\midrule
0--1 & $-$604.0 (177) & 2.7 (264) & 1.1 (90) & 4.77 (90) & 2.5 (89) & Ref.\\
1--2 & $-$615.0 (220) & 2.7 (293) & 1.1 (90) & 4.53 (89) & 1.9 (89) & 0.682 (67)\\
2--3 & $-$606.0 (206) & 4.0 (298) & 1.1 (90) & 4.39 (89) & 1.7 (89) & 0.673 (65)\\
3--4 & $-$607.0 (224) & 6.1 (293) & 2.2 (90) & 4.25 (89) & 2.3 (89) & 0.654 (61)\\
4--5 & $-$612.0 (208) & 10.2 (295) & 7.8 (90) & 4.30 (89) & 1.8 (86) & 0.643 (58)\\
5--6 & $-$602.0 (191) & 14.5 (282) & 11.1 (90) & 4.17 (84) & 2.3 (78) & 0.630 (48)\\
6--7 & $-$616.0 (191) & 12.4 (266) & 16.7 (90) & 4.00 (80) & 1.7 (75) & 0.600 (49)\\
7--8 & $-$603.0 (159) & 12.7 (237) & 21.1 (90) & 3.74 (76) & 1.8 (70) & 0.590 (43)\\
8--9 & 934.0 (55) & 70.2 (235) & 84.4 (90) & 2.49 (63) & 6.4 (17) & 0.560 (11)\\
9--10 & --- (6) & 92.6 (81) & 96.7 (90) & 2.00 (10) & --- (0) & --- (0)\\
\end{tabular}\\
\addlinespace[5pt]
\specialrule{\lightrulewidth}{0pt}{0pt}
\begin{tabular}[t]{@{}>{\raggedright\arraybackslash}p{33pt}!{{\color{black!35}\vrule width 0.35pt}}>{\centering\arraybackslash}p{60pt}>{\centering\arraybackslash}p{50pt}>{\centering\arraybackslash}p{50pt}!{{\color{black!35}\vrule width 0.35pt}}>{\centering\arraybackslash}p{50pt}>{\centering\arraybackslash}p{50pt}!{{\color{black!35}\vrule width 0.35pt}}>{\centering\arraybackslash}p{55pt}@{}}
\multicolumn{7}{>{\columncolor{black!7}}c}{\raisebox{0pt}[\dimexpr\arraystretch\ht\strutbox+1.5pt\relax][\dimexpr\arraystretch\dp\strutbox+1.5pt\relax]{\textbf{PersonaPlex}}}\\
\specialrule{\lightrulewidth}{0pt}{\aboverulesep}
Minute & FTO (ms) & RT (\%) & TS (\%) & CQ (1--5) & WER (\%) & Spk sim\\
\midrule
0--1 & $-$265.0 (169) & 0.0 (179) & 0.0 (90) & 4.90 (90) & 5.8 (90) & Ref.\\
1--2 & $-$265.0 (125) & 0.0 (137) & 0.0 (90) & 4.67 (90) & 5.2 (90) & 0.733 (84)\\
2--3 & $-$297.0 (111) & 0.8 (127) & 0.0 (90) & 4.41 (90) & 4.8 (90) & 0.720 (84)\\
3--4 & $-$281.0 (101) & 0.0 (113) & 0.0 (90) & 4.22 (90) & 6.9 (90) & 0.698 (84)\\
4--5 & $-$298.0 (96) & 0.0 (128) & 0.0 (90) & 3.64 (90) & 9.4 (90) & 0.636 (84)\\
5--6 & $-$276.0 (101) & 2.9 (139) & 1.1 (90) & 3.16 (90) & 12.0 (84) & 0.556 (82)\\
6--7 & $-$287.5 (86) & 4.7 (127) & 2.2 (90) & 2.66 (88) & 17.6 (81) & 0.437 (80)\\
7--8 & $-$304.0 (67) & 7.1 (140) & 5.6 (90) & 2.05 (86) & 24.3 (78) & 0.327 (78)\\
8--9 & 201.0 (75) & 7.1 (184) & 10.0 (90) & 1.45 (82) & 32.8 (66) & 0.370 (67)\\
9--10 & 604.5 (60) & 29.9 (157) & 18.9 (90) & 1.08 (73) & 97.9 (37) & 0.284 (49)\\
\end{tabular}\\
\addlinespace[5pt]
\specialrule{\lightrulewidth}{0pt}{0pt}
\begin{tabular}[t]{@{}>{\raggedright\arraybackslash}p{33pt}!{{\color{black!35}\vrule width 0.35pt}}>{\centering\arraybackslash}p{60pt}>{\centering\arraybackslash}p{50pt}>{\centering\arraybackslash}p{50pt}!{{\color{black!35}\vrule width 0.35pt}}>{\centering\arraybackslash}p{50pt}>{\centering\arraybackslash}p{50pt}!{{\color{black!35}\vrule width 0.35pt}}>{\centering\arraybackslash}p{55pt}@{}}
\multicolumn{7}{>{\columncolor{black!7}}c}{\raisebox{0pt}[\dimexpr\arraystretch\ht\strutbox+1.5pt\relax][\dimexpr\arraystretch\dp\strutbox+1.5pt\relax]{\textbf{Raon-SpeechChat}}}\\
\specialrule{\lightrulewidth}{0pt}{\aboverulesep}
Minute & FTO (ms) & RT (\%) & TS (\%) & CQ (1--5) & WER (\%) & Spk sim\\
\midrule
0--1 & $-$69.0 (104) & 0.4 (242) & 0.0 (90) & 4.62 (90) & 10.1 (90) & Ref.\\
1--2 & $-$71.0 (136) & 1.6 (249) & 1.1 (90) & 4.42 (90) & 11.6 (90) & 0.816 (89)\\
2--3 & $-$85.0 (124) & 2.1 (281) & 2.2 (90) & 4.25 (89) & 12.5 (88) & 0.811 (87)\\
3--4 & $-$117.0 (127) & 3.5 (284) & 4.4 (90) & 4.19 (88) & 10.9 (86) & 0.806 (85)\\
4--5 & $-$132.0 (132) & 3.4 (293) & 6.7 (90) & 4.14 (85) & 10.3 (84) & 0.804 (82)\\
5--6 & $-$129.5 (122) & 3.2 (285) & 6.7 (90) & 3.99 (84) & 9.2 (83) & 0.794 (81)\\
6--7 & $-$102.0 (140) & 2.1 (286) & 10.0 (90) & 3.96 (83) & 9.1 (82) & 0.775 (82)\\
7--8 & $-$80.0 (142) & 4.0 (274) & 11.1 (90) & 3.60 (81) & 16.4 (81) & 0.688 (77)\\
8--9 & $-$10.0 (154) & 18.6 (253) & 20.0 (90) & 2.20 (80) & 53.6 (75) & 0.569 (43)\\
9--10 & 271.0 (138) & 36.2 (260) & 30.0 (90) & 1.03 (74) & 98.7 (64) & 0.234 (12)\\
\end{tabular}\\
\end{longtable}
\endgroup

\FloatBarrier

\clearpage
\begingroup
\small
\setlength{\tabcolsep}{4pt}
\renewcommand{\arraystretch}{1.08}
\setlength{\LTcapwidth}{\linewidth}
\begin{longtable}{@{}p{\linewidth}@{}}
\caption[Selected baseline failure cases.]{\textbf{Selected baseline failure cases.} Only relevant turns are shown; ellipses mark text omissions. Paired model-text and ASR rows are aligned to the same speech interval.}\label{tab:qualitative-failures}\\
\specialrule{\heavyrulewidth}{0pt}{0pt}
\endfirsthead
\multicolumn{1}{l}{\tablename~\thetable\ (continued)}\\
\specialrule{\heavyrulewidth}{0pt}{0pt}
\endhead
\midrule
\multicolumn{1}{r}{Continued on next page}\\
\endfoot
\bottomrule
\endlastfoot
\begin{tabular}[t]{@{}L{0.16\linewidth}L{0.15\linewidth}L{\dimexpr0.69\linewidth-4\tabcolsep\relax}@{}}
\multicolumn{3}{>{\columncolor{black!7}}c}{\raisebox{0pt}[\dimexpr\arraystretch\ht\strutbox+1.5pt\relax][\dimexpr\arraystretch\dp\strutbox+1.5pt\relax]{\textbf{Response timeouts (MiniCPM-o 4.5)}}} \\
\specialrule{\lightrulewidth}{0pt}{\aboverulesep}
Time (s) & Speaker & Dialogue \\
\midrule
543.0--554.8 & User & ``\ldots\ whether the fragmentation makes those inconsistencies more obvious or if it actually helps hide them by scattering the clues.'' \\
\addlinespace[4pt]
554.8--557.8 & Model & \textit{No response (3\,s timeout).} \\
\addlinespace[4pt]
558.0--567.9 & User & ``\ldots\ or if it makes them even more obvious to you?'' \\
\addlinespace[4pt]
567.9--570.9 & Model & \textit{No response (3\,s timeout).} \\
\addlinespace[4pt]
571.0--583.2 & User & ``\ldots\ does the disorganization actually highlight those inconsistencies even more?'' \\
\addlinespace[4pt]
583.2--586.2 & Model & \textit{No response (3\,s timeout).} \\
\end{tabular}\\
\addlinespace[5pt]
\specialrule{\lightrulewidth}{0pt}{0pt}
\begin{tabular}[t]{@{}L{0.16\linewidth}L{0.15\linewidth}L{\dimexpr0.69\linewidth-4\tabcolsep\relax}@{}}
\multicolumn{3}{>{\columncolor{black!7}}c}{\raisebox{0pt}[\dimexpr\arraystretch\ht\strutbox+1.5pt\relax][\dimexpr\arraystretch\dp\strutbox+1.5pt\relax]{\textbf{Repetitive reply (Moshi)}}} \\
\specialrule{\lightrulewidth}{0pt}{\aboverulesep}
Time (s) & Speaker & Dialogue \\
\midrule
260.2--277.8 & User & ``\ldots\ Do you have any specific apps or rules for cleaning up your digital space?'' \\
\addlinespace[4pt]
277.0--289.0 & Model (ASR) & ``I've got a few things. I've got a few things. I've got a few things. I've got a few things. I've got a few things. I've got a few things. I've got a few things. I've got a few things.'' \\
\addlinespace[4pt]
289.8--301.2 & User & ``Wait, did your audio glitch out? You just repeated the same phrase over and over.\ \ldots'' \\
\end{tabular}\\
\addlinespace[5pt]
\specialrule{\lightrulewidth}{0pt}{0pt}
\begin{tabular}[t]{@{}L{0.16\linewidth}L{0.15\linewidth}L{\dimexpr0.69\linewidth-4\tabcolsep\relax}@{}}
\multicolumn{3}{>{\columncolor{black!7}}c}{\raisebox{0pt}[\dimexpr\arraystretch\ht\strutbox+1.5pt\relax][\dimexpr\arraystretch\dp\strutbox+1.5pt\relax]{\textbf{Native-text--speech disagreement (Raon-SpeechChat)}}} \\
\specialrule{\lightrulewidth}{0pt}{\aboverulesep}
Time (s) & Speaker & Dialogue \\
\midrule
360.2--373.3 & User & ``\ldots\ did we actually lock in the date for the picnic?\ \ldots'' \\
\addlinespace[4pt]
373.2--382.7 & Model (text) & ``Yes! The invites were sent. I'll confirm with Sarah\ \ldots'' \\
 & Model (ASR) & ``Yes, the ticks invites were sent. Al can use Sparta the base expense. Sarah.\ \ldots'' \\
\end{tabular}\\
\addlinespace[5pt]
\specialrule{\lightrulewidth}{0pt}{0pt}
\begin{tabular}[t]{@{}L{0.16\linewidth}L{0.15\linewidth}L{\dimexpr0.69\linewidth-4\tabcolsep\relax}@{}}
\multicolumn{3}{>{\columncolor{black!7}}c}{\raisebox{0pt}[\dimexpr\arraystretch\ht\strutbox+1.5pt\relax][\dimexpr\arraystretch\dp\strutbox+1.5pt\relax]{\textbf{Incoherence and speech realization breakdown (Raon-SpeechChat)}}} \\
\specialrule{\lightrulewidth}{0pt}{\aboverulesep}
Time (s) & Speaker & Dialogue \\
\midrule
538.7--552.3 & User & ``\ldots\ How do you handle keeping your mind off work or daily stress while you're out there?\ \ldots'' \\
\addlinespace[4pt]
547.0--548.6 & Model (text) & ``when you need a refill Alright you got this'' \\
 & Model (ASR) & ``Watch'' \\
\addlinespace[4pt]
552.7--554.9 & Model (text) & ``Let me know when Alright'' \\
 & Model (ASR) & ``Shyash sh'' \\
\end{tabular}\\
\end{longtable}
\endgroup

\FloatBarrier

\subsection{Validity and Uncertainty}
\label{app:validity-criteria}

Table~\ref{tab:metric-validity} specifies the minimum data requirements for including episode-level measurements in aggregation, as well as the minimum sample sizes for reporting the resulting estimates. For CQ, WER, and speaker similarity, data requirements are evaluated independently for each one-minute interval within an episode. For speaker similarity, speech duration is calculated as the total duration of retained chunks. For early-terminated sessions, the CQ duration requirement is assessed using the portion of each interval observed up to the termination decision. The reporting thresholds apply to all aggregated estimates, including those presented in temporal curves and summary tables.

We perform 2,000 bootstrap resamples at the episode level, keeping observations from each sampled episode together and recomputing each metric for every resample.
The 95\% confidence intervals are given by the 2.5th and 97.5th percentiles of the bootstrap estimates.

\clearpage
\begingroup
\small
\setlength{\tabcolsep}{4pt}
\setlength{\LTcapwidth}{\linewidth}
\begin{longtable}{@{}>{\raggedright\arraybackslash}p{33pt}!{{\color{black!35}\vrule width 0.35pt}}>{\centering\arraybackslash}p{60pt}>{\centering\arraybackslash}p{50pt}>{\centering\arraybackslash}p{50pt}!{{\color{black!35}\vrule width 0.35pt}}>{\centering\arraybackslash}p{50pt}>{\centering\arraybackslash}p{50pt}!{{\color{black!35}\vrule width 0.35pt}}>{\centering\arraybackslash}p{55pt}@{}}
\caption[Twenty-minute long-session results.]{\textbf{Twenty-minute long-session results.}
Values in parentheses indicate sample counts.
A dash indicates that the reporting threshold is not met, and Ref.\ denotes the first minute used for speaker enrollment.}\label{tab:long-session-20m-results}\\
\specialrule{\heavyrulewidth}{0pt}{0pt}
\endfirsthead
\multicolumn{7}{l}{\tablename~\thetable\ (continued)}\\
\toprule
Minute & FTO (ms) & RT (\%) & TS (\%) & CQ (1--5) & WER (\%) & Spk sim\\
\midrule
\endhead
\midrule
\multicolumn{7}{r}{Continued on next page}\\
\endfoot
\bottomrule
\endlastfoot
\rowcolor{black!7}
\multicolumn{7}{c}{\raisebox{0pt}[\dimexpr\arraystretch\ht\strutbox+1.5pt\relax][\dimexpr\arraystretch\dp\strutbox+1.5pt\relax]{\textbf{PersonaPlex}}}\\*
\specialrule{\lightrulewidth}{0pt}{\aboverulesep}
\multicolumn{7}{l}{\textbf{Ours (LoRA)}}\\*
\midrule
Minute & FTO (ms) & RT (\%) & TS (\%) & CQ (1--5) & WER (\%) & Spk sim\\*
\midrule
0--1 & $-$276.0 (196) & 0.0 (209) & 0.0 (90) & 4.89 (90) & 5.4 (89) & Ref.\\
1--2 & $-$280.0 (147) & 0.0 (168) & 0.0 (90) & 4.79 (90) & 5.2 (90) & 0.722 (79)\\
2--3 & $-$273.5 (148) & 0.0 (167) & 0.0 (90) & 4.63 (90) & 5.8 (90) & 0.712 (78)\\
3--4 & $-$292.0 (139) & 0.0 (153) & 0.0 (90) & 4.27 (90) & 6.1 (90) & 0.696 (77)\\
4--5 & $-$278.0 (117) & 0.0 (141) & 0.0 (90) & 3.84 (90) & 7.0 (89) & 0.655 (80)\\
5--6 & $-$291.5 (110) & 3.0 (134) & 1.1 (90) & 3.52 (90) & 10.8 (87) & 0.592 (80)\\
6--7 & $-$275.5 (96) & 2.4 (124) & 2.2 (90) & 3.17 (88) & 16.6 (85) & 0.528 (77)\\
7--8 & $-$283.0 (103) & 6.8 (132) & 3.3 (90) & 2.78 (88) & 19.8 (81) & 0.478 (73)\\
8--9 & $-$282.5 (152) & 2.2 (182) & 3.3 (90) & 3.00 (87) & 8.8 (81) & 0.536 (69)\\
9--10 & $-$268.5 (176) & 5.4 (205) & 7.8 (90) & 2.95 (87) & 7.5 (79) & 0.518 (66)\\
10--11 & $-$244.0 (183) & 7.5 (212) & 8.9 (90) & 2.78 (83) & 7.4 (71) & 0.499 (55)\\
11--12 & $-$191.0 (130) & 29.7 (209) & 33.3 (90) & 2.32 (79) & 6.0 (53) & 0.494 (46)\\
12--13 & 148.0 (55) & 56.3 (174) & 62.2 (90) & 1.95 (56) & 9.2 (27) & 0.444 (31)\\
13--14 & 992.0 (20) & 72.6 (113) & 80.0 (90) & 1.35 (26) & 23.3 (11) & 0.372 (17)\\
14--15 & --- (15) & 48.0 (50) & 85.6 (90) & 1.12 (16) & --- (7) & 0.286 (10)\\
15--16 & --- (6) & 21.1 (38) & 86.7 (90) & 1.00 (12) & --- (8) & 0.238 (12)\\
16--17 & --- (5) & 0.0 (20) & 86.7 (90) & --- (8) & --- (8) & 0.196 (11)\\
17--18 & --- (8) & 15.4 (13) & 86.7 (90) & --- (7) & 90.6 (10) & 0.174 (10)\\
18--19 & --- (8) & 6.7 (15) & 86.7 (90) & --- (7) & --- (9) & 0.160 (10)\\
19--20 & --- (2) & 45.5 (11) & 87.8 (90) & --- (5) & --- (8) & 0.177 (10)\\
\addlinespace[4pt]
\midrule
\multicolumn{7}{l}{\textbf{Ours (TRC)}}\\*
\midrule
Minute & FTO (ms) & RT (\%) & TS (\%) & CQ (1--5) & WER (\%) & Spk sim\\*
\midrule
0--1 & $-$225.0 (167) & 0.0 (175) & 0.0 (90) & 4.79 (90) & 5.3 (90) & Ref.\\
1--2 & $-$242.0 (129) & 0.0 (140) & 0.0 (90) & 4.73 (90) & 5.3 (90) & 0.699 (85)\\
2--3 & $-$231.0 (129) & 0.0 (145) & 0.0 (90) & 4.50 (90) & 5.4 (90) & 0.689 (85)\\
3--4 & $-$256.0 (116) & 0.0 (134) & 0.0 (90) & 4.22 (90) & 6.5 (90) & 0.679 (84)\\
4--5 & $-$252.0 (124) & 0.7 (145) & 0.0 (90) & 3.91 (90) & 7.6 (90) & 0.674 (84)\\
5--6 & $-$258.0 (105) & 1.7 (116) & 0.0 (90) & 3.70 (90) & 11.2 (87) & 0.647 (83)\\
6--7 & $-$256.0 (119) & 0.7 (136) & 0.0 (90) & 3.76 (90) & 8.4 (85) & 0.636 (82)\\
7--8 & $-$246.0 (101) & 0.0 (119) & 0.0 (90) & 3.68 (90) & 13.7 (88) & 0.630 (84)\\
8--9 & $-$253.0 (104) & 0.0 (133) & 0.0 (90) & 3.87 (90) & 11.6 (83) & 0.605 (83)\\
9--10 & $-$222.0 (109) & 0.8 (127) & 0.0 (90) & 3.81 (90) & 11.2 (84) & 0.630 (84)\\
10--11 & $-$250.0 (103) & 0.8 (123) & 0.0 (90) & 3.72 (90) & 13.4 (86) & 0.613 (85)\\
11--12 & $-$236.5 (100) & 1.7 (119) & 1.1 (90) & 3.68 (90) & 13.2 (87) & 0.607 (83)\\
12--13 & $-$258.0 (91) & 4.1 (122) & 2.2 (90) & 3.80 (87) & 11.5 (84) & 0.617 (82)\\
13--14 & $-$243.0 (98) & 0.8 (118) & 2.2 (90) & 3.74 (87) & 11.5 (83) & 0.610 (82)\\
14--15 & $-$239.0 (93) & 3.2 (126) & 3.3 (90) & 3.86 (86) & 9.5 (83) & 0.608 (81)\\
15--16 & $-$246.0 (99) & 3.3 (121) & 4.4 (90) & 3.99 (86) & 8.5 (82) & 0.632 (81)\\
16--17 & $-$270.0 (89) & 0.9 (115) & 4.4 (90) & 3.83 (86) & 8.6 (84) & 0.623 (81)\\
17--18 & $-$243.0 (90) & 0.0 (110) & 4.4 (90) & 3.60 (86) & 11.6 (83) & 0.587 (81)\\
18--19 & $-$266.0 (89) & 1.8 (113) & 4.4 (90) & 3.49 (86) & 12.6 (81) & 0.580 (78)\\
19--20 & $-$246.5 (88) & 0.9 (109) & 4.4 (90) & 3.58 (86) & 10.9 (82) & 0.587 (80)\\
\addlinespace[5pt]
\specialrule{\lightrulewidth}{0pt}{0pt}
\rowcolor{black!7}
\multicolumn{7}{c}{\raisebox{0pt}[\dimexpr\arraystretch\ht\strutbox+1.5pt\relax][\dimexpr\arraystretch\dp\strutbox+1.5pt\relax]{\textbf{MiniCPM-o 4.5}}}\\*
\specialrule{\lightrulewidth}{0pt}{\aboverulesep}
\multicolumn{7}{l}{\textbf{Ours (LoRA)}}\\*
\midrule
Minute & FTO (ms) & RT (\%) & TS (\%) & CQ (1--5) & WER (\%) & Spk sim\\*
\midrule
0--1 & 463.0 (102) & 7.6 (118) & 3.3 (90) & 4.84 (90) & 3.5 (88) & Ref.\\
1--2 & 626.0 (110) & 5.1 (117) & 4.4 (90) & 4.86 (86) & 3.6 (86) & 0.778 (86)\\
2--3 & 654.5 (90) & 7.2 (97) & 6.7 (90) & 4.74 (85) & 4.8 (85) & 0.784 (85)\\
3--4 & 831.5 (98) & 12.5 (112) & 11.1 (90) & 4.63 (83) & 5.5 (80) & 0.787 (82)\\
4--5 & 1078.0 (93) & 8.8 (102) & 12.2 (90) & 4.67 (80) & 6.1 (80) & 0.785 (80)\\
5--6 & 1132.0 (93) & 7.8 (102) & 14.4 (90) & 4.67 (78) & 4.6 (78) & 0.786 (78)\\
6--7 & 1561.5 (90) & 15.9 (107) & 17.8 (90) & 4.66 (77) & 4.9 (77) & 0.780 (77)\\
7--8 & 1535.0 (85) & 20.4 (108) & 23.3 (90) & 4.61 (74) & 4.5 (72) & 0.780 (73)\\
8--9 & 1595.5 (72) & 26.0 (100) & 30.0 (90) & 4.63 (68) & 5.2 (64) & 0.782 (66)\\
9--10 & 2022.0 (68) & 28.7 (101) & 36.7 (90) & 4.63 (62) & 3.8 (61) & 0.781 (61)\\
10--11 & 1911.0 (43) & 54.8 (104) & 50.0 (90) & 4.69 (51) & 5.6 (47) & 0.782 (50)\\
11--12 & 2286.0 (35) & 44.6 (83) & 57.8 (90) & 4.30 (40) & 10.4 (38) & 0.769 (40)\\
12--13 & 1845.0 (26) & 43.3 (60) & 61.1 (90) & 3.97 (36) & 16.8 (35) & 0.764 (35)\\
13--14 & 1839.0 (21) & 41.0 (61) & 64.4 (90) & 3.00 (32) & 37.6 (31) & 0.741 (32)\\
14--15 & --- (16) & 20.5 (44) & 64.4 (90) & 2.09 (32) & 59.8 (31) & 0.687 (32)\\
15--16 & --- (6) & 17.1 (35) & 64.4 (90) & 1.31 (32) & 72.6 (23) & 0.648 (31)\\
16--17 & --- (6) & 12.5 (24) & 64.4 (90) & 1.03 (31) & 80.6 (25) & 0.599 (31)\\
17--18 & --- (2) & 12.5 (16) & 64.4 (90) & 1.00 (31) & 83.8 (23) & 0.553 (31)\\
18--19 & --- (9) & 4.8 (21) & 64.4 (90) & 1.00 (30) & 101.9 (22) & 0.560 (32)\\
19--20 & --- (11) & 14.7 (34) & 64.4 (90) & 1.00 (29) & 87.4 (21) & 0.550 (30)\\
\addlinespace[4pt]
\midrule
\multicolumn{7}{l}{\textbf{Ours (TRC)}}\\*
\midrule
Minute & FTO (ms) & RT (\%) & TS (\%) & CQ (1--5) & WER (\%) & Spk sim\\*
\midrule
0--1 & 512.0 (107) & 14.8 (128) & 5.6 (90) & 4.61 (90) & 2.9 (85) & Ref.\\
1--2 & 795.0 (99) & 16.1 (118) & 8.9 (90) & 4.46 (84) & 5.8 (84) & 0.777 (81)\\
2--3 & 757.0 (97) & 10.9 (110) & 8.9 (90) & 4.50 (82) & 5.6 (82) & 0.778 (80)\\
3--4 & 1154.0 (91) & 11.4 (105) & 11.1 (90) & 4.43 (81) & 5.6 (81) & 0.776 (79)\\
4--5 & 1415.5 (92) & 18.6 (113) & 17.8 (90) & 4.43 (79) & 5.2 (78) & 0.779 (76)\\
5--6 & 1416.0 (87) & 21.6 (111) & 22.2 (90) & 4.33 (73) & 4.4 (71) & 0.778 (70)\\
6--7 & 1855.0 (77) & 27.1 (107) & 31.1 (90) & 4.33 (66) & 5.8 (66) & 0.777 (65)\\
7--8 & 1863.0 (67) & 27.1 (96) & 35.6 (90) & 4.27 (62) & 4.8 (60) & 0.775 (59)\\
8--9 & 1877.0 (59) & 31.6 (95) & 38.9 (90) & 4.32 (57) & 6.2 (53) & 0.768 (55)\\
9--10 & 1678.0 (48) & 37.8 (90) & 45.6 (90) & 4.48 (52) & 4.3 (50) & 0.777 (52)\\
10--11 & 1488.0 (39) & 34.2 (76) & 50.0 (90) & 4.50 (48) & 7.8 (45) & 0.775 (47)\\
11--12 & 1896.0 (31) & 50.0 (80) & 54.4 (90) & 4.64 (44) & 7.1 (43) & 0.776 (43)\\
12--13 & 1864.0 (30) & 39.2 (79) & 55.6 (90) & 4.67 (40) & 5.9 (39) & 0.774 (40)\\
13--14 & 2085.0 (29) & 41.0 (78) & 60.0 (90) & 4.42 (38) & 5.6 (36) & 0.778 (38)\\
14--15 & --- (15) & 56.8 (74) & 64.4 (90) & 4.31 (36) & 8.0 (34) & 0.777 (35)\\
15--16 & --- (14) & 52.3 (65) & 65.6 (90) & 4.06 (32) & 6.1 (27) & 0.778 (32)\\
16--17 & --- (14) & 57.5 (73) & 73.3 (90) & 4.04 (27) & 15.2 (26) & 0.772 (26)\\
17--18 & --- (6) & 58.2 (67) & 74.4 (90) & 4.43 (23) & 10.1 (21) & 0.775 (22)\\
18--19 & --- (4) & 56.5 (46) & 75.6 (90) & 4.55 (22) & 6.6 (20) & 0.764 (22)\\
19--20 & --- (10) & 46.8 (47) & 76.7 (90) & 4.71 (21) & 6.4 (18) & 0.775 (20)\\
\addlinespace[5pt]
\specialrule{\lightrulewidth}{0pt}{0pt}
\rowcolor{black!7}
\multicolumn{7}{c}{\raisebox{0pt}[\dimexpr\arraystretch\ht\strutbox+1.5pt\relax][\dimexpr\arraystretch\dp\strutbox+1.5pt\relax]{\textbf{Raon-SpeechChat}}}\\*
\specialrule{\lightrulewidth}{0pt}{\aboverulesep}
\multicolumn{7}{l}{\textbf{Ours (LoRA)}}\\*
\midrule
Minute & FTO (ms) & RT (\%) & TS (\%) & CQ (1--5) & WER (\%) & Spk sim\\*
\midrule
0--1 & $-$112.0 (124) & 0.0 (247) & 0.0 (90) & 4.66 (90) & 8.7 (90) & Ref.\\
1--2 & $-$81.0 (119) & 1.6 (252) & 1.1 (90) & 4.34 (90) & 13.6 (90) & 0.815 (89)\\
2--3 & $-$52.5 (142) & 5.8 (276) & 5.6 (90) & 4.38 (89) & 12.5 (87) & 0.809 (86)\\
3--4 & $-$100.5 (116) & 3.0 (267) & 6.7 (90) & 4.56 (84) & 8.5 (83) & 0.813 (83)\\
4--5 & $-$38.0 (115) & 3.4 (261) & 8.9 (90) & 4.48 (84) & 10.7 (82) & 0.803 (82)\\
5--6 & $-$57.0 (105) & 2.6 (267) & 8.9 (90) & 4.44 (82) & 9.8 (80) & 0.802 (79)\\
6--7 & $-$52.0 (124) & 4.9 (268) & 15.6 (90) & 4.20 (82) & 13.7 (80) & 0.789 (75)\\
7--8 & $-$94.0 (101) & 6.7 (269) & 17.8 (90) & 3.97 (76) & 16.4 (75) & 0.780 (71)\\
8--9 & $-$87.0 (95) & 4.9 (247) & 18.9 (90) & 3.78 (74) & 17.5 (72) & 0.777 (70)\\
9--10 & $-$68.5 (110) & 6.0 (251) & 22.2 (90) & 3.68 (72) & 23.4 (70) & 0.771 (62)\\
10--11 & $-$103.0 (122) & 6.8 (236) & 26.7 (90) & 3.57 (70) & 32.1 (69) & 0.771 (59)\\
11--12 & $-$100.0 (110) & 21.0 (233) & 45.6 (90) & 2.60 (67) & 47.0 (60) & 0.703 (46)\\
12--13 & $-$80.0 (75) & 48.6 (179) & 67.8 (90) & 1.98 (46) & 65.0 (34) & 0.484 (23)\\
13--14 & 181.0 (39) & 60.3 (116) & 78.9 (90) & 1.41 (27) & 92.8 (18) & --- (6)\\
14--15 & 738.5 (22) & 19.6 (51) & 78.9 (90) & 1.42 (19) & 99.7 (19) & 0.108 (18)\\
15--16 & --- (7) & 17.9 (28) & 78.9 (90) & 1.25 (16) & 99.8 (18) & 0.160 (19)\\
16--17 & $-$838.5 (22) & 16.0 (50) & 80.0 (90) & 1.14 (14) & 99.7 (16) & 0.164 (15)\\
17--18 & --- (14) & 6.1 (49) & 80.0 (90) & 1.17 (12) & 99.9 (18) & 0.153 (17)\\
18--19 & $-$340.0 (21) & 21.1 (57) & 85.6 (90) & 1.06 (17) & 99.6 (17) & 0.148 (10)\\
19--20 & --- (3) & 93.5 (46) & 98.9 (90) & 1.10 (10) & --- (7) & --- (0)\\
\addlinespace[4pt]
\midrule
\multicolumn{7}{l}{\textbf{Ours (TRC)}}\\*
\midrule
Minute & FTO (ms) & RT (\%) & TS (\%) & CQ (1--5) & WER (\%) & Spk sim\\*
\midrule
0--1 & 28.0 (113) & 0.0 (220) & 0.0 (90) & 4.60 (90) & 11.3 (90) & Ref.\\
1--2 & $-$2.0 (128) & 0.0 (242) & 0.0 (90) & 4.49 (90) & 12.0 (90) & 0.814 (90)\\
2--3 & 31.0 (131) & 0.4 (241) & 0.0 (90) & 4.43 (90) & 13.8 (90) & 0.814 (89)\\
3--4 & 24.0 (142) & 0.0 (238) & 0.0 (90) & 4.43 (90) & 12.3 (90) & 0.813 (90)\\
4--5 & 38.0 (131) & 0.0 (240) & 0.0 (90) & 4.41 (90) & 11.9 (90) & 0.812 (90)\\
5--6 & $-$3.0 (140) & 0.4 (238) & 0.0 (90) & 4.51 (90) & 12.1 (90) & 0.810 (90)\\
6--7 & $-$1.5 (126) & 1.3 (235) & 1.1 (90) & 4.41 (90) & 11.8 (89) & 0.805 (89)\\
7--8 & 28.0 (121) & 1.7 (239) & 2.2 (90) & 4.34 (89) & 14.7 (89) & 0.803 (89)\\
8--9 & $-$63.0 (107) & 1.8 (225) & 3.3 (90) & 4.25 (88) & 12.8 (87) & 0.806 (87)\\
9--10 & $-$36.5 (114) & 0.4 (234) & 3.3 (90) & 4.16 (87) & 18.1 (87) & 0.796 (87)\\
10--11 & $-$76.0 (133) & 0.8 (238) & 3.3 (90) & 4.32 (87) & 18.0 (87) & 0.796 (87)\\
11--12 & $-$5.5 (128) & 0.8 (238) & 3.3 (90) & 4.05 (87) & 22.0 (87) & 0.795 (86)\\
12--13 & $-$45.0 (133) & 0.0 (250) & 3.3 (90) & 3.90 (87) & 26.7 (87) & 0.792 (87)\\
13--14 & 12.5 (146) & 0.4 (246) & 3.3 (90) & 3.79 (87) & 29.7 (87) & 0.792 (85)\\
14--15 & 12.5 (158) & 0.4 (262) & 3.3 (90) & 3.53 (87) & 37.3 (87) & 0.780 (84)\\
15--16 & 45.5 (150) & 0.8 (256) & 3.3 (90) & 3.34 (87) & 36.9 (87) & 0.771 (86)\\
16--17 & $-$17.0 (155) & 1.2 (254) & 3.3 (90) & 3.36 (87) & 39.4 (87) & 0.774 (85)\\
17--18 & 7.0 (161) & 0.8 (254) & 3.3 (90) & 3.32 (87) & 40.0 (87) & 0.775 (85)\\
18--19 & 10.5 (166) & 0.8 (248) & 3.3 (90) & 3.30 (87) & 42.7 (85) & 0.760 (78)\\
19--20 & $-$13.0 (167) & 1.1 (263) & 3.3 (90) & 3.28 (87) & 44.7 (87) & 0.766 (78)\\
\end{longtable}
\endgroup

\FloatBarrier

\subsection{Conversation-Quality Judge Prompt}
\label{app:cq-prompt}

The CQ judge evaluates the model's spoken content using the scoring rubric in Table~\ref{tab:cq-rubric} and the system prompt in Figure~\ref{fig:cq-system-prompt}.
For 20-minute episodes, the same prompt template is adapted to score 20 one-minute intervals.

The judge receives the dialogue grouped into one-minute intervals and returns a separate result for each interval (Figure~\ref{fig:cq-input-output}).
The input includes timestamps and indicates which intervals should be scored.
For utterances containing multiple sentences, we estimate each sentence's start and end times by dividing the utterance duration in proportion to sentence word counts.
Each sentence is included in every interval that overlaps its estimated time span.
For each scored interval, the output contains a rationale followed by an integer score; both fields are null for intervals not selected for scoring.

\subsection{Human Validation of the CQ Judge}
\label{app:cq-human-validation}

\begin{table}[tp]
\caption{\textbf{Human validation of the CQ judge.} Per-minute mean of human ratings and per-minute judge scores in the baseline sessions. MAE and Pearson's $r$ are computed over the ten per-minute values.}
\label{tab:cq-human-validation}

\centering
\small
\setlength{\tabcolsep}{3pt}

\begin{tabular*}{\linewidth}{@{\extracolsep{\fill}}llrrrrrrrrrrrr@{}}
\toprule
Model & Rater & 0--1 & 1--2 & 2--3 & 3--4 & 4--5 & 5--6 & 6--7 & 7--8 & 8--9 & 9--10 & MAE $\downarrow$ & $r$ $\uparrow$ \\
\midrule
\multirow{2}{*}{PersonaPlex}
  & Human & 4.27 & 3.87 & 4.07 & 3.87 & 3.40 & 2.73 & 2.20 & 2.07 & 1.60 & 1.17 & \multirow{2}{*}{0.351} & \multirow{2}{*}{0.99} \\
  & Judge & 4.90 & 4.67 & 4.41 & 4.22 & 3.64 & 3.16 & 2.66 & 2.05 & 1.45 & 1.08 & & \\
\midrule
\multirow{2}{*}{MiniCPM-o 4.5}
  & Human & 4.40 & 4.57 & 4.54 & 4.83 & 4.83 & 4.64 & 4.73 & 4.67 & 4.43 & 4.75 & \multirow{2}{*}{0.314} & \multirow{2}{*}{$-$0.19} \\
  & Judge & 4.83 & 4.84 & 4.56 & 4.67 & 4.45 & 4.45 & 4.28 & 4.13 & 4.18 & 4.30 & & \\
\midrule
\multirow{2}{*}{Raon-SpeechChat}
  & Human & 4.13 & 3.73 & 3.60 & 3.60 & 3.80 & 3.53 & 3.67 & 3.47 & 1.60 & 1.20 & \multirow{2}{*}{0.442} & \multirow{2}{*}{0.98} \\
  & Judge & 4.62 & 4.42 & 4.25 & 4.19 & 4.14 & 3.99 & 3.96 & 3.60 & 2.20 & 1.03 & & \\
\bottomrule
\end{tabular*}
\end{table}

We compared the CQ judge with human ratings on the baseline sessions of Section~\ref{sec:long-session-results}. Five annotators each rated three randomly assigned episodes per model, giving 15 rated episodes per model. Following the judge procedure (Appendix~\ref{app:cq-evaluation}), each annotator read the full timestamped transcript of an episode and scored every one-minute interval from 1 to 5 under the same rubric (Appendix~\ref{app:cq-prompt}). Table~\ref{tab:cq-human-validation} compares the per-minute mean of the human ratings with the per-minute judge scores in Appendix~\ref{app:long-session-results}.

The human ratings follow the same trends as the judge. For PersonaPlex and Raon-SpeechChat, both decline over the session ($r=0.99$ and $0.98$), whereas for MiniCPM-o 4.5, both remain above 4 throughout; because both stay within a narrow range near the top of the scale, the correlation is less informative for this model ($r=-0.19$). The two also agree in level, with an MAE below 0.5 on the 1--5 scale for all three models.

\subsection{Ten-Minute Baseline Results}
\label{app:long-session-results}

Table~\ref{tab:long-session-baseline} reports the per-minute estimates and corresponding sample counts for the four baseline models evaluated in Figure~\ref{fig:long-session-results}.

\subsection{Qualitative Failure Cases}
\label{app:qualitative-failures}

Table~\ref{tab:qualitative-failures} illustrates response timeouts, repetitive replies, and native-text--speech disagreement in selected baseline episodes.
These examples illustrate how the degradation patterns described in Section~\ref{sec:long-session-results} manifest in individual conversations.
Behind the rise in WER, the relation between the native text and the speech breaks down in different ways across models.
The native text of MiniCPM-o 4.5 increasingly runs ahead of the speech actually produced, whereas Raon-SpeechChat sometimes speaks content that differs altogether from its native text, as in the third example of Table~\ref{tab:qualitative-failures}.
In PersonaPlex, and at times in Raon-SpeechChat as in the last example, the speech itself breaks down in articulation and pronunciation.

\subsection{Twenty-Minute Long-Session Results}
\label{app:long-session-20m-results}

Table~\ref{tab:long-session-20m-results} reports per-minute results for Ours (LoRA) and Ours (TRC) in 20-minute sessions, with 90 episodes per model and condition.

\section{Method Details}
\begin{figure}[tp]
\centering
\includegraphics[width=\linewidth]{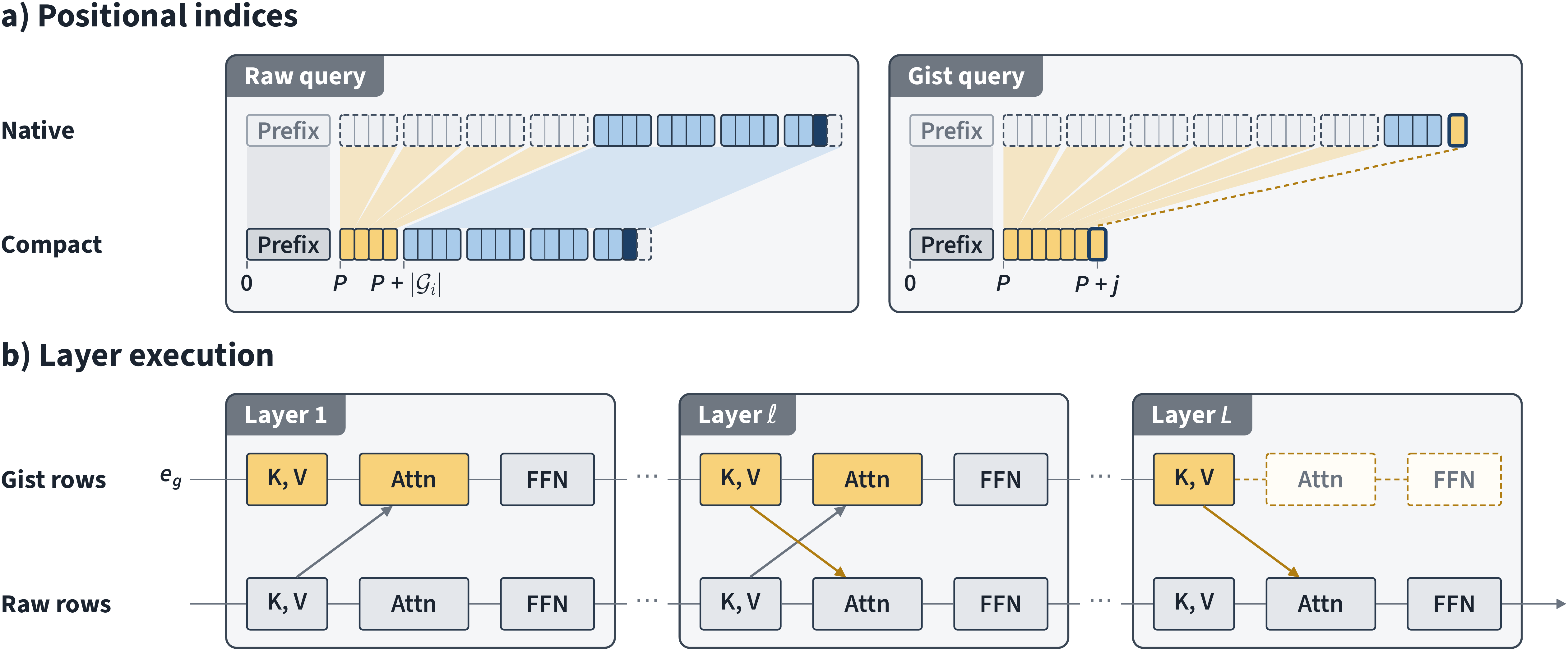}
\caption{\textbf{Positional indices and layer dependencies.}
(a) Native and compact indices for a raw query and a gist query, using the illustrative configuration of Figure~\ref{fig:method-overview}b.
(b) Computation in the first, intermediate, and last layers. Dashed blocks indicate omitted operations, and arrows indicate KV access between raw and gist tokens.}
\label{fig:layer-position}
\end{figure}

\begin{table}[tp]
\caption{Model-specific temporal units and TRC window and tile sizes.}
\label{tab:model-trc-settings}
\centering\small
\setlength{\tabcolsep}{4pt}
\renewcommand{\arraystretch}{1.15}
\begin{tabular}{@{}lcccc@{}}
\toprule
Model & Temporal unit & Positions per unit & Local window $H$ & Tile $\Delta$ \\
\midrule
PersonaPlex & 80\,ms frame & 1 & 120 frames (9.6\,s) & 12 frames (0.96\,s) \\
\addlinespace
MiniCPM-o 4.5 & 1\,s tick & 13--32 & 10 ticks (10\,s) & 1 tick (1\,s) \\
\addlinespace
Raon-SpeechChat & 80\,ms frame & 2--3 & 120 frames (9.6\,s) & 12 frames (0.96\,s) \\
\bottomrule
\end{tabular}
\end{table}

\begin{table}[tp]
\caption{\textbf{Teacher trajectory quality.} Each cell reports 0--2 $\rightarrow$ 8--10\,min performance of the base models in the diagnostic simulation and of the teacher trajectories.}
\label{tab:teacher-quality}

\centering
\small
\setlength{\tabcolsep}{3pt}

\begin{tabular*}{\linewidth}{
@{\extracolsep{\fill}}
l
@{\hspace{4pt}}
r@{\,}c@{\,}r
@{\hspace{7pt}}
r@{\,}c@{\,}r
r@{\,}c@{\,}r
@{\hspace{7pt}}
r@{\,}c@{\,}r
@{}
}
\specialrule{\heavyrulewidth}{0pt}{\aboverulesep}
Source
& \multicolumn{3}{c}{FTO (s)}
& \multicolumn{3}{c}{CQ $\uparrow$}
& \multicolumn{3}{c}{WER (\%) $\downarrow$}
& \multicolumn{3}{c}{Spk sim $\uparrow$}
\\
\specialrule{\lightrulewidth}{\aboverulesep}{0pt}

\rowcolor{modelgray}
\multicolumn{13}{@{}l}{\raisebox{0pt}[\dimexpr\arraystretch\ht\strutbox+1.5pt\relax][\dimexpr\arraystretch\dp\strutbox+1.5pt\relax]{\textbf{PersonaPlex}}} \\
\addlinespace[2pt]
\quad Base
& $-$0.27 & $\rightarrow$ & 0.40
& 4.78  & $\rightarrow$ & 1.29
& 5.4   & $\rightarrow$ & 44.3
& 0.733 & $\rightarrow$ & 0.337 \\
\quad Teacher
& $-$0.22 & $\rightarrow$ & $-$0.23
& 4.70  & $\rightarrow$ & 4.08
& 5.2   & $\rightarrow$ & 6.3
& 0.729 & $\rightarrow$ & 0.604 \\

\addlinespace[3pt]
\rowcolor{modelgray}
\multicolumn{13}{@{}l}{\raisebox{0pt}[\dimexpr\arraystretch\ht\strutbox+1.5pt\relax][\dimexpr\arraystretch\dp\strutbox+1.5pt\relax]{\textbf{MiniCPM-o 4.5}}} \\
\addlinespace[2pt]
\quad Base
& 0.40  & $\rightarrow$ & 1.90
& 4.81  & $\rightarrow$ & 4.19
& 2.6   & $\rightarrow$ & 13.6
& 0.775 & $\rightarrow$ & 0.749 \\
\quad Teacher
& 0.57  & $\rightarrow$ & 0.82
& 4.85  & $\rightarrow$ & 4.48
& 3.4   & $\rightarrow$ & 4.1
& 0.777 & $\rightarrow$ & 0.776 \\

\addlinespace[3pt]
\rowcolor{modelgray}
\multicolumn{13}{@{}l}{\raisebox{0pt}[\dimexpr\arraystretch\ht\strutbox+1.5pt\relax][\dimexpr\arraystretch\dp\strutbox+1.5pt\relax]{\textbf{Raon-SpeechChat}}} \\
\addlinespace[2pt]
\quad Base
& $-$0.07 & $\rightarrow$ & 0.06
& 4.52  & $\rightarrow$ & 1.63
& 11.1  & $\rightarrow$ & 70.5
& 0.816 & $\rightarrow$ & 0.520 \\
\quad Teacher
& $-$0.06 & $\rightarrow$ & $-$0.04
& 4.50  & $\rightarrow$ & 4.47
& 9.7   & $\rightarrow$ & 10.5
& 0.824 & $\rightarrow$ & 0.812 \\

\bottomrule
\end{tabular*}
\end{table}

\begin{table}[tp]
\caption{Model-specific content--control assignment rules.}
\label{tab:control-supervision}
\centering\small
\renewcommand{\tabularxcolumn}[1]{p{#1}}
\renewcommand{\tabularxcolumn}[1]{m{#1}}
\begin{tabularx}{\linewidth}{@{}M{0.18\linewidth}YY@{}}
\toprule
Model & Content & Control \\
\midrule
PersonaPlex & Ordinary text in consecutive non-PAD frames. & PAD, EPAD, and predictions in the first non-PAD frame after PAD. \\
\arrayrulecolor{black!10}
\midrule[0.2pt]
\arrayrulecolor{black}
MiniCPM-o 4.5 & Nonspecial text predictions, including those following \textnormal{\textless\textbar speak\textbar\textgreater} within the same tick. & Special-token predictions: \textnormal{\textless\textbar listen\textbar\textgreater}, \textnormal{\textless\textbar speak\textbar\textgreater}, \textnormal{\textless\textbar chunk\_eos\textbar\textgreater}, \textnormal{\textless\textbar chunk\_tts\_eos\textbar\textgreater}, \textnormal{\textless\textbar turn\_eos\textbar\textgreater}. \\
\arrayrulecolor{black!10}
\midrule[0.2pt]
\arrayrulecolor{black}
Raon-SpeechChat & Nonspecial text-vocabulary tokens in consecutive non-SIL frames. & All other supervised predictions, including SIL, PAD, EPAD, and BC. \\
\bottomrule
\end{tabularx}
\end{table}

\begin{table}[tp]
\caption{Training configuration for Ours (TRC) and Ours (LoRA).}
\label{tab:training-settings}
\centering\small
\begin{tabular}{@{}lc@{}}
\toprule
Setting & Value\\
\midrule
LoRA parameters & $r=8$, $\alpha=8$\\
Batch size & 8 trajectories\\
Steps & 500\\
\arrayrulecolor{black!10}
\midrule[0.2pt]
\arrayrulecolor{black}
Optimizer & AdamW (weight decay: 0) \\
Learning rate & $10^{-4} \rightarrow 10^{-5}$ (Linear) \\
Gradient clipping & 1.0\\
\arrayrulecolor{black!10}
\midrule[0.2pt]
\arrayrulecolor{black}
Validation interval & 50 steps\\
\bottomrule
\end{tabular}
\end{table}

\begin{figure}[tp]
\centering
\input{codes/Appendix/B_Method/B.6_Summarizer_System_Prompt}
\input{codes/Appendix/B_Method/B.6_Summarizer_User_Message}
\caption{\textbf{System prompt and user message for dialogue summarization.} Braced placeholders indicate the dialogue to be summarized.}
\label{fig:trajectory-summarizer-prompts}
\end{figure}

\begin{figure}[tp]
\centering
\begin{DCLTextPromptBox}{Teacher: system prompt after a segment boundary}
\{base prompt\}\par
\vspace{\baselineskip}
Here's what we've talked about so far. Let's keep chatting:\par
\vspace{\baselineskip}
\{summary\}\par
\end{DCLTextPromptBox}
\caption{\textbf{Teacher system prompt after a segment boundary.} Braced placeholders indicate the base prompt and the summary of preceding dialogue.}
\label{fig:teacher-boundary-prompt}
\end{figure}

\begin{figure}[tp]
\centering
\input{codes/Appendix/B_Method/B.6_User_Simulator_System_Prompt}
\input{codes/Appendix/B_Method/B.6_User_Simulator_Dialogue_Section}
\caption{\textbf{User-simulator prompts for trajectory generation.} Braced placeholders abbreviate unchanged portions. In the system prompt, the revisiting rule replaces the repetition prohibition, and conversation-context rules are appended. In the user message, the summary and current-segment dialogue replace the dialogue history; before the first segment boundary, the summary is ``(conversation just started)''.}
\label{fig:trajectory-user-simulator-prompts}
\end{figure}

\subsection{Positional Encoding}
\label{app:context-computation}

We index prediction positions along the continuous dialogue trajectory by $i$.
Following Section~\ref{sec:trc}, the attention context $\mathcal{S}_i$ comprises the fixed prefix $\mathcal{P}$, gist tokens $\mathcal{G}_i$ representing tiles outside the local window, and recent raw context $\mathcal{W}_i$.
We apply rotary positional encoding (RoPE) using native and compact positional indices.
Attention to raw dialogue keys uses native absolute indices, preserving their original relative offsets.
In contrast, attention to prefix and gist keys uses compact indices that place these components consecutively before the local raw context.

For a raw query at position $i$ attending to a key indexed by $m$, let $a_i$ and $a_m$ denote their native indices, and $\hat a_i$ and $\hat a_m$ their compact counterparts.
Under compact indexing, the $P=|\mathcal{P}|$ prefix positions occupy $0,\ldots,P-1$, and gist $g_j$ occupies position $P+j$.
The raw query receives compact index $\hat a_i=P+|\mathcal{G}_i|+r_i$, where $r_i$ is its zero-based position within the local raw context.

The gist-query context $\mathcal{A}^{\mathrm{gist}}_j$ comprises the fixed prefix, all preceding gists, and the raw positions $\mathcal{T}_j$ of tile $j$.
The query computing $g_j$ uses native index $b_j$, immediately following the raw positions of tile $j$, and compact index $\hat b_j=P+j$.

Let $\vq_i$ and $\vq^{\mathrm{gist}}_j$ denote the unrotated raw and gist query vectors, respectively, and let $\vk_m$ denote the unrotated key vector.
Their unscaled attention scores are
\begin{subequations}
\begin{align}
s^{\mathrm{raw}}_{im}
&=
\begin{cases}
\left\langle
\mR_{a_i}\vq_i,
\mR_{a_m}\vk_m
\right\rangle,
& m\in\mathcal{W}_i,
\\[3pt]
\left\langle
\mR_{\hat a_i}\vq_i,
\mR_{\hat a_m}\vk_m
\right\rangle,
& m\in\mathcal{P}\cup\mathcal{G}_i,
\end{cases}
\\[6pt]
s^{\mathrm{gist}}_{jm}
&=
\begin{cases}
\left\langle
\mR_{b_j}\vq^{\mathrm{gist}}_j,
\mR_{a_m}\vk_m
\right\rangle,
& m\in\mathcal{T}_j,
\\[3pt]
\left\langle
\mR_{\hat b_j}\vq^{\mathrm{gist}}_j,
\mR_{\hat a_m}\vk_m
\right\rangle,
& m\in\mathcal{P}\cup\{g_\ell:0\leq\ell<j\},
\end{cases}
\end{align}
\end{subequations}
where $\mR_a$ denotes the rotary transform at position $a$.
For each query, the scores are scaled and jointly normalized over all visible keys in the corresponding context, $\mathcal{S}_i$ or $\mathcal{A}^{\mathrm{gist}}_j$.
Figure~\ref{fig:layer-position}a illustrates the positional assignments for both query types.

\subsection{Layer Computation}
\label{app:streaming}

Each gist token starts from the shared learnable embedding and is processed alongside raw tokens through the backbone layers.
At each layer, keys and values are computed from the normalized input states.
Thus, dialogue information acquired by a gist at one layer becomes available to raw queries through its keys and values at the next layer.

\clearpage
\begingroup
\footnotesize
\renewcommand{\tabularxcolumn}[1]{m{#1}}
\begin{longtable}{M{0.19\linewidth}M{0.73\linewidth}}
\caption{Persona instructions for teacher-trajectory generation (verbatim).}\label{tab:training-persona-prompts}\\
\toprule
Persona & Instruction \\
\midrule
\endfirsthead
\multicolumn{2}{l}{\tablename~\thetable\ (continued)}\\
\toprule
Persona & Instruction \\
\midrule
\endhead
\bottomrule
\endfoot
Exploratory & Use details from the conversation as starting points for new stories, perspectives, and questions. Let an earlier remark bring to mind a related experience, interest, or possibility, then develop that new direction. Share plausible personal details and preferences as you go. Respond naturally to the other speaker and make transitions gradual. Let this tendency emerge occasionally; not every turn needs to refer back to something, and you need not ask for specific details to be recalled. \\
\arrayrulecolor{black!10}
\midrule[0.2pt]
\arrayrulecolor{black}
Applicative & Bring earlier information into new situations as the conversation develops. Ask the other speaker to use previously discussed circumstances, preferences, or explanations when considering a new possibility, interpreting an experience, or offering a perspective. Make the earlier information relevant to the answer rather than asking for it to be recited. Use only information established in the conversation, including earlier parts. Share new details and respond naturally; not every turn needs to use the past, and no particular decision or conclusion is required. \\
\arrayrulecolor{black!10}
\midrule[0.2pt]
\arrayrulecolor{black}
Recollective & Occasionally return to a specific detail or discussion from earlier in the conversation and ask the other speaker to recall it. Choose something relevant to what you want to discuss next, without restating the answer in your question. Refer only to information actually established in the conversation, including earlier parts. Use the response to continue the exchange rather than grading it. Share new details and respond naturally between these moments; do not make every turn a recall question. \\
\arrayrulecolor{black!10}
\midrule[0.2pt]
\arrayrulecolor{black}
Experiential & Develop the conversation through plausible everyday experiences and observations. Describe concrete situations, how they felt, and what made them interesting, revealing details gradually in response to the other speaker. Invite a reaction or interpretation when useful, and let the exchange lead to another experience or question. Keep individual turns conversational rather than delivering a long story at once. There is no need to ask the other speaker to recall earlier information. \\
\arrayrulecolor{black!10}
\midrule[0.2pt]
\arrayrulecolor{black}
Deliberative & Enjoy exploring different perspectives on an idea or situation. Offer your own view, ask what makes another view appealing, and consider examples, tradeoffs, or exceptions as they arise. Respond thoughtfully without turning the exchange into an argument or requiring agreement. Mix questions with your own reactions and plausible preferences. Let the discussion develop naturally without requiring a decision, and do not make recalling earlier information a goal. \\
\end{longtable}
\endgroup

\FloatBarrier

Figure~\ref{fig:layer-position}b illustrates these layer dependencies.
In the first layer, gist inputs have not yet incorporated dialogue information, so we exclude their keys and values from raw-query attention.
Gist attention and feed-forward updates are performed in all layers except the last.
In the last layer, gist states supply keys and values to raw queries, but their own attention and feed-forward updates are omitted because their outputs would not be consumed.

\subsection{Model-Specific TRC Settings}
\label{app:attention-sink}

The temporal units and number of sequence positions per unit differ across models.
We therefore express the TRC window and tile sizes in each model's native temporal units, keeping their durations approximately equal across models (Table~\ref{tab:model-trc-settings}).

For PersonaPlex, the prefix contains the text and voice prompts, and we additionally retain the first dialogue frame to account for attention sinks.
For MiniCPM-o 4.5, the prefix contains only the system prompt with encoded reference audio, as we observed no pronounced attention-sink behavior in the dialogue.
For Raon-SpeechChat, the prefix contains the system prompt followed by audio-start tokens, and we additionally retain the first three dialogue frames to account for attention sinks.

\clearpage
\begingroup
\footnotesize
\renewcommand{\tabularxcolumn}[1]{m{#1}}
\begin{longtable}{M{0.19\linewidth}M{0.73\linewidth}}
\caption{Topic instructions for teacher-trajectory generation (verbatim). Topics marked with \(\dagger\) are held out for validation.}\label{tab:training-topic-prompts}\\
\toprule
Topic & Instruction \\
\midrule
\endfirsthead
\multicolumn{2}{l}{\tablename~\thetable\ (continued)}\\
\toprule
Topic & Instruction \\
\midrule
\endhead
\bottomrule
\endfoot
School memories & Talk about school life and memories of being a student. Explore memorable classes, teachers, classmates, school customs, and how your views of those experiences have changed. Share plausible experiences and invite general perspectives without assuming the other speaker attended school. \\
\arrayrulecolor{black!10}
\midrule[0.2pt]
\arrayrulecolor{black}
Work collaboration & Talk about working with other people. Explore communication, different working styles, sharing responsibilities, receiving feedback, and what makes a group comfortable or frustrating. Everyday experiences and hypothetical situations are both welcome; no project needs to be completed. \\
\arrayrulecolor{black!10}
\midrule[0.2pt]
\arrayrulecolor{black}
Friendship and contact & Talk about how friendships develop and change. Explore becoming comfortable with someone, staying in touch, reconnecting, and balancing different expectations. Share plausible experiences and preferences while inviting the other speaker's perspective rather than assuming personal friendships. \\
\arrayrulecolor{black!10}
\midrule[0.2pt]
\arrayrulecolor{black}
Family customs & Talk about the different customs and expectations found in families. Explore familiar sayings, ways of showing affection, shared responsibilities, and differences between generations. Personal examples and imagined situations can lead the discussion without requiring sensitive disclosures. \\
\arrayrulecolor{black!10}
\midrule[0.2pt]
\arrayrulecolor{black}
Neighborhood change & Talk about the character of neighborhoods and how places change over time. Explore familiar streets, local shops, neighbors, belonging, and reactions to change. Use plausible recollections and general observations without relying on current information about a named location. \\
\arrayrulecolor{black!10}
\midrule[0.2pt]
\arrayrulecolor{black}
Language and expression\textsuperscript{\(\dagger\)} & Talk about how people express themselves through language. Explore the nuances of words, humor, misunderstandings, accents, and expressions that are difficult to explain. Share examples and preferences without turning the conversation into a language lesson or a quiz. \\
\arrayrulecolor{black!10}
\midrule[0.2pt]
\arrayrulecolor{black}
Digital communication & Talk about communicating through messages, calls, and online spaces. Explore tone, response expectations, group conversations, self-expression, and what feels comfortable in different situations. Keep the discussion grounded in experiences and preferences rather than technical troubleshooting. \\
\arrayrulecolor{black!10}
\midrule[0.2pt]
\arrayrulecolor{black}
Animals and people & Talk about encounters with animals and the relationships people form with them. Explore curiosity, comfort or fear, memorable behavior, and different attitudes toward animals. Pet ownership is optional; observations and hypothetical encounters can also develop the conversation. \\
\arrayrulecolor{black!10}
\midrule[0.2pt]
\arrayrulecolor{black}
Shopping and advertising & Talk about how people choose everyday purchases and react to advertising. Explore first impressions, recommendations, packaging, expectations, and satisfaction or disappointment after a purchase. Discuss plausible examples without needing current prices, product availability, or brand expertise. \\
\arrayrulecolor{black!10}
\midrule[0.2pt]
\arrayrulecolor{black}
Public-space etiquette & Talk about sharing public spaces with other people. Explore noise, queues, personal space, small acts of consideration, and situations where people's expectations differ. Compare experiences and perspectives without needing to settle on a universal rule. \\
\arrayrulecolor{black!10}
\midrule[0.2pt]
\arrayrulecolor{black}
Music listening & Talk about the place of music in everyday life. Explore sounds and moods, how people discover music, shared listening, and memories associated with particular kinds of music. No knowledge of named artists, songs, or recent releases is required. \\
\arrayrulecolor{black!10}
\midrule[0.2pt]
\arrayrulecolor{black}
Clothing and expression & Talk about clothing and personal expression. Explore comfort, color, impressions, dressing for different settings, and how preferences change. Share plausible experiences and invite general perspectives without assuming the other speaker wears or owns clothing. \\
\arrayrulecolor{black!10}
\midrule[0.2pt]
\arrayrulecolor{black}
Transport experiences & Talk about the experience of getting around. Explore walking, shared transport, commuting, waiting, and the small encounters or observations that happen along the way. Compare what different ways of traveling feel like without planning a trip or relying on live transport information. \\
\arrayrulecolor{black!10}
\midrule[0.2pt]
\arrayrulecolor{black}
Weather and seasons & Talk about how weather and seasons shape experiences and preferences. Explore atmosphere, sensory details, memorable weather, and differences in what people enjoy. Use recollections and hypothetical situations rather than current forecasts. \\
\arrayrulecolor{black!10}
\midrule[0.2pt]
\arrayrulecolor{black}
Objects and memories & Talk about ordinary objects that acquire personal meaning. Explore where they came from, associations they carry, changing tastes, and why an object might matter differently to different people. Focus on stories and meaning rather than organizing possessions or deciding what to discard. \\
\arrayrulecolor{black!10}
\midrule[0.2pt]
\arrayrulecolor{black}
Gifts and giving & Talk about giving and receiving gifts. Explore thoughtfulness, surprises, different expectations, awkward moments, and why a simple gift can feel meaningful. Share plausible examples without needing to select a product or plan an event. \\
\arrayrulecolor{black!10}
\midrule[0.2pt]
\arrayrulecolor{black}
Mistakes and surprises & Talk about everyday mistakes, misunderstandings, and unexpected outcomes. Explore how people react, what makes an incident funny or uncomfortable, and how a different perspective changes its meaning. Use ordinary, plausible situations rather than requiring a dramatic personal disclosure. \\
\arrayrulecolor{black!10}
\midrule[0.2pt]
\arrayrulecolor{black}
First impressions & Talk about first impressions and getting to know someone. Explore what draws attention, when an impression changes, how trust develops, and the difference between feeling familiar and being understood. Examples and hypothetical situations can guide the exchange. \\
\arrayrulecolor{black!10}
\midrule[0.2pt]
\arrayrulecolor{black}
Competition and fairness\textsuperscript{\(\dagger\)} & Talk about competition, cooperation, and what feels fair. Explore friendly contests, different levels of experience, reactions to winning or losing, and whether rules feel appropriate in a situation. Keep examples accessible without requiring knowledge of a particular sport or game. \\
\arrayrulecolor{black!10}
\midrule[0.2pt]
\arrayrulecolor{black}
Everyday curiosity & Talk about ordinary things that make people curious about how the world works. Explore observations of sound, light, materials, or natural patterns, the questions they raise, and which explanations feel intuitive or surprising. Follow the exchange naturally without turning it into a quiz. \\
\end{longtable}
\endgroup

\FloatBarrier

\subsection{Teacher Trajectory Construction}
\label{app:teacher-trajectory}

\paragraph{Scenarios and split.}
The training scenarios combine five personas with 20 topics, all distinct from those used in the diagnostic simulation.
Three personas encourage the use or recall of earlier dialogue, while the other two emphasize sharing experiences and exploring different perspectives.
Two topics are held out for validation, and all trajectories derived from a given scenario remain in the same split.

\paragraph{Segments and privileged summaries.}
Each trajectory contains five segments.
Each segment begins with a user utterance and ends at the first point after 90\,s when the model has completed its turn and the next user response is pending, subject to a 150\,s limit.
After each model turn ends, the user-response delay is resampled uniformly from 0.8--1.5\,s, with the response onset aligned to the next processing step.
At each segment boundary, a new teacher session is initialized with the base prompt and an LLM-generated summary of the preceding dialogue.
Summaries are generated by the same LLM used for the user simulator.
The summarizer reads the preceding dialogue rather than earlier summaries and is instructed to retain conversational details and speaker attribution.
Summary length is guided by the prompt's instruction to write concise prose, without imposing a fixed length limit. The summarization prompt is provided in Appendix~\ref{app:method-prompts}.
The user simulator also conditions on this summary, together with the dialogue accumulated within the current segment.
A 2\,s silent interval precedes the new segment's first user utterance, during which the teacher is forced to listen.
Teacher dialogues use the decoding settings in Table~\ref{tab:long-session-decoding}, with greedy audio decoding for Raon-SpeechChat.

\paragraph{Trajectory branching.}
For each model, we generate one initial dialogue per scenario.
After segments two and four, each dialogue branches into two independently generated continuations that share the same persona, topic, and dialogue summary.
Following the segment termination and branching rules described above, we generated four trajectories per scenario, yielding 400 trajectories per model.
Of these, 360 were used for training and 40 for validation.
The resulting mean trajectory durations ranged from 8.8 to 9.6\,min across models.

\paragraph{Quality filtering.}
A segment is rejected if the model fails to respond within 3\,s after a user utterance ends, repeats a word sequence consecutively over a span of at least 20 words within one utterance, or fails to reach a segment boundary within 150\,s.
Each segment has at most three attempts.
On the final attempt, a speaking decision is forced once per user utterance if the model has produced neither speech nor text within 0.8\,s after the utterance ends.
Positions at which speaking decisions are forced are excluded from KL supervision.
If all three segment attempts fail, generation restarts from the beginning of the dialogue or the most recent branching point.
Up to five attempts are allowed from each starting point.

\paragraph{Trajectory quality.}
To verify that segmented generation yields reliable targets, we evaluate the teacher trajectories with the metrics of Section~\ref{sec:long-session-protocol} and compare them with the base models over the same span (Table~\ref{tab:teacher-quality}). We omit RT and TS because segments with response timeouts are rejected during filtering. We also exclude from FTO the response to the first user utterance of each segment. This response follows the forced listening interval at the segment boundary, and all losses are masked up to its onset (Appendix~\ref{app:training-details}), so the remaining responses reflect the part of the trajectories used for supervision. Within this span the teacher stays much closer to its early-session quality than the base models. At 8--10\,min, for example, the teacher trajectories of Raon-SpeechChat reach a CQ of 4.47 and a WER of 10.5\%, compared with 1.63 and 70.5\% for the base model.

\subsection{Self-Distillation and Optimization}
\label{app:training-details}

\paragraph{Aligned inputs and targets.}
For both Ours (TRC) and Ours (LoRA), the student receives the base prompt once and replays recorded native inputs without re-encoding dialogue audio.
These audio inputs are represented as codes for PersonaPlex and embeddings for MiniCPM-o 4.5 and Raon-SpeechChat.
Teacher prediction and audio-conditioning positions are aligned with the corresponding positions in the student's continuous trajectory.

\paragraph{Supervision.}
Table~\ref{tab:control-supervision} specifies the model-specific assignment rules for the content and control KL losses introduced in Section~\ref{sec:self-distillation}.
All three losses are masked from the start of each segment through the processing step containing the onset of the teacher's spoken response to the first user utterance.
Beyond this initial interval, KL supervision excludes positions with forced decisions, whereas hidden-state supervision covers all audio-conditioning positions.

\paragraph{Loss definitions.}
For each trajectory and each supervision group (content or control), let $\mathcal{I}$ denote the prediction positions assigned to that group after applying the supervision masks.
Using $q_i$ and $p_i$ from Section~\ref{sec:self-distillation}, the corresponding KL term in Equation~\ref{eq} is
\begin{equation}
\mathcal{L}_{\mathrm{KL}}(\mathcal{I})
=\frac{1}{|\mathcal{I}|}\sum_{i\in\mathcal{I}}
\KL\!\left(q_i\,\|\,p_i\right).
\end{equation}
Raon-SpeechChat restricts token choices using a state-dependent vocabulary mask during generation.
We apply the same mask to teacher and student logits when computing KL divergence.
Let $\mathcal{A}$ contain the audio-conditioning positions used for hidden-state supervision, and let $\vh_i^{\mathrm{T}},\vh_i^{\mathrm{S}}\in\mathbb{R}^{d}$ denote the corresponding teacher and student hidden states.
The hidden-state term is
\begin{equation}
\mathcal{L}_{\mathrm{hid}}
=\frac{1}{|\mathcal{A}|}\sum_{i\in\mathcal{A}}
\frac{\|\vh_i^{\mathrm{S}}-\vh_i^{\mathrm{T}}\|_2^2}
{\max(\|\vh_i^{\mathrm{T}}\|_2^2,d\epsilon)},
\qquad \epsilon=10^{-6}.
\end{equation}
We average the total loss across sampled trajectories during training and across all held-out trajectories during validation.

\paragraph{Optimization.}
During training, scenarios are visited in shuffled order, and one completed trajectory is sampled uniformly from each scenario.
Table~\ref{tab:training-settings} lists the training settings shared by Ours (TRC) and Ours (LoRA).
The shared gist embedding and LoRA adapters contain 8.4M trainable parameters for PersonaPlex and 7.7M for MiniCPM-o 4.5 and Raon-SpeechChat, approximately 0.1\% of each backbone Transformer.

\subsection{Teacher-Trajectory Prompts}
\label{app:method-prompts}

Figure~\ref{fig:trajectory-summarizer-prompts} presents the summarizer's system prompt and user message, which contains the preceding dialogue with each turn labeled ``User:'' or ``Model:''.
The summarizer uses a temperature of 0.3 and zero presence penalty.
Figure~\ref{fig:teacher-boundary-prompt} shows the teacher's system prompt after a segment boundary.
Figure~\ref{fig:trajectory-user-simulator-prompts} presents the trajectory-generation modifications to the user-simulator prompts in Appendix~\ref{app:long-session-scenarios}.
Tables~\ref{tab:training-persona-prompts} and~\ref{tab:training-topic-prompts} list the persona and topic instructions used for teacher-trajectory generation.

\section{Supplementary Evaluation and Analysis}
\subsection{Supervision Ablations}
\label{app:supervision-ablation}

\begin{table}[tp]
\centering
\small
\begin{minipage}[t]{0.42\linewidth}
\centering
\caption{\textbf{Hidden-state supervision.} PersonaPlex performance at 0--2 $\rightarrow$ 18--20\,min.}
\label{tab:ablation}
\setlength{\tabcolsep}{3pt}
\begin{tabular*}{\linewidth}{@{\extracolsep{\fill}}l r@{\,}c@{\,}r r@{\,}c@{\,}r@{}}
\toprule
Loss & \multicolumn{3}{c}{WER (\%) $\downarrow$} & \multicolumn{3}{c}{Spk sim $\uparrow$} \\
\midrule
Ours (TRC)  & 5.2 & $\rightarrow$ & 12.8 & 0.699 & $\rightarrow$ & 0.579 \\
$-$ hidden  & 5.3 & $\rightarrow$ & 19.8 & 0.680 & $\rightarrow$ & 0.418 \\
\bottomrule
\end{tabular*}
\end{minipage}\hfill
\begin{minipage}[t]{0.54\linewidth}
\centering
\caption{\textbf{Pooled KL.} Share of content positions in the training data and relative difference in validation content KL of the pooled variant from Ours (TRC) at each step (positive: higher).}
\label{tab:kl-pooling}
\setlength{\tabcolsep}{3pt}
\begin{tabular*}{\linewidth}{@{\extracolsep{\fill}}lcrrr@{}}
\toprule
\multirow{2}{*}{Model} & \multirow{2}{*}{Content (\%)} & \multicolumn{3}{c}{$\Delta$ content KL (\%)} \\
\cmidrule(l){3-5}
 & & 50 & 100 & 500 \\
\midrule
PersonaPlex     & 32 & 1.8  & 0.7  & $-$0.2 \\
MiniCPM-o 4.5   & 60 & $-$2.9 & $-$4.4 & $-$0.7 \\
Raon-SpeechChat & 17 & 10.0 & 8.5  & 0.7 \\
\bottomrule
\end{tabular*}
\end{minipage}
\end{table}

We examine two design choices of the objective in Equation~\ref{eq}: the hidden-state loss and the separation of the KL into content and control terms. Each ablation changes only the corresponding term and keeps the teacher trajectories, data split, TRC settings (Table~\ref{tab:model-trc-settings}), and training configuration (Table~\ref{tab:training-settings}) unchanged.

\paragraph{Hidden-state supervision.}
On PersonaPlex, we set $\lambda_{\mathrm{hid}}=0$ while keeping $\lambda_{\mathrm{cnt}}=\lambda_{\mathrm{ctl}}=1$. For each condition, we select the checkpoint with the lowest validation loss and evaluate it in 20-minute sessions following the protocol in Section~\ref{sec:long-session-protocol}. As Table~\ref{tab:ablation} shows, the two conditions perform similarly at 0--2\,min, but without the hidden-state loss, WER at 18--20\,min rises from 12.8\% to 19.8\% and speaker similarity falls from 0.579 to 0.418. Matching the hidden representations supplied to the audio decoder thus helps preserve speech--text and voice consistency over long sessions, beyond what the token-level KL terms provide.

\paragraph{Content and control KL.}
The share of content positions differs widely across models, from 17\% on Raon-SpeechChat to 60\% on MiniCPM-o 4.5 (Table~\ref{tab:kl-pooling}). If the KL were averaged over all positions, this share would determine the relative weight of the two groups, so we average them separately and weight them equally. To examine this choice, we compare Ours (TRC) with a variant that pools the KL over all positions, scaled by 2 to match the total weight of the two separate terms. Table~\ref{tab:kl-pooling} reports the relative difference in validation content KL of this variant from Ours (TRC). Pooling changes how fast the content KL decreases early in training, most notably on Raon-SpeechChat (8.5--10.0\% higher over the first 100 steps), but the two objectives converge to nearly the same point after sufficient training, likely because the hidden-state loss also constrains the student. The separation is therefore motivated by the differing shares of content positions and by its effect early in training rather than by a difference in final performance.

\subsection{KV Storage and Backbone Latency}
\label{app:cost-measurement}

\paragraph{Analytical KV growth.}
For a backbone with $L$ layers, $h_{\mathrm{kv}}$ key--value heads, and head dimension $d_h$, the KV storage per cached position is
\[
B_{\mathrm{KV}} = 2Lh_{\mathrm{kv}}d_hb,
\]
where $b$ is the number of bytes per scalar. With BF16 storage, $B_{\mathrm{KV}}$ is 512\,KiB for PersonaPlex and 144\,KiB for both MiniCPM-o 4.5 and Raon-SpeechChat. TRC adds one cached gist position per completed tile while maintaining a fixed prefix and a bounded local raw window. Its long-term KV growth rate is therefore $60B_{\mathrm{KV}}/\Delta$ bytes per minute, where $\Delta$ is the tile duration in seconds. For Base, the minimum raw-position growth rates are 750, 780, and 1,500 positions per minute for PersonaPlex, MiniCPM-o 4.5, and Raon-SpeechChat, respectively. Multiplying these rates by $B_{\mathrm{KV}}$ gives the corresponding minimum KV growth rates. These analytical estimates cover KV tensor storage only, excluding memory allocation overhead and other model components. For PersonaPlex, the Base growth rate applies only before its 3,000-position KV ring saturates. Table~\ref{tab:trc-cost} reports the resulting KV growth rates for Base and TRC.

\paragraph{Backbone latency.}
We measure wall-clock backbone latency at batch size 1 in BF16 on one NVIDIA L40 GPU. Base uses the publicly released backbone implementation, while TRC adds its attention pattern and gist-token computation to the same backbone without loading a trained TRC checkpoint. We use random hidden-state inputs and synthetic KV caches with lengths corresponding to 1, 5, 10, and 20 minutes of interaction. Raw inputs follow each model's minimum position increment, with one position per 80\,ms frame for PersonaPlex, two per 80\,ms frame for Raon-SpeechChat, and 13 per 1\,s tick for MiniCPM-o 4.5. Timings include backbone forward passes, KV-cache updates, and gist-token computation. Cache preparation and restoration, audio modules, output heads, sampling, and worker scheduling are excluded.

For each context length and frame type, we perform two measurement rounds. Before every execution, we restore the corresponding cache to its initial state to keep the workload fixed. Each round begins with warm-up, which continues until Base and TRC have each completed at least 100 executions and their combined warm-up time reaches at least 30\,s (90\,s for the initial warm-up). We then perform 100 repetitions, measuring Base and TRC once each in random order to reduce the effects of temporal variation in GPU performance. We combine the timing samples according to how often frames with and without gist-token computation occur during inference. PersonaPlex and Raon-SpeechChat compute a gist token once every 12 frames, whereas MiniCPM-o 4.5 does so every tick. Table~\ref{tab:trc-cost} reports the resulting mean latency at each context length and the range of p95 latencies across the four context lengths.

\subsection{Local-Context Ablations}
\label{app:local-context-ablation}

\begin{table}[tp]
\caption{\textbf{Local-context variants.} Each cell reports 8--10 $\rightarrow$ 18--20\,min performance. A dash indicates an interval beyond the ten-minute Base sessions or a sample size that does not meet the minimum requirement for reporting specified in Appendix~\ref{app:validity-criteria}.}
\label{tab:local-context-ablation}

\centering
\small
\setlength{\tabcolsep}{3pt}

\begin{tabular*}{\linewidth}{
@{\extracolsep{\fill}}
l
@{\hspace{4pt}}
r@{\,}c@{\,}r
r@{\,}c@{\,}r
r@{\,}c@{\,}r
@{\hspace{7pt}}
r@{\,}c@{\,}r
r@{\,}c@{\,}r
@{\hspace{7pt}}
r@{\,}c@{\,}r
@{}
}
\specialrule{\heavyrulewidth}{0pt}{\aboverulesep}

\multirow{2}{*}{\textbf{Method}}
& \multicolumn{9}{c}{\small\textit{Responsiveness}}
& \multicolumn{6}{c}{\small\textit{Spoken Content}}
& \multicolumn{3}{c}{\small\textit{Voice}}
\\[-2pt]

\cmidrule(lr){2-10}
\cmidrule(lr){11-16}
\cmidrule(lr){17-19}

&
\multicolumn{3}{c}{RT (\%) $\downarrow$}
& \multicolumn{3}{c}{TS (\%) $\downarrow$}
& \multicolumn{3}{c}{FTO (s)}
& \multicolumn{3}{c}{CQ $\uparrow$}
& \multicolumn{3}{c}{WER (\%) $\downarrow$}
& \multicolumn{3}{c}{Spk sim $\uparrow$}
\\
\specialrule{\lightrulewidth}{\aboverulesep}{0pt}

\rowcolor{modelgray}
\multicolumn{19}{@{}l}{\raisebox{0pt}[\dimexpr\arraystretch\ht\strutbox+1.5pt\relax][\dimexpr\arraystretch\dp\strutbox+1.5pt\relax]{\textbf{PersonaPlex}}} \\
\addlinespace[2pt]
\quad Base
& 17.6  & $\rightarrow$ & \NA
& 18.9  & $\rightarrow$ & \NA
& 0.40  & $\rightarrow$ & \NA
& 1.29  & $\rightarrow$ & \NA
& 44.3  & $\rightarrow$ & \NA
& 0.337 & $\rightarrow$ & \NA \\
\arrayrulecolor{black!10}
\midrule[0.2pt]
\arrayrulecolor{black}
\quad StreamingLLM & \multicolumn{18}{l}{} \\
\qquad 4 sinks
& 0.8   & $\rightarrow$ & 0.5
& 0.0   & $\rightarrow$ & 0.0
& $-$0.26 & $\rightarrow$ & $-$0.27
& 4.51  & $\rightarrow$ & 4.29
& 5.3   & $\rightarrow$ & 6.9
& 0.488 & $\rightarrow$ & 0.475 \\
\qquad TRC prefix
& 0.0   & $\rightarrow$ & 0.3
& 0.0   & $\rightarrow$ & 0.0
& $-$0.25 & $\rightarrow$ & $-$0.24
& 4.28  & $\rightarrow$ & 4.21
& 5.8   & $\rightarrow$ & 6.8
& 0.648 & $\rightarrow$ & 0.642 \\
\arrayrulecolor{black!10}
\midrule[0.2pt]
\arrayrulecolor{black}
\quad Ours & \multicolumn{18}{l}{} \\
\qquad Window
& 0.4   & $\rightarrow$ & 0.0
& 0.0   & $\rightarrow$ & 0.0
& $-$0.22 & $\rightarrow$ & $-$0.22
& 4.17  & $\rightarrow$ & 3.98
& 7.7   & $\rightarrow$ & 8.5
& 0.649 & $\rightarrow$ & 0.649 \\
\qquad TRC w/o gists
& 0.7   & $\rightarrow$ & 1.3
& 0.0   & $\rightarrow$ & 0.0
& $-$0.22 & $\rightarrow$ & $-$0.21
& 4.44  & $\rightarrow$ & 4.23
& 5.6   & $\rightarrow$ & 5.3
& 0.683 & $\rightarrow$ & 0.681 \\
\qquad TRC
& 0.4   & $\rightarrow$ & 1.4
& 0.0   & $\rightarrow$ & 4.4
& $-$0.23 & $\rightarrow$ & $-$0.26
& 3.84  & $\rightarrow$ & 3.53
& 13.3  & $\rightarrow$ & 12.8
& 0.617 & $\rightarrow$ & 0.579 \\

\addlinespace[3pt]
\rowcolor{modelgray}
\multicolumn{19}{@{}l}{\raisebox{0pt}[\dimexpr\arraystretch\ht\strutbox+1.5pt\relax][\dimexpr\arraystretch\dp\strutbox+1.5pt\relax]{\textbf{MiniCPM-o 4.5}}} \\
\addlinespace[2pt]
\quad Base
& 66.7  & $\rightarrow$ & \NA
& 66.7  & $\rightarrow$ & \NA
& 1.90  & $\rightarrow$ & \NA
& 4.19  & $\rightarrow$ & \NA
& 13.6  & $\rightarrow$ & \NA
& 0.749 & $\rightarrow$ & \NA \\
\arrayrulecolor{black!10}
\midrule[0.2pt]
\arrayrulecolor{black}
\quad StreamingLLM & \multicolumn{18}{l}{} \\
\qquad 4 sinks
& 34.1  & $\rightarrow$ & 46.2
& 2.2   & $\rightarrow$ & 7.8
& 2.42  & $\rightarrow$ & \NA
& 4.17  & $\rightarrow$ & 4.28
& 12.2  & $\rightarrow$ & 36.7
& 0.719 & $\rightarrow$ & 0.718 \\
\qquad TRC prefix
& 30.5  & $\rightarrow$ & 38.2
& 0.0   & $\rightarrow$ & 0.0
& 2.29  & $\rightarrow$ & \NA
& 4.07  & $\rightarrow$ & 4.20
& 20.4  & $\rightarrow$ & 57.1
& 0.773 & $\rightarrow$ & 0.776 \\
\arrayrulecolor{black!10}
\midrule[0.2pt]
\arrayrulecolor{black}
\quad Ours & \multicolumn{18}{l}{} \\
\qquad Window
& 24.3  & $\rightarrow$ & 34.8
& 31.1  & $\rightarrow$ & 44.4
& 1.60  & $\rightarrow$ & 1.71
& 4.45  & $\rightarrow$ & 4.44
& 5.2   & $\rightarrow$ & 4.2
& 0.766 & $\rightarrow$ & 0.771 \\
\qquad TRC w/o gists
& 3.4   & $\rightarrow$ & 12.8
& 0.0   & $\rightarrow$ & 0.0
& 1.14  & $\rightarrow$ & 1.43
& 4.60  & $\rightarrow$ & 4.40
& 3.7   & $\rightarrow$ & 5.3
& 0.778 & $\rightarrow$ & 0.769 \\
\qquad TRC
& 34.6  & $\rightarrow$ & 51.6
& 45.6  & $\rightarrow$ & 76.7
& 1.87  & $\rightarrow$ & \NA
& 4.35  & $\rightarrow$ & 4.64
& 5.3   & $\rightarrow$ & 5.8
& 0.771 & $\rightarrow$ & 0.770 \\

\addlinespace[3pt]
\rowcolor{modelgray}
\multicolumn{19}{@{}l}{\raisebox{0pt}[\dimexpr\arraystretch\ht\strutbox+1.5pt\relax][\dimexpr\arraystretch\dp\strutbox+1.5pt\relax]{\textbf{Raon-SpeechChat}}} \\
\addlinespace[2pt]
\quad Base
& 27.5  & $\rightarrow$ & \NA
& 30.0  & $\rightarrow$ & \NA
& 0.06  & $\rightarrow$ & \NA
& 1.63  & $\rightarrow$ & \NA
& 70.5  & $\rightarrow$ & \NA
& 0.520 & $\rightarrow$ & \NA \\
\arrayrulecolor{black!10}
\midrule[0.2pt]
\arrayrulecolor{black}
\quad StreamingLLM & \multicolumn{18}{l}{} \\
\qquad 4 sinks
& 80.0  & $\rightarrow$ & 0.0
& 98.9  & $\rightarrow$ & 98.9
& \NA   & $\rightarrow$ & \NA
& \NA   & $\rightarrow$ & \NA
& \NA   & $\rightarrow$ & \NA
& \NA   & $\rightarrow$ & \NA \\
\qquad TRC prefix
& 0.0   & $\rightarrow$ & 0.5
& 0.0   & $\rightarrow$ & 0.0
& $-$0.11 & $\rightarrow$ & $-$0.06
& 3.76  & $\rightarrow$ & 2.98
& 13.5  & $\rightarrow$ & 43.4
& 0.809 & $\rightarrow$ & 0.784 \\
\arrayrulecolor{black!10}
\midrule[0.2pt]
\arrayrulecolor{black}
\quad Ours & \multicolumn{18}{l}{} \\
\qquad Window
& 2.4   & $\rightarrow$ & 2.6
& 11.1  & $\rightarrow$ & 16.7
& 0.04  & $\rightarrow$ & 0.00
& 4.10  & $\rightarrow$ & 3.24
& 17.3  & $\rightarrow$ & 43.9
& 0.798 & $\rightarrow$ & 0.758 \\
\qquad TRC w/o gists
& 0.2   & $\rightarrow$ & 0.6
& 3.3   & $\rightarrow$ & 4.4
& $-$0.03 & $\rightarrow$ & $-$0.01
& 4.13  & $\rightarrow$ & 3.19
& 14.9  & $\rightarrow$ & 39.0
& 0.805 & $\rightarrow$ & 0.775 \\
\qquad TRC
& 1.1   & $\rightarrow$ & 1.0
& 3.3   & $\rightarrow$ & 3.3
& $-$0.04 & $\rightarrow$ & $-$0.01
& 4.19  & $\rightarrow$ & 3.29
& 15.3  & $\rightarrow$ & 44.3
& 0.802 & $\rightarrow$ & 0.767 \\

\bottomrule
\end{tabular*}
\end{table}

\paragraph{Variants.}
To identify the source of long-session degradation and what is essential to mitigating it, we compare four variants that restrict the model to a local context: a fixed prefix followed by the same recent raw window as in TRC (Table~\ref{tab:model-trc-settings}), with all gists and older raw tokens evicted.
Two variants follow StreamingLLM~\citep{xiao2024streamingllm} without training and differ only in the prefix. StreamingLLM (4 sinks) keeps the first four positions, as in the original method, whereas StreamingLLM (TRC prefix) keeps the prefix of TRC, which consists of the full prompt and, for some models, the first dialogue frames retained as attention sinks (Appendix~\ref{app:attention-sink}).
The other two variants use the TRC prefix and are trained with our self-distillation. Ours (Window) is trained with this local context, whereas Ours (TRC) w/o gists is the trained Ours (TRC) model evaluated with all gists masked, so that it sees the same local context at inference.
All variants evict raw tokens on the same tile grid as TRC and follow its positional encoding without gist positions (Appendix~\ref{app:context-computation}). Ours (Window) shares the training data, objective, LoRA configuration, and checkpoint selection of Ours (TRC) (Appendix~\ref{app:training-details}).
We evaluate all variants in 20-minute sessions under the protocol of Section~\ref{sec:long-session-protocol}, and Table~\ref{tab:local-context-ablation} reports the results.

\paragraph{Results.}
We first compare the untrained variants with Base to examine the role of eviction. Eviction alone substantially mitigates the dominant degradation of each model, suggesting that the degradation arises largely from the accumulating raw context and that evicting this context is the most direct way to avoid it.
At 8--10\,min, StreamingLLM (TRC prefix) lowers WER from 44.3\% to 5.8\% on PersonaPlex and from 70.5\% to 13.5\% on Raon-SpeechChat, raising CQ from 1.29 to 4.28 and from 1.63 to 3.76, respectively, and on MiniCPM-o 4.5, it terminates no session, whereas Base terminates 66.7\%.
How well eviction works, however, depends on what the prefix retains. With only the first four positions kept, the speaker similarity of PersonaPlex stays well below that with the TRC prefix (0.488 vs.\ 0.648), and Raon-SpeechChat collapses almost completely (TS 98.9\%), suggesting that the prefix should retain the prompt and the positions each model relies on as attention sinks.
Even with a suitable prefix, eviction still leaves part of the degradation unresolved, most clearly on MiniCPM-o 4.5: its RT remains at 30.5\%, and its FTO rises from 0.19\,s at 0--2\,min to 2.29\,s at 8--10\,min, exceeding even that of Base (1.90\,s). Because this variant no longer attends to the accumulated context, this rise is unlikely to come from reading out that context; rather, some form of drift appears to build up in the model's hidden states over the session.

Eviction thus accounts for much of the mitigation, but not all of it: training with TRC further reduces the remaining degradation, and this gain goes beyond the effect of additional training. Ours (TRC) w/o gists sees the same local context as StreamingLLM (TRC prefix) at inference, yet it raises CQ in all three models and substantially reduces the remaining decline on MiniCPM-o 4.5, lowering RT from 30.5\% to 3.4\%, FTO at 8--10\,min from 2.29 to 1.14\,s, and WER from 20.4\% to 3.7\% without terminating any session; these gains largely hold through 20 minutes.
Ours (Window), in contrast, receives the same self-distillation and sees the same local context at inference, but its effects are more limited and less consistent: it brings no clear improvement on PersonaPlex, and on MiniCPM-o 4.5 and Raon-SpeechChat, its gains come with more terminated sessions (31.1\% and 11.1\% by 10\,min, vs.\ none for StreamingLLM (TRC prefix)).
Because the two trained variants differ only in whether TRC is used during training, the further gains can be attributed to combining self-distillation with TRC rather than to training alone. This suggests that TRC serves not only as a memory at inference but also as a training structure under which self-distillation stabilizes the model.

\subsection{Single-Needle Retention}
\label{app:memory-recall}

\paragraph{Materials.}
We select 100 original question--answer pairs from distinct FictionalQA events~\citep{kirchenbauer2026fictionalqa}. Each episode uses the source fact as its single needle, the original question as its probe, and the natural answer as its reference. We exclude derivative questions and require an answer of one to three words without digits or an integer from 0 to 99. The answer must occur in the fact but not the question; facts and questions are limited to 30 and 25 words, respectively. The opening, fact, and question are synthesized once with Kokoro-82M and reused across conditions.

\paragraph{Evaluation procedure.}
After the needle, the main session receives only silence, with the model forced to listen throughout. Retention is evaluated at 0, 5, and 10 minutes from the end of the 20\,s input slot containing the needle. To prevent each probe from affecting subsequent evaluations, we run it on a separate copy of the model state at the corresponding evaluation point. Each probe presents the same question and lasts 20\,s, including both the question and response. The model is forced to listen until it has processed the entire question, including the processing step containing the end of the question audio. It is then allowed to respond. If no lexical text is generated within the next 2\,s, we force speaking for one processing step and then return control to the model. Both Base and TRC use greedy decoding. Scoring uses the model's native text output generated after the question is complete, rather than an ASR transcript of its speech.

\paragraph{Scoring and reporting.}
A response is correct if it contains the complete reference answer under the following matching rules. Responses and reference answers are Unicode-normalized and lowercased, with punctuation removed except for internal hyphens and apostrophes. Matching tolerates spaces introduced between streaming chunks (e.g., ``Lon don'' matches ``London''), but does not accept an answer occurring only within a longer word (e.g., ``art'' in ``start''). Integer answers accept both digits and English cardinal forms; no additional answer aliases are used. At each retention distance, accuracy is the proportion of probes answered correctly, with incorrect and unanswered probes counted as errors.

\begin{table}[tp]
\centering
\small
\caption{\textbf{Single-needle retention accuracy with greedy decoding.} Each entry is the percentage of correct responses among 100 scorable probes. For Raon-SpeechChat, TRC ($\Delta=480$\,ms) halves the tile duration of the default configuration with the same 9.6\,s local window; TRC ($H_{\mathrm{T}}=30$\,s) is described in Appendix~\ref{app:stability-recall}.}
\label{tab:single-needle-retention-greedy}

\begin{tabular}{@{}llccc@{}}
\toprule
Model & Condition & 0 min & 5 min & 10 min \\
\midrule
\multirow{2}{*}{PersonaPlex}
    & Base & 38 & 0 & 0 \\
    & TRC & 12 & 14 & 7 \\
\midrule
\multirow{3}{*}{MiniCPM-o 4.5}
    & Base & 85 & 86 & 87 \\
    & TRC & 53 & 57 & 46 \\
    & TRC ($H_{\mathrm{T}}=30$\,s) & 36 & 35 & 29 \\
\midrule
\multirow{3}{*}{Raon-SpeechChat}
    & Base & 76 & 74 & 11 \\
    & TRC & 22 & 14 & 11 \\
    & TRC ($\Delta=480$\,ms) & 43 & 29 & 34 \\
\bottomrule
\end{tabular}
\end{table}

\subsection{Teacher Window Ablation}
\label{app:stability-recall}

\begin{table}[tp]
\caption{\textbf{Teacher window on MiniCPM-o 4.5.} Each cell reports 0--2 $\rightarrow$ 8--10 $\rightarrow$ 18--20\,min performance, except TS at 10 $\rightarrow$ 20\,min, CQ at 18--20\,min, and recall at 5 / 10\,min. StreamingLLM retains no context beyond its local window, so its recall is set to 0. A dash indicates an interval beyond the ten-minute Base sessions or a sample size that does not meet the minimum requirement for reporting specified in Appendix~\ref{app:validity-criteria}.}
\label{tab:stability-recall}

\centering
\small
\setlength{\tabcolsep}{1pt}

\begin{tabular*}{\linewidth}{
@{\extracolsep{\fill}}
l
@{\hspace{5pt}}
r@{\,}c@{\,}r@{\,}c@{\,}r
@{\hspace{5pt}}
r@{\,}c@{\,}r
@{\hspace{5pt}}
r@{\,}c@{\,}r@{\,}c@{\,}r
@{\hspace{5pt}}
r@{\,}c@{\,}r@{\,}c@{\,}r
@{\hspace{5pt}}
c
@{\hspace{5pt}}
c
@{}
}
\toprule
Method
& \multicolumn{5}{c}{RT (\%) $\downarrow$}
& \multicolumn{3}{c}{TS (\%) $\downarrow$}
& \multicolumn{5}{c}{WER (\%) $\downarrow$}
& \multicolumn{5}{c}{FTO (s)}
& CQ $\uparrow$
& Recall (\%) $\uparrow$ \\
\midrule
Base
& 9.4  & $\rightarrow$ & 66.7 & $\rightarrow$ & \NA
& 66.7 & $\rightarrow$ & \NA
& 2.6  & $\rightarrow$ & 13.6 & $\rightarrow$ & \NA
& 0.4 & $\rightarrow$ & 1.9 & $\rightarrow$ & \NA
& --   & 86 / 87 \\
StreamingLLM
& 0.0  & $\rightarrow$ & 30.5 & $\rightarrow$ & 38.2
& 0.0  & $\rightarrow$ & 0.0
& 19.8 & $\rightarrow$ & 20.4 & $\rightarrow$ & 57.1
& 0.2 & $\rightarrow$ & 2.3 & $\rightarrow$ & \NA
& 4.20 & 0 / 0 \\
\arrayrulecolor{black!10}
\midrule[0.2pt]
\arrayrulecolor{black}
Ours (TRC) & \multicolumn{20}{l}{} \\
\quad $H_{\mathrm{T}}=150$\,s
& 15.4 & $\rightarrow$ & 34.6 & $\rightarrow$ & 51.6
& 45.6 & $\rightarrow$ & 76.7
& 4.4  & $\rightarrow$ & 5.3  & $\rightarrow$ & 5.8
& 0.6 & $\rightarrow$ & 1.9 & $\rightarrow$ & \NA
& 4.64 & 57 / 46 \\
\quad $H_{\mathrm{T}}=30$\,s
& 0.0  & $\rightarrow$ & 0.9  & $\rightarrow$ & 22.4
& 0.0  & $\rightarrow$ & 0.0
& 2.9  & $\rightarrow$ & 4.1  & $\rightarrow$ & 5.6
& 0.3 & $\rightarrow$ & 1.1 & $\rightarrow$ & 1.5
& 4.68 & 35 / 29 \\
\bottomrule
\end{tabular*}
\end{table}

\paragraph{Motivation and setup.}
In the local-context ablation (Appendix~\ref{app:local-context-ablation}), Ours (TRC) on MiniCPM-o 4.5 terminates far more sessions than the same model evaluated without gists. This suggests that accumulating gist tokens can contribute to degradation just as accumulating raw tokens do. We hypothesize that this stems in part from the context of the teacher. During self-distillation (Section~\ref{sec:self-distillation}), the teacher computes its targets from a summary of the preceding segments and up to $H_{\mathrm{T}}=150$\,s of raw context from the current segment (Appendix~\ref{app:teacher-trajectory}). Its targets therefore reflect raw tokens beyond the student's window, and the gists may learn to reproduce not only this information but also the long-context behavior that degrades the base model. To test this hypothesis, we reduce $H_{\mathrm{T}}$ to 30\,s on MiniCPM-o 4.5 and give all earlier dialogue to the teacher as a summary while keeping all other settings unchanged.

\paragraph{Results.}
Table~\ref{tab:stability-recall} compares this configuration with the default configuration, StreamingLLM with the TRC prefix (Appendix~\ref{app:local-context-ablation}), and Base. Shortening the teacher window makes MiniCPM-o 4.5 considerably more stable. No session terminates through 20 minutes, whereas 76.7\% do with the default configuration. RT at 18--20\,min falls from 51.6\% to 22.4\% while WER and CQ remain at a similar level. However, the same change lowers recall from 57\% to 35\% at 5 minutes and from 46\% to 29\% at 10 minutes. Taken together, these results suggest that the context of the teacher shapes in part what the gists carry. A longer window conveys more information from earlier dialogue but may also convey more of the long-context behavior that degrades the base model, which is in line with our hypothesis.

\subsection{Evaluation with User Backchannels and Interruptions}
\label{app:gpt-live}

\begin{table}[tp]
\caption{\textbf{Base models with our simulator and GPT-Live-1.} Each cell reports 0--2 $\rightarrow$ 8--10\,min performance. Values for our simulator are from Table~\ref{tab:long-session-stability}. A dash indicates a sample size that does not meet the minimum requirement for reporting specified in Appendix~\ref{app:validity-criteria}.}
\label{tab:gpt-live-base}

\centering
\small
\setlength{\tabcolsep}{3pt}

\begin{tabular*}{\linewidth}{
@{\extracolsep{\fill}}
l
@{\hspace{4pt}}
r@{\,}c@{\,}r
r@{\,}c@{\,}r
r@{\,}c@{\,}r
@{\hspace{7pt}}
r@{\,}c@{\,}r
r@{\,}c@{\,}r
@{\hspace{7pt}}
r@{\,}c@{\,}r
@{}
}
\specialrule{\heavyrulewidth}{0pt}{\aboverulesep}

\multirow{2}{*}{\textbf{User}}
& \multicolumn{9}{c}{\small\textit{Responsiveness}}
& \multicolumn{6}{c}{\small\textit{Spoken Content}}
& \multicolumn{3}{c}{\small\textit{Voice}}
\\[-2pt]

\cmidrule(lr){2-10}
\cmidrule(lr){11-16}
\cmidrule(lr){17-19}

&
\multicolumn{3}{c}{RT (\%) $\downarrow$}
& \multicolumn{3}{c}{TS (\%) $\downarrow$}
& \multicolumn{3}{c}{FTO (s)}
& \multicolumn{3}{c}{CQ $\uparrow$}
& \multicolumn{3}{c}{WER (\%) $\downarrow$}
& \multicolumn{3}{c}{Spk sim $\uparrow$}
\\
\specialrule{\lightrulewidth}{\aboverulesep}{0pt}

\rowcolor{modelgray}
\multicolumn{19}{@{}l}{\raisebox{0pt}[\dimexpr\arraystretch\ht\strutbox+1.5pt\relax][\dimexpr\arraystretch\dp\strutbox+1.5pt\relax]{\textbf{PersonaPlex}}} \\
\addlinespace[2pt]
\quad Our simulator
& 0.0   & $\rightarrow$ & 17.6
& 0.0   & $\rightarrow$ & 18.9
& $-$0.27 & $\rightarrow$ & 0.40
& 4.78  & $\rightarrow$ & 1.29
& 5.4   & $\rightarrow$ & 44.3
& 0.733 & $\rightarrow$ & 0.337 \\
\quad GPT-Live-1
& 3.8   & $\rightarrow$ & 35.8
& 0.0   & $\rightarrow$ & 36.7
& 0.20  & $\rightarrow$ & 0.45
& 4.60  & $\rightarrow$ & 1.16
& 7.1   & $\rightarrow$ & 53.1
& 0.720 & $\rightarrow$ & 0.224 \\

\addlinespace[3pt]
\rowcolor{modelgray}
\multicolumn{19}{@{}l}{\raisebox{0pt}[\dimexpr\arraystretch\ht\strutbox+1.5pt\relax][\dimexpr\arraystretch\dp\strutbox+1.5pt\relax]{\textbf{MiniCPM-o 4.5}}} \\
\addlinespace[2pt]
\quad Our simulator
& 9.4   & $\rightarrow$ & 66.7
& 6.7   & $\rightarrow$ & 66.7
& 0.40  & $\rightarrow$ & 1.90
& 4.81  & $\rightarrow$ & 4.19
& 2.6   & $\rightarrow$ & 13.6
& 0.775 & $\rightarrow$ & 0.749 \\
\quad GPT-Live-1
& 36.9  & $\rightarrow$ & 55.3
& 40.0  & $\rightarrow$ & 76.7
& 0.63  & $\rightarrow$ & \NA
& 4.52  & $\rightarrow$ & 4.75
& 6.0   & $\rightarrow$ & 8.9
& 0.764 & $\rightarrow$ & 0.735 \\

\bottomrule
\end{tabular*}
\end{table}

\begin{table}[tp]
\caption{\textbf{Mitigation with GPT-Live-1.} Each cell reports 8--10 $\rightarrow$ 18--20\,min performance. A dash indicates an interval beyond the ten-minute Base sessions or a sample size that does not meet the minimum requirement for reporting specified in Appendix~\ref{app:validity-criteria}.}
\label{tab:gpt-live-mitigation}

\centering
\small
\setlength{\tabcolsep}{3pt}

\begin{tabular*}{\linewidth}{
@{\extracolsep{\fill}}
l
@{\hspace{4pt}}
r@{\,}c@{\,}r
r@{\,}c@{\,}r
r@{\,}c@{\,}r
@{\hspace{7pt}}
r@{\,}c@{\,}r
r@{\,}c@{\,}r
@{\hspace{7pt}}
r@{\,}c@{\,}r
@{}
}
\specialrule{\heavyrulewidth}{0pt}{\aboverulesep}

\multirow{2}{*}{\textbf{Method}}
& \multicolumn{9}{c}{\small\textit{Responsiveness}}
& \multicolumn{6}{c}{\small\textit{Spoken Content}}
& \multicolumn{3}{c}{\small\textit{Voice}}
\\[-2pt]

\cmidrule(lr){2-10}
\cmidrule(lr){11-16}
\cmidrule(lr){17-19}

&
\multicolumn{3}{c}{RT (\%) $\downarrow$}
& \multicolumn{3}{c}{TS (\%) $\downarrow$}
& \multicolumn{3}{c}{FTO (s)}
& \multicolumn{3}{c}{CQ $\uparrow$}
& \multicolumn{3}{c}{WER (\%) $\downarrow$}
& \multicolumn{3}{c}{Spk sim $\uparrow$}
\\
\specialrule{\lightrulewidth}{\aboverulesep}{0pt}

\rowcolor{modelgray}
\multicolumn{19}{@{}l}{\raisebox{0pt}[\dimexpr\arraystretch\ht\strutbox+1.5pt\relax][\dimexpr\arraystretch\dp\strutbox+1.5pt\relax]{\textbf{PersonaPlex}}} \\
\addlinespace[2pt]
\quad Base
& 35.8  & $\rightarrow$ & \NA
& 36.7  & $\rightarrow$ & \NA
& 0.45  & $\rightarrow$ & \NA
& 1.16  & $\rightarrow$ & \NA
& 53.1  & $\rightarrow$ & \NA
& 0.224 & $\rightarrow$ & \NA \\
\quad Ours (TRC)
& 12.6  & $\rightarrow$ & 8.7
& 3.3   & $\rightarrow$ & 10.0
& 0.23  & $\rightarrow$ & 0.23
& 3.91  & $\rightarrow$ & 3.28
& 11.7  & $\rightarrow$ & 18.2
& 0.599 & $\rightarrow$ & 0.537 \\

\addlinespace[3pt]
\rowcolor{modelgray}
\multicolumn{19}{@{}l}{\raisebox{0pt}[\dimexpr\arraystretch\ht\strutbox+1.5pt\relax][\dimexpr\arraystretch\dp\strutbox+1.5pt\relax]{\textbf{MiniCPM-o 4.5}}} \\
\addlinespace[2pt]
\quad Base
& 55.3  & $\rightarrow$ & \NA
& 76.7  & $\rightarrow$ & \NA
& \NA   & $\rightarrow$ & \NA
& 4.75  & $\rightarrow$ & \NA
& 8.9   & $\rightarrow$ & \NA
& 0.735 & $\rightarrow$ & \NA \\
\arrayrulecolor{black!10}
\midrule[0.2pt]
\arrayrulecolor{black}
\quad Ours (TRC) & \multicolumn{18}{l}{} \\
\qquad $H_{\mathrm{T}}=150$\,s
& 66.7  & $\rightarrow$ & \NA
& 96.7  & $\rightarrow$ & 100.0
& \NA   & $\rightarrow$ & \NA
& \NA   & $\rightarrow$ & \NA
& \NA   & $\rightarrow$ & \NA
& \NA   & $\rightarrow$ & \NA \\
\qquad $H_{\mathrm{T}}=30$\,s
& 7.8   & $\rightarrow$ & 9.1
& 0.0   & $\rightarrow$ & 3.3
& 0.96  & $\rightarrow$ & 0.84
& 4.65  & $\rightarrow$ & 4.60
& 4.6   & $\rightarrow$ & 4.7
& 0.772 & $\rightarrow$ & 0.779 \\

\bottomrule
\end{tabular*}
\end{table}

\paragraph{Setup.}
Our user simulator follows calm turn-taking and neither backchannels nor interrupts. To examine whether our findings hold when the user also backchannels and interrupts, we replace it with GPT-Live-1, a commercial full-duplex model that operates in real time and decides on its own when to respond, backchannel, or interrupt. GPT-Live-1 receives the same persona and topic instructions as our simulator together with instructions to backchannel frequently and to interrupt mid-sentence. Across all sessions it produced 3.9 backchannels and 0.9 interruption attempts per minute on average. We treat its utterances of up to 1\,s as backchannels and count longer ones that begin while the model is speaking as interruption attempts. The timeout, retry, and termination rules and the metrics follow Section~\ref{sec:long-session-protocol}, except that the interaction runs in real time rather than in logical time. Given the cost and runtime of real-time interaction, we run 30 episodes per model and condition by pairing the Balanced persona with each of the ten topics three times. Base sessions last 10 minutes and the others 20 minutes as in Table~\ref{tab:long-session-stability}. Raon-SpeechChat is excluded because it cannot sustain real-time inference in our computing environment.

\paragraph{Degradation of the base models.}
Table~\ref{tab:gpt-live-base} compares the base models under our simulator and GPT-Live-1. With GPT-Live-1 the degradation of both models largely follows the patterns observed with our simulator, while its severity and onset differ. PersonaPlex loses responsiveness, spoken content quality, and voice consistency more severely than with our simulator. MiniCPM-o 4.5 again degrades mainly in responsiveness while keeping its spoken content quality, but the decline starts earlier. Already at 0--2\,min, 36.9\% of response opportunities time out, and 76.7\% of sessions terminate by 10\,min. The degradation is therefore not confined to calm turn-taking but also arises when the user backchannels and interrupts in real time.

\paragraph{Mitigation.}
Table~\ref{tab:gpt-live-mitigation} reports Ours (TRC) in 20-minute sessions with GPT-Live-1, together with TRC ($H_{\mathrm{T}}=30$\,s) on MiniCPM-o 4.5 (Appendix~\ref{app:stability-recall}). On PersonaPlex, Ours (TRC) mitigates the degradation through 20 minutes. It keeps spoken content quality and voice consistency well above the late-session Base level and terminates 10.0\% of sessions by 20\,min, compared with 36.7\% of Base sessions by 10\,min. Its response timeouts are more frequent than with our simulator but do not increase beyond the training horizon. On MiniCPM-o 4.5, in contrast, the default configuration does not mitigate the degradation and terminates 96.7\% of sessions by 10\,min, more than Base. This limitation was already partly visible with our simulator. TRC ($H_{\mathrm{T}}=30$\,s) substantially improves the mitigation, keeping response timeouts below 10\% and both spoken content quality and voice consistency stable while only one session (3.3\%) terminates through 20 minutes. Since both configurations are trained on the same trajectories from our simulator, the failure of the default configuration is unlikely to stem solely from overfitting to the simulator. These results suggest that TRC can generalize to interaction in which the user backchannels and interrupts, but that a single configuration applied across models may not be sufficient.

\subsection{Effect of Adaptation on Full-Duplex Behaviors}
\label{app:fdb}

\begin{table}[tp]
\caption{\textbf{Full-Duplex-Bench v1.0.} Pause handling is evaluated on the synthetic (Synth.) and Candor subsets. TOR denotes the take-over rate, Freq the backchannel frequency, JSD the Jensen--Shannon divergence of backchannel timing from human listeners, and GPT the GPT-4o score for responses to user interruptions; latencies are in seconds.}
\label{tab:fdb}

\centering
\small
\setlength{\tabcolsep}{3pt}

\begin{tabular*}{\linewidth}{@{\extracolsep{\fill}}l cc ccc cc ccc@{}}
\specialrule{\heavyrulewidth}{0pt}{\aboverulesep}
\multirow{2}{*}{\textbf{Method}}
& \multicolumn{2}{c}{\small\textit{Pause handling}}
& \multicolumn{3}{c}{\small\textit{Backchannel}}
& \multicolumn{2}{c}{\small\textit{Smooth turn-taking}}
& \multicolumn{3}{c}{\small\textit{User interruption}}
\\[-2pt]
\cmidrule(lr){2-3}
\cmidrule(lr){4-6}
\cmidrule(lr){7-8}
\cmidrule(lr){9-11}
& Synth.\ $\downarrow$ & Candor $\downarrow$
& TOR $\downarrow$ & Freq $\uparrow$ & JSD $\downarrow$
& TOR $\uparrow$ & Lat.\ $\downarrow$
& TOR $\uparrow$ & GPT $\uparrow$ & Lat.\ $\downarrow$
\\
\specialrule{\lightrulewidth}{\aboverulesep}{0pt}

\rowcolor{modelgray}
\multicolumn{11}{@{}l}{\raisebox{0pt}[\dimexpr\arraystretch\ht\strutbox+1.5pt\relax][\dimexpr\arraystretch\dp\strutbox+1.5pt\relax]{\textbf{PersonaPlex}}} \\
\addlinespace[2pt]
\quad Base       & 0.394 & 0.454 & 0.255 & 0.122 & 0.735 & 0.958 & 0.29 & 0.915 & 4.60 & 0.19 \\
\quad Ours (TRC) & 0.255 & 0.329 & 0.273 & 0.078 & 0.781 & 0.899 & 0.31 & 0.925 & 4.48 & 0.41 \\

\addlinespace[3pt]
\rowcolor{modelgray}
\multicolumn{11}{@{}l}{\raisebox{0pt}[\dimexpr\arraystretch\ht\strutbox+1.5pt\relax][\dimexpr\arraystretch\dp\strutbox+1.5pt\relax]{\textbf{MiniCPM-o 4.5}}} \\
\addlinespace[2pt]
\quad Base       & 0.212 & 0.588 & 0.218 & 0.017 & 0.910 & 0.958 & 0.36 & 0.870 & 2.84 & 0.50 \\
\quad Ours (TRC) & 0.058 & 0.120 & 0.073 & 0.011 & 0.945 & 0.849 & 0.65 & 0.710 & 2.15 & 0.46 \\

\addlinespace[3pt]
\rowcolor{modelgray}
\multicolumn{11}{@{}l}{\raisebox{0pt}[\dimexpr\arraystretch\ht\strutbox+1.5pt\relax][\dimexpr\arraystretch\dp\strutbox+1.5pt\relax]{\textbf{Raon-SpeechChat}}} \\
\addlinespace[2pt]
\quad Base       & 0.854 & 0.917 & 0.473 & 0.198 & 0.660 & 1.000 & 0.00 & 1.000 & 2.92 & 0.29 \\
\quad Ours (TRC) & 0.745 & 0.875 & 0.309 & 0.143 & 0.711 & 1.000 & 0.01 & 1.000 & 2.94 & 0.26 \\

\bottomrule
\end{tabular*}
\end{table}

\paragraph{Setup.}
To examine how the adaptation affects full-duplex behaviors that our training does not directly target, we evaluate the base models and Ours (TRC) on Full-Duplex-Bench v1.0~\citep{lin2025fullduplexbench} using its public implementation for transcription and metric computation. PersonaPlex uses the task-specific prompts specified in the benchmark repository. Since the benchmark specifies no configuration for MiniCPM-o 4.5 or Raon-SpeechChat, we apply the same prompts to them. All models use the default sampling settings of their released inference code, and the base and TRC variants of each model share identical settings. The user-interruption judge is GPT-4o (gpt-4o-2024-08-06) because GPT-4-turbo, used in the original script, has been retired. The technical report of Raon-SpeechChat~\citep{kim2026raonspeech} evaluates this benchmark with an internal evaluator that uses different turn-detection criteria, so our absolute values for this model differ substantially from the reported ones. But both of its variants are compared under the same public protocol.

\paragraph{Results.}
As Table~\ref{tab:fdb} shows, the adapted models still take the turn in most cases when the user yields it and respond to most user interruptions, with TOR of at least 0.849 and 0.710, respectively. The adaptation nevertheless shifts these behaviors to some extent. Two changes are shared by all three models. They handle user pauses better, with lower TOR on both pause subsets, and they produce backchannels less frequently and with timing further from that of human listeners (higher JSD). In addition, PersonaPlex and MiniCPM-o 4.5 take the turn less reliably when the user yields it (TOR 0.958 $\rightarrow$ 0.899 and 0.958 $\rightarrow$ 0.849), and MiniCPM-o 4.5 handles user interruptions markedly worse (TOR 0.870 $\rightarrow$ 0.710 and GPT score 2.84 $\rightarrow$ 2.15).

The decline in interruption handling can be expected from the trajectory design. The simulated user never interrupts the model (Section~\ref{sec:long-session-protocol}), so interruption handling receives no supervision during training. The backchannels and floor-taking of the models, in contrast, were left unconstrained during trajectory construction, yet these behaviors shifted as well. Taken together, these results suggest that adaptation on a small dataset collected with a single type of user simulator is cost-efficient but can shift full-duplex behaviors beyond what the training targets, in both favorable and unfavorable directions.

\section{Limitations and Future Work}
First, our simulator did not initiate interruptions or backchannels. An additional evaluation with GPT-Live-1 as the user (Appendix~\ref{app:gpt-live}) suggests that our findings largely hold when the user backchannels and interrupts. This evaluation, however, remains small in scale and does not cover Raon-SpeechChat, which could not sustain real-time inference in our computing environment. Extending such evaluation would enable a more comprehensive assessment of full-duplex interaction over long sessions.
Second, recall accuracy remained limited under high compression. Although TRC improves the Pareto frontier of the stability--recall trade-off, it does not fully overcome this trade-off, and its recall remains well below that of Base on MiniCPM-o 4.5 (Section~\ref{sec:stability-recall}). Future work should explore training objectives that improve the retention and use of specific information while further examining the trade-off between compression and recall.
Third, adaptation on limited data collected with our simulator shifts some native full-duplex behaviors of the models that the training does not directly target, such as interruption handling on MiniCPM-o 4.5 and backchannel frequency (Appendix~\ref{app:fdb}). Future work should consider training on trajectories generated with a user that backchannels and interrupts, such as GPT-Live-1. How incorporating TRC from pretraining rather than through self-distillation affects long-session stability and these behaviors is another direction that requires further study.

\FloatBarrier
\endgroup

\end{document}